\documentclass[runningheads]{llncs}
\usepackage{float}
\usepackage{eccv}

\usepackage{eccvabbrv}

\usepackage{graphicx}
\usepackage{booktabs}
\usepackage{xcolor} 
\usepackage{multirow}
\usepackage{makecell}
\usepackage{colortbl}
\usepackage{algorithm}
\usepackage{algorithmic}
\usepackage{pifont}
\usepackage{wrapfig}
\usepackage{enumitem}

\usepackage[accsupp]{axessibility}  %
\usepackage{tcolorbox}
\usepackage{float}
\tcbuselibrary{breakable}
\usepackage{listings}

\definecolor{promptbg}{HTML}{FBF3FB}      
\definecolor{promptframe}{HTML}{F0E0F0}   

\definecolor{cellRed}{HTML}{F3D9DA}        
\definecolor{cellRedLight}{HTML}{FAECEC}   

\definecolor{cellYellow}{HTML}{ECD8F0}     
\definecolor{cellYellowLight}{HTML}{F6EBF8} 

\definecolor{cellBlue}{HTML}{C9EAF6}      
\definecolor{cellBlueLight}{HTML}{E4F4FB} 

\definecolor{cellPurple}{HTML}{ECD8F0}     
\definecolor{cellPurpleLight}{HTML}{F6EBF8} 

\newtcolorbox{promptbox}[1][]{%
  colback=promptbg,
  colframe=promptframe,
  coltitle=black,
  fonttitle=\small\bfseries,
  title={#1},
  breakable,
  boxrule=0.6pt,
  arc=2pt,
  left=6pt, right=6pt, top=4pt, bottom=4pt,
}

\newcommand{\hlp}[1]{\colorbox{green!15}{#1}}
\definecolor{degraded}{RGB}{255,224,194}   
\newcommand{\cc}{\cellcolor{degraded}}

\usepackage{hyperref}

\begin{document}

\title{O-VAD: Industrial Video Anomaly Detection through Object-Centric Tracking and Reasoning}

\titlerunning{O-VAD}

\author{Mei Yuan\inst{1} \and
Qi Long\inst{1}\and
Qifeng Wu\inst{1}\and
Zhenyang Li\inst{2}\and
Yizhou Zhao\inst{1}\and
Lei Wang\inst{3}\and
Yang Liu\inst{1}\and
Min Xu\inst{1}\thanks{Corresponding author.}
}

\authorrunning{M.~Yuan et al.}

\institute{Carnegie Mellon University \and
University of Alabama at Birmingham \and
Griffith University \\
\email{meiyuan@andrew.cmu.edu}, \email{mxu1@cs.cmu.edu} \\
\url{https://o-vad.github.io/}}

\maketitle

\begin{figure}[H]
    \centering
    \includegraphics[width=\linewidth]{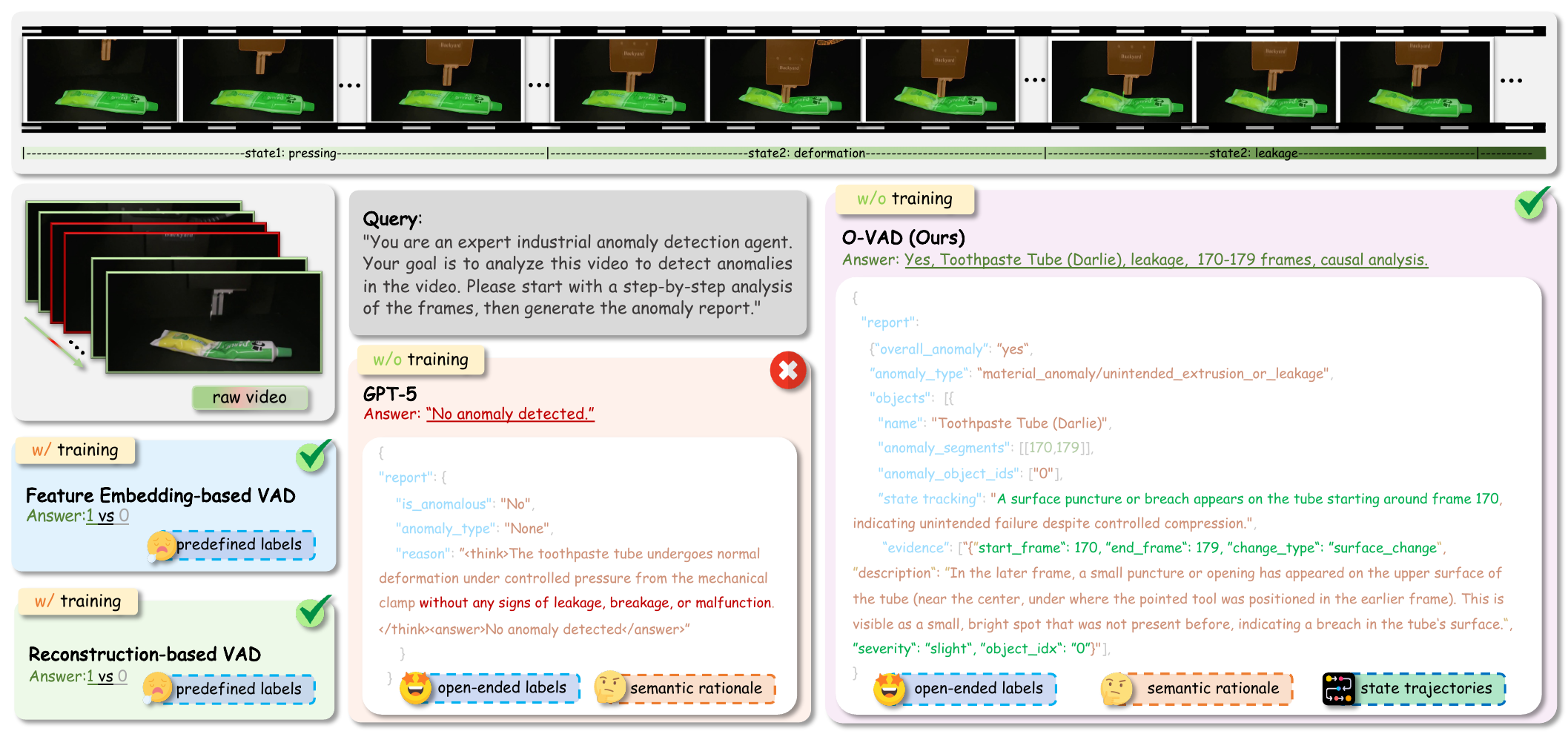}
    \caption{\textbf{Why object state evolution matters for industrial anomaly detection.} The toothpaste tube undergoes pressing and deformation before a subtle leakage emerges. Traditional methods predict binary classifications. VLM prompting fails to detect it due to the lack of evidence. O-VAD tracks object-wise state changes and produces an open-ended anomaly report with in-depth analysis.}
    \label{fig:intro}
\end{figure}

\begin{abstract}
Industrial Video Anomaly Detection (IVAD) aims to identify anomalous objects and events in an industrial process, which is crucial for modern manufacturing and quality control systems.
Existing VLM-based anomaly reasoning methods are capable of detecting open-ended anomalies in general domains. However, their performance declines in industrial settings characterized by intricate object transformations, strict physics, and procedural constraints.
To tackle the complexity of such interaction-intensive detection, we introduce a training-free agentic framework \textit{\textbf{O-VAD}} for anomaly detection free of domain-specific knowledge, emphasizing object state evolution like humans inspectors.
It is designed to track spatial-temporal dynamics and underlying transformations of detected objects over time, and then reason over the object-wise temporal state trajectories to identify abnormal objects in grounded frames.
Our method overcomes limitations of prior approaches that rely on retraining on normal clips or injecting domain knowledge as context for test-time inference.
Extensive experiments on three IVAD datasets demonstrate that our method outperforms frontier VLMs, agentic frameworks, and traditional VAD methods fine-tuned on the respective datasets, while providing interpretable reports over anomaly processes and types.

\end{abstract}

\section{Introduction}

Industrial Video Anomaly Detection (IVAD) aims to identify anomalous objects and events in an industrial video process. 
It plays a critical role in modern manufacturing and quality control systems, where automated visual inspection can significantly reduce costs and improve production reliability~\cite{liu2024ipad}. 
Industrial processes are extremely challenging and different from general domains, since they are usually characterized by: (i) \textit{complex object transformations}: objects undergo substantial physical and functional state changes through cutting, assembly, heating, pressing, and other manufacturing operations;
(ii) \textit{strict physical and procedural constraints}: anomalies manifest as violations of expected spatial configurations, periodic procedures, or physical laws;
(iii) \textit{high interpretability requirements}: 
operators need explicit explanations of detected anomalies for root cause analysis and corrective actions. 

Traditional Video Anomaly Detection (VAD) methods primarily adopt two paradigms: reconstruction-based approaches~\cite{he2024mambaad, fan2024revitalizing, zavrtanik2021draem, zhang2024realnet, zhangdiffusionad, fang2023fastrecon} reconstruct anomalous samples to their corresponding normal counterparts and calculate the reconstruction error, and embedding-based~\cite{jiang2022softpatch, hyun2024reconpatch, roth2022towards, defard2021padim, guo2025dinomaly} methods focus on modeling the feature embeddings of normal samples and measure deviations.
But they typically follow the “one-class-one-model” learning paradigm, requiring plentiful normal samples for each object class to learn its distribution~\cite{gu2024anomalygpt}, 
which makes it impractical for industrial anomaly detection settings and less suitable for dynamic production environments.
Given that \textbf{\textit{traditional VADs have limited generalization abilities and binary classifications provide no semantic rationale for their decisions}}, recent works~\cite{huang2025vad, lin2025unified, lin2025vlm, zhao2025omniad, kangjudo, gu2024anomalygpt, zhu2025vau, zou2026unlocking} propose a series of VLM-based anomaly detection methods, featuring semantic anomaly understanding and test-time detection inference.

Although VLMs have unlocked their potential for detecting anomalies augmented with semantic-visual understanding in general domains, there are still two issues that remain up in the air:
(1) Their performance in industrial settings declined due to \textit{lack of domain-specific knowledge and fine-grained annotations}.
Recent training-free studies~\cite{jiang2024mmad} provide external knowledge or normal samples in context for test-time inference to alleviate this limitation. 
However, \textbf{\textit{they heavily rely on sophisticated prompt design to caption videos, inject knowledge, and then assign anomaly scores by off-the-shelf VLMs}}. The overly dependence on external context leads to misaligned responses that prioritize contextual plausibility over accuracy~\cite{kangjudo}.
(2) Video/frame level features or descriptions have \textit{weak understanding of localized details within objects over time}, which hinders their reliable and accurate interaction-intensive IVAD reasoning.
Some VLM-based methods exploited the spatio-temporal fusion~\cite{wu2024weakly, li2022scale,liu2019exploring, sun2023long, wu2021weakly}, and detect anomalies from when and where perspectives~\cite{lin2025unified, huang2022hierarchical}.
However, \textbf{\textit{they lose focus on object-centric physics and interactions over time}}. It is the core when human inspectors detect object-centric state changes throughout industrial processes, especially when objects undergo substantial physical and functional state changes.

To tackle these two issues, we propose \textbf{\textit{O-VAD}}, a training-free agentic framework that detects anomalies by tracking object state evolution over time---free of any domain-specific knowledge, labels, or predefined anomaly taxonomy.
(1) For the first issue, rather than injecting external knowledge or relying on complex prompt design, O-VAD enables the VLM to \emph{build its own understanding} through a ``ground$\rightarrow$track$\rightarrow$reason'' agentic pipeline. Objects are first detected and segmented via \textbf{\textit{VLM-grounded masking}} with SAM3~\cite{carion2025sam}, then continuously tracked and queried across frames to accumulate structured evidence. A cascaded chain-of-thought (CoT)
then derives anomaly judgments from the VLM's own internalized commonsense and physical knowledge, mirroring the cognitive process of human inspectors.
(2) For the second issue, we shift the unit of analysis from holistic frames to individual objects. An \textbf{\emph{object state tracker}} maintains per-object representations and queries the VLM at key temporal transitions to capture fine-grained state changes, \eg deformation, material release, surface alteration. These structured state trajectories ground the downstream reasoning, making detection both \emph{object-centric} and \emph{spatio-temporally aware}, and yielding \emph{open-ended}, \emph{fine-grained} outputs: anomaly types, grounded abnormal frames, affected objects, and causal analyses.

Result-wise, despite using no labor-intensive annotations, the proposed agentic framework delivers state-of-the-art performance on any level anomaly detection, surpassing frontier VLMs and agentic frameworks on quantitative and qualitative evaluations. 

To summarize our contributions:

(1) We present a generalizable data curation pipeline and apply it to three object-centric IVAD datasets (IPAD, Phys-AD and LiquidAD) with fine-grained annotations of object trajectories across frames.

(2) We propose a training-free, three-stage agentic reasoning framework that requires no domain-specific knowledge or fine-tuning, performing object discovery, spatiotemporal tubelet construction withy open-ended state change detection, and chain-of-thought anomaly reasoning with visual verification.

(3) Our framework achieves state-of-the-art performance on three IVAD datasets at both video- and frame- levels, outperforming frontier VLMs (Qwen3-VL-32B and GPT-5), two agentic framework for VAD, and fine-tuned traditional AD methods, while providing interpretable reasoning over anomaly processes and open-ended types.

\section{Related Work}
\label{sec:related}

\paragraph{\textbf{Traditional Video Anomaly Detection.}}
Traditional video anomaly detection methods primarily adopt two paradigms: reconstruction-based approaches~\cite{he2024mambaad, fan2024revitalizing, zavrtanik2021draem, zhang2024realnet, zhangdiffusionad, fang2023fastrecon} flag anomalies by their elevated reconstruction error, while embedding-based methods~\cite{jiang2022softpatch, hyun2024reconpatch, roth2022towards, defard2021padim, guo2025dinomaly} model the feature distribution of normal samples and measure deviations from it. Despite strong benchmark performance, both share three limitations. First, they follow a ``one-class-one-model'' paradigm~\cite{gu2024anomalygpt}, requiring abundant normal samples and a separate model per object class, and must be re-trained for every unseen domain or anomaly type~\cite{li2026vadtree}, making them impractical for novel categories and dynamic production lines. Second, they are largely black-box, emitting an anomaly score without semantic rationale~\cite{wu2024weakly}, which has motivated growing interest in explainable, language-grounded anomaly understanding~\cite{li2026vadtree, zheng2026iad}. Third, they compress entire frames into holistic features, neglecting region-level cues and over-relying on dominant background context~\cite{li2025video}. Our O-VAD addresses these limitations with a training-free, object-centric framework that requires no per-class training, produces semantic explanations for its decisions, and reasons over region-level evidence.

\paragraph{\textbf{Multimodal Video Anomaly Detection}}

Recent works~\cite{huang2025vad, lin2025unified, lin2025vlm, zhao2025omniad, kangjudo, gu2024anomalygpt, zhu2025vau, zou2026unlocking} revisit anomaly detection with VLM-based methods, valued for their generalizability and explainability, especially in the zero/few-shot setting. These approaches rely on sophisticated prompt design that captions videos and assigns anomaly scores via off-the-shelf VLMs. To strengthen spatio-temporal awareness, some methods couple this with auxiliary temporal modeling or prompting~\cite{wu2024weakly, li2022scale, liu2019exploring, sun2023long, wu2021weakly}; for instance, a unified zero-shot framework~\cite{lin2025unified} chains temporal detection, spatial localization, and textual explanation through a test-time reasoning process over foundation models, requiring no additional training. Yet these methods largely overlook that superficial temporal caption sequences lose focus on the object-centric dynamics and interactions over time—precisely the cues human inspectors rely on to detect anomalous objects and events throughout industrial processes. In contrast, our O-VAD explicitly models object-centric spatio-temporal evidence while retaining the training-free and explainable advantages of VLM-based detection.

\section{Method}
\label{sec:method}

\begin{figure}[t]
    \centering
    \includegraphics[width=1\linewidth]{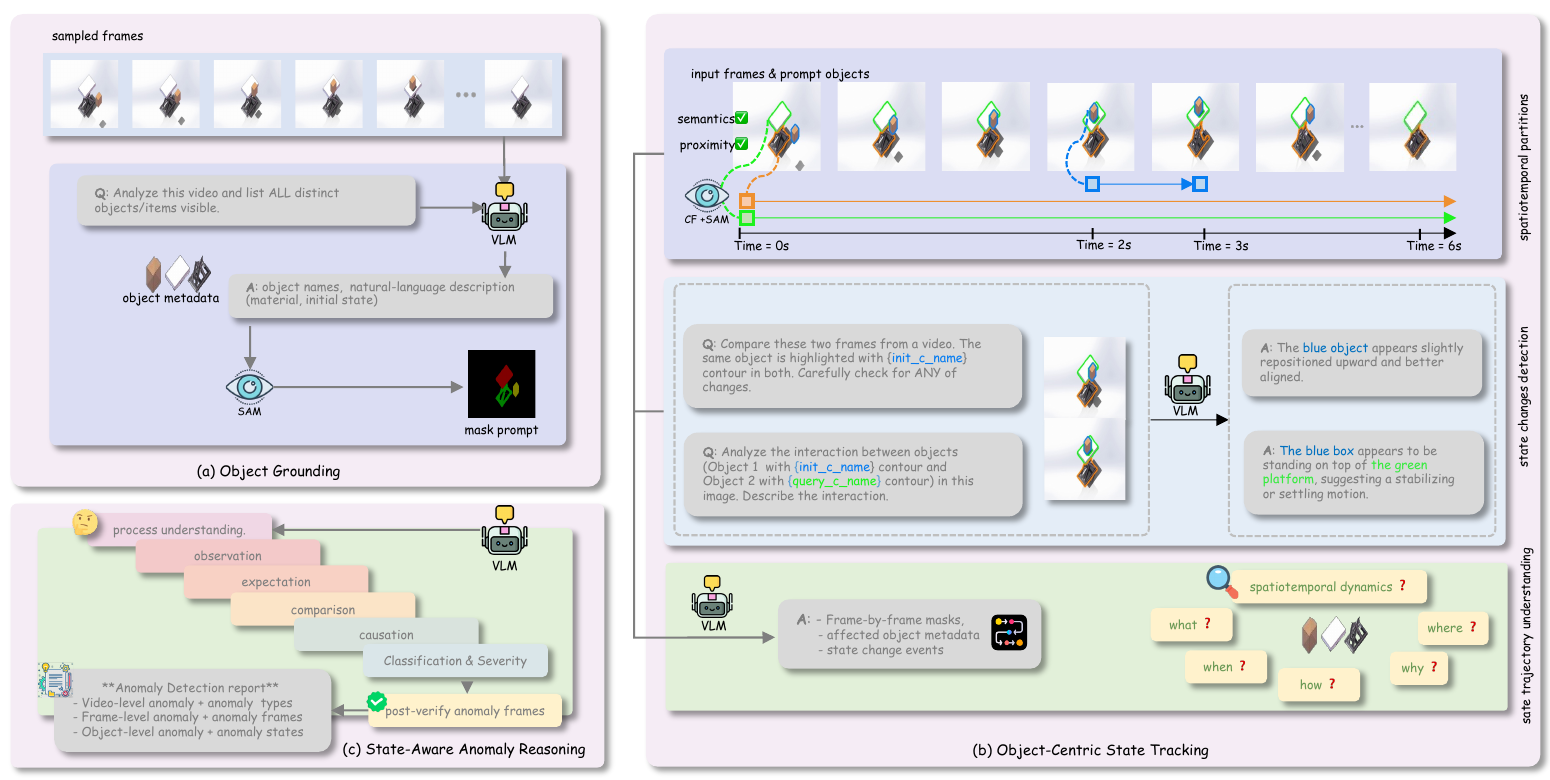}
    \caption{\textbf{Overview of O-VAD.} \textbf{(a)}~Stage~1 samples multiple frames and queries a VLM to discover all objects, then segments them with SAM to produce initial masks and metadata. \textbf{(b)}~Stage~2 constructs spatiotemporal tubelets via CropFormer~\cite{qi2022high} and SAM2\cite{ravi2024sam2}, recovers missing tracks through semantic and proximity priors, and query the VLM for open-ended state change detection and inter-object interaction analysis. \textbf{(c)}~Stage~3 feeds the accumulated object metadata and state change events into a multi-step chain-of-thought reasoning chain that distinguishes process actions from failure outcomes, followed by visual verification of candidate anomalies.}
    \label{fig:framework}
\end{figure}

\paragraph{\textbf{Overview.}}

Given an industrial process video $\mathcal{V}=\{I_t\}_{t=1}^{T}$, O-VAD produces a structured anomaly report for each detected anomaly, it outputs the anomaly type, severity, affected object, temporal localization, and a natural-language causal explanation---all without any domain-specific knowledge, predefined taxonomy, or training data.
To achieve this, O-VAD adopts a three-stage agentic pipeline (Fig.~\ref{fig:framework}).
Stage~1 (\S\ref{sec:stage1}) discovers and segments all task-relevant objects via VLM-grounded masking with SAM~\cite{carion2025sam}, rather than manual annotation.
Stage~2 (\S\ref{sec:stage2}) tracks each object through potential state trajectories via spatiotemporal partitioning, querying the VLM at temporal transitions to detect fine-grained state changes and produce per-object state trajectories.
Stage~3 (\S\ref{sec:stage3}) reasons over the accumulated evidence through a multi-step chain-of-thought that separates abnormal stage changes from expected process outcomes, yielding anomaly judgments grounded in the VLM's own commonsense and physical reasoning.

\subsection{Problem Formulation}
\label{sec:formulation}
Our O-VAD decomposes the problem into an object-centric formulation. We first detect $K$ objects $\mathcal{O}=\{o_k\}_{k=1}^{K}$ with masks $\{m_k^{(t)}\}$ via VLM-grounded segmentation, then track each object's state trajectory $\mathcal{S}_k = \{(m_k^{(t)}, \phi_k^{(t)})\}_{t=1}^{T}$, where $m_k^{(t)}$ is the mask of object $k$ at frame $t$ and $\phi_k^{(t)}$ is a VLM-generated natural-language description of its physical state (\eg, name, material, surface condition, and detected state changes). Object-level anomaly reasoning is then performed as:
\begin{equation}
    \hat{y}_{t,k} = \theta_{\text{VLM}}\!\big(p_{\text{CoT}} \oplus \mathcal{S}_k \oplus c\big), \qquad
    \hat{y}^{\text{frm}}_t = \max_{k}\; \hat{y}_{t,k}, \qquad
    \hat{y}^{\text{vid}} = \max_{t}\; \hat{y}^{\text{frm}}_t,
\end{equation}
where $\hat{y}_{t,k}\in[0,1]$ is the anomaly confidence for object $k$ at frame $t$, $c$ is the optional video caption, $p_{\text{CoT}}$ is a cascaded chain-of-thought prompt that guides the VLM through cognitive anomaly reasoning, and $\oplus$ denotes concatenation of the prompt context. The frame- and video-level scores are obtained by max-pooling over objects and frames, respectively. By conditioning on per-object state trajectories $\mathcal{S}_k$ rather than raw frames alone, O-VAD produces fine-grained, open-ended anomaly predictions with grounded frames, affected objects, and causal explanations; the discrete anomaly set $\mathcal{A}$ used for reporting is obtained from these scores after the reasoning and verification of Stage~3 (\S\ref{sec:stage3}).

\subsection{Stage~1: Automated Object Grounding}
\label{sec:stage1}

Rather than relying on a single reference frame, we sample a set of candidate frames $\{I_{t_j}\}_{j=1}^{N_f}$ spanning the video so that objects initially occluded or absent can still be discovered.
For each candidate frame $I_{t_j}$, a VLM generates a structured object inventory:
\begin{equation}
  \mathcal{O}_{t_j}
  = \phi_{\mathrm{VLM}}(I_{t_j})
  = \bigl\{(n_k,\, d_k,\, b_k)\bigr\}_{k=1}^{N_{t_j}},
  \label{eq:vlm_inventory}
\end{equation}
where $n_k$ is the object name, $d_k$ a natural-language description (material, initial state), and $b_k$ a spatial cue.
Inventories across frames are merged into a deduplicated set $\mathcal{O}=\bigcup_j \mathcal{O}_{t_j}$.
Each object $o_k\!\in\!\mathcal{O}$ is then segmented in a reference frame $I_1$ using SAM3~\cite{carion2025sam}:
\begin{equation}
  m_k^{(1)} = \mathrm{SAM}(I_1,\, o_k),
  \qquad
  \mathcal{M}_1 = \bigl\{m_k^{(1)}\bigr\}_{k=1}^{|\mathcal{O}|}.
  \label{eq:sam_masks}
\end{equation}
The output of Stage~1, including initial masks $\mathcal{M}_1$ together with object metadata, grounds all subsequent tracking and reasoning in concrete, per-object evidence.

\subsection{Stage~2: Object-Centric State Tracking}
\label{sec:stage2}

The second stage tracks each object across all frames, detecting and describing state changes even when objects undergo significant transformations. We build upon TubeletGraph~\cite{sun2025tracking} for tracking through transformations, and extend it with two novel components: object interaction analysis  and open-ended state change annotation.

\paragraph{\textbf{Spatiotemporal Partitioning and Tubelet Construction}.}

We construct a spatiotemporal partition $\mathcal{P}$ of the video that associates every pixel region across all frames with a tracked entity. An entity segmentation model (CropFormer~\cite{qi2022high}) produces per-frame spatial partitions $\mathcal{E}_t = \mathrm{CF}(I_t)$, and each entity $e_1^i \in \mathcal{E}_1$ (including the initial object masks $\mathcal{M}_1$) is tracked forward via SAM2~\cite{ravi2024sam2} to form tubelets:
\begin{equation}
    \label{eq:partition}
    \mathcal{P}_\text{init} = \{P_i\}_{i=1}^{|\mathcal{E}_1|}, \quad P_i = \{e_t^i\}_{t=1}^{T}.
\end{equation}
New tubelets are incrementally added whenever track-less regions emerge. For each entity $\hat{e}_t^j \in \mathcal{E}_t$ at frame $t > 1$, a new tubelet is initiated if less than $\tau_\text{coverage}$ of its area is covered by existing tubelets.

\paragraph{\textbf{State Tracking.}}

To recover missing object tracks after state transformations (\eg, an object breaking apart or releasing material), candidate tracks that emerge after the initial frame are evaluated using spatial proximity and semantic consistency priors.

\textit{Spatial Proximity.} For a candidate track $C = \{c_s, c_{s+1}, \ldots, c_T\}$ beginning at frame~$s$ and the prompt object track $P = \{p_1, p_2, \ldots, p_T\}$, the spatial proximity score is:
\begin{equation}
    S_\text{prox}(C, P) = \max_{j \in \{1,2,3\}} \frac{|c_s \cap m_s^j|}{|c_s|},
\end{equation}
where $\{m_s^j\}_{j=1}^{3}$ are the three candidate masks from SAM2's multi-mask output at frame~$s$. A candidate is considered proximal if $S_\text{prox}(C, P) > \tau_\text{prox}$.

\textit{Semantic Consistency.}\quad For a mask $M$ and frame $I$, we compute the masked CLIP~\cite{radford2021clip} feature $f(M, I) = \mathrm{Pool}(\mathrm{CLIP}(I), M)$ via mask-pooling. The semantic similarity is:
\begin{equation}
    S_\text{sem}(C, P) = \max_{\substack{i \in \{1,\ldots,s{-}1\},\\ j \in \{s,\ldots,T\}}} f(p_i, I_i) \cdot f(c_j, I_j)^\top.
\end{equation}
The final set of valid continuation tracks is:
\begin{equation}
    \mathcal{V} = \{C \in \mathcal{P} \mid C \text{ begins at } t > 0,\; S_\text{prox}(C, P) > \tau_\text{prox},\; S_\text{sem}(C, P) > \tau_\text{sem}\}.
\end{equation}
The complete tracking result $\mathcal{T} = P \cup \mathcal{V}$ captures the object through transformations.

\paragraph{\textbf{State Change Detection and Understanding.}}
For object-aware frame pairs, the VLM is prompted with masked visualizations of both frames and asked to identify state changes through visual understanding.
Critically, we adopt an \textit{open-ended labeling scheme}: both the change type (\eg, compression deformation, material release, rotational resistance) and the change cause (\eg, object leakage, environmental lighting) are generated as free-form natural language by the VLM. This enables the system to describe novel phenomena for generalizing across diverse industrial processes.
Each detected state change event $e$ is represented as a tuple:
\begin{equation}
   e=(t_{\text{start}},\; t_{\text{end}},\; \textit{type},\; \textit{cause},\; \textit{desc},\; \textit{sev},\; k),
   \label{eq:state_event}
\end{equation}
where $t_{\text{start}}$ and $t_{\text{end}}$ are the bounding frame indices, \textit{type} is the free-form change type, \textit{cause} is the causal label, \textit{desc} is a natural-language description, $\textit{sev} \in \{\text{none}, \text{slight}, \text{moderate}, \text{severe}\}$ is the severity, and $k$ is the affected object index.

\subsection{Stage~3: State-Aware Anomaly Reasoning}
\label{sec:stage3}

We formulate \textit{cognitive anomaly reasoning} as a cascaded chain-of-thought (CoT) task over the accumulated object states. 
The final stage synthesizes the tracking metadata, state change events, and visual evidence into an anomaly detection decision via multi-step reasoning. 
The key design principle is open-ended classification: the VLM is not constrained to a fixed taxonomy but reasons freely from its domain knowledge, grounded in the tracking evidence.
Finally, we have both the anomaly set and the complete reasoning trace for full interpretability.

\paragraph{\textbf{Cognitive Anomaly Reasoning.}}
\label{sec:cot}

The core anomaly reasoning proceeds through a cascaded chain-of-thought (CoT) that mirrors how a human quality inspector diagnoses failures.
The VLM receives, in a single call, the object metadata, filtered state changes, inferred task context, video caption, and sampled frames, and is instructed to reason through sequential steps with confidence scores. 
\begin{enumerate}
    \item \textit{Process Understanding}. The VLM identifies the industrial process and its purpose, then triages each state change into expected process, potential anomaly, and normal noise. 
    
    \item \textit{Observation}. The VLM cites specific object IDs, frame ranges, change types, and severities from the tracking data, focusing on events triaged as potential anomalies.

    \item {\textit{Expectation}.} The VLM articulates the pass/fail criteria for the identified test, distinguishing expected mechanical responses from failure-indicating outcomes.

    \item {\textit{Comparison}.} For each candidate anomaly, the VLM determines whether it represents a genuine failure outcome or a mechanical response that was mis-categorized.

    \item {\textit{Causation}}. The VLM reasons about root causes without being constrained to predefined categories.

    \item {\textit{Classification} \& \textit{Severity}.} A free-form anomaly type is assigned and each confirmed anomaly receives a severity rating based on its impact on task completion, safety implications, and reversibility.

\end{enumerate}

The reasoning chain outputs a set of candidate anomalies $\{a\}$, where each $a$ carries a free-form anomaly type, an affected object index, a frame range, a severity level, and an initial confidence $c_{\text{orig}}(a)\in[0,1]$ produced by the VLM. These candidates are then refined by the verification step below to form the final anomaly set $\mathcal{A}$.

\paragraph{\textbf{Post Visual Verification.}}
\label{sec:verify}
To suppress false positives, each candidate anomaly $a$ is gated by its initial confidence $c_{\text{orig}}(a)$: anomalies with $c_{\text{orig}}(a) > \tau_{\text{hi}}$ bypass verification (they are sufficiently certain), those with $c_{\text{orig}}(a) < \tau_{\text{lo}}$ are discarded outright, and only the remaining candidates are verified against their evidence frames:
\begin{equation}
  (\texttt{verified},\, c_{\text{ver}}) = \phi_{\text{VLM}}\big(\{I_{t}\}_{t \in \text{evidence}},\, c,\, a\big).
\end{equation}
The verification step explicitly checks whether the claimed anomaly is actually a normal process behavior visible in the caption $c$ or frames. The final confidence is then computed via a multiplicative gating scheme:
\begin{equation}
  c_{\text{final}}(a) = \begin{cases}
  c_{\text{orig}}(a) \cdot c_{\text{ver}} & \text{if } \texttt{verified} = \texttt{true}, \\
  c_{\text{orig}}(a) \cdot (1 - c_{\text{ver}}) & \text{if } \texttt{verified} = \texttt{false},
  \end{cases}
  \label{eq:conf_gate}
\end{equation}
and the anomaly is retained only if $c_{\text{final}}(a) \geq \tau_{\text{conf}}$. We use $\tau_{\text{hi}} = 0.8$, $\tau_{\text{lo}} = 0.2$, and $\tau_{\text{conf}} = 0.3$ in all experiments. The final anomaly set $\mathcal{A}$ and the complete reasoning trace together form the output report $\mathcal{R}$, providing full interpretability.

\section{Experiment}

\subsection{Experiment Setup.}

\paragraph{\textbf{Datasets \& evaluation metrics.}}
We evaluate on the official test splits of three industrial benchmarks spanning diverse anomaly types and interaction modalities. (i) \textit{Phys-AD}~\cite{li2025towards}: a large-scale, physics-grounded dataset from a real robot arm and motor, with 22 object categories and 47 defect types across 6,434 videos (60~FPS); its anomalies require physical reasoning over interactions such as pressing, rotating, and grasping. We report per-category video-level AUROC, aggregate Acc/P/R/F1 at the video and anomaly-type levels, and BERTScore for anomaly-type semantic alignment. (ii) \textit{LiquidAD}~\cite{dabouei2025deep}: liquid-transfer anomalies in automated labs, 2,251 videos (30~FPS) of 8 parallel pipettes dispensing into 8 test tubes; we report video- and frame-level Acc/P/R/F1/AUROC. (iii) \textit{IPAD}~\cite{liu2024ipad}: industrial-manufacturing processes ($\sim$2{,}000 clips over 16 device types, synthetic and real) that define anomalies as deviations from reference normal samples rather than physical commonsense; we evaluate under this reference-based protocol and report video- and frame-level metrics. For VLM-based methods, an LLM-as-judge protocol assesses whether a predicted anomaly type semantically matches the ground truth, enabling fair comparison across heterogeneous outputs. Full dataset details and per-scenario results are in the supplement (\S\ref{sec:supp_datasets} and \S\ref{sec:res-IPAD}).

\paragraph{\textbf{Hyperparameters \& experiment details.}}

O-VAD is fully training-free; unless noted, we use the following defaults. \textit{Stage 1 (VLM-grounded masking)} uses GPT-5~\cite{singh2025openai} as the VLM and SAM3~\cite{carion2025sam} for concept-aware zero-shot segmentation, with a permissive detection threshold of 0.1 to maximize discovery recall. \textit{Stage 2 (object state tracking)} partitions each video into tubelets via CropFormer~\cite{qi2022high} for per-frame entity segmentation and SAM2~\cite{ravi2024sam2} for temporal propagation, sampling every 10th frame; track recovery is gated by spatial-proximity and semantic-consistency thresholds. \textit{Stage 3 (cognitive anomaly reasoning)} applies the 6-step CoT prompt over the accumulated state trajectories, key frames, and object metadata to produce the structured report, after which confidence-tiered visual verification suppresses false positives. Phys-AD videos are subsampled with dynamic FPS to reduce redundancy while preserving key state transitions. All experiments run on a single NVIDIA H200 GPU (141~GB VRAM). Full threshold hyperparameters and model-selection rationale are in the supplement (\S\ref{sec:hyperparams} and \S\ref{sec:model-selection}).

\paragraph{\textbf{Baseline Methods.}}

We compare O-VAD against three groups. (i) \textit{Traditional VADs}: the best method from each paradigm on Phys-AD---MNAD.p~\cite{park2020learning} (unsupervised, prediction-based memory normality) and S3R~\cite{wu2022self} (weakly-supervised sparse representation); both require training data and produce only binary scores without explanations. (ii) \textit{Direct Prompting}: frontier VLMs in a zero-shot ``caption-then-classify'' setup---Qwen3-VL-32B~\cite{bai2025qwen3} (extended thinking) and GPT-5~\cite{singh2025openai}---receiving raw frames and a standard anomaly-detection prompt, without object grounding or state tracking. (iii) \textit{Agentic Workflows}: two recent training-free frameworks. URF-ZS-HVAA~\cite{lin2025unified} chains temporal detection, spatial localization, and textual explanation via intra- and inter-task reasoning; VERA~\cite{ye2025vera} verbalizes learnable guiding questions for a frozen VLM, decomposing the task into yes/no perceptual queries aggregated into a video-level verdict.

\begin{table}[t]
\centering
\caption{\textbf{Video-level, type-level, and frame-level Acc/P/R/F1/AUROC (\%) on Phys-AD~\cite{li2025towards}, LiquidAD~\cite{dabouei2025deep}, and IPAD~\cite{liu2024ipad}.}
Methods are grouped into traditional VADs (\textbf{[Trad.]}), direct-prompting VLMs (\textbf{[Direct Prompting]}), agentic workflows (\textbf{[Workflow]}), and our \textbf{[O-VAD]}. Type-level BERTScore and LLM-judge scores are reported only for methods that emit free-form anomaly-type descriptions.}
\label{tab:main}
\setlength{\tabcolsep}{2pt}
\renewcommand{\arraystretch}{1.3}
\resizebox{\columnwidth}{!}{%
\tiny
\begin{tabular}{lll cc cc cc c}
\toprule
\multicolumn{3}{l}{}
& \multicolumn{2}{c}{\textbf{[Trad.]}}
& \multicolumn{2}{c}{\textbf{[Direct Prompting]}}
& \multicolumn{2}{c}{\textbf{[Workflow]}}
& \multicolumn{1}{c}{\textbf{[O-VAD(Ours)]}} \\
\cmidrule(lr){4-5} \cmidrule(lr){6-7} \cmidrule(lr){8-9} \cmidrule(lr){10-10}
\textbf{Dataset} & \textbf{Level} & \textbf{Metric}
& MNAD.p$^\dagger$ & S3R$^\dagger$
& Qwen3$^\ast$ & GPT-5$^\ast$
& URF$^\ast$ & VERA$^\dagger$
& \textbf{O-VAD}$^\ast$ \\
\midrule

\multirow{7}{*}{\textbf{Phys-AD}} & \multirow{5}{*}{video}
  & Acc       & \cellcolor{cellRedLight}0.635 & 0.591 & 0.454 & \cellcolor{cellRed}0.640 & 0.329 & 0.470 & 0.592 \\
& & P         & 0.713 & \cellcolor{cellRedLight}0.739 & 0.717 & 0.690 & \cellcolor{cellRed}0.853 & 0.500 & 0.724 \\
& & R         & \cellcolor{cellRed}0.803 & \cellcolor{cellRedLight}0.645 & 0.368 & 0.580 & 0.053 & 0.050 & 0.625 \\
& & F1        & \cellcolor{cellRed}0.755 & \cellcolor{cellRedLight}0.689 & 0.486 & 0.630 & 0.100 & 0.091 & 0.621 \\
& & AUC       & 0.495 & \cellcolor{cellRedLight}0.555 & 0.513 & 0.502 & 0.426 & 0.456 & \cellcolor{cellRed}\textbf{0.584} \\
\cline{2-10}
& \multirow{2}{*}{type}
  & BERTscore & -- & -- & 0.798 & \cellcolor{cellYellow}{0.878} & -- & -- & \cellcolor{cellYellowLight}{0.803} \\
& & LLM-judge & -- & -- & 0.372 & \cellcolor{cellYellowLight}{0.580} & -- & -- & \cellcolor{cellYellow}{0.595} \\
\midrule

\multirow{10}{*}{\textbf{LiquidAD}} & \multirow{5}{*}{video}
  & Acc & 0.184 & 0.631 & 0.150 & 0.462 & \cellcolor{cellRed}0.903 & 0.091 & \cellcolor{cellRedLight}0.868 \\
& & P   & \cellcolor{cellRed}1.000 & \cellcolor{cellRedLight}0.966 & 0.886 & 0.896 & 0.912 & 0.000 & 0.910 \\
& & R   & 0.102 & 0.616 & 0.075 & 0.462 & \cellcolor{cellRed}0.988 & 0.000 & \cellcolor{cellRedLight}0.948 \\
& & F1  & 0.185 & 0.752 & 0.139 & 0.610 & \cellcolor{cellRed}0.949 & 0.000 & \cellcolor{cellRedLight}0.929 \\
& & AUC & 0.453 & \cellcolor{cellRedLight}0.651 & 0.489 & 0.465 & 0.365 & 0.534 & \cellcolor{cellRed}\textbf{0.692} \\
\cline{2-10}
& \multirow{5}{*}{frame}
  & Acc & 0.489 & 0.601 & \cellcolor{cellBlue}0.876 & 0.633 & 0.564 & \cellcolor{cellBlueLight}0.709 & 0.458 \\
& & P   & 0.395 & \cellcolor{cellBlue}0.800 & 0.000 & 0.253 & \cellcolor{cellBlueLight}0.602 & 0.000 & 0.431 \\
& & R   & 0.589 & 0.590 & 0.000 & 0.134 & \cellcolor{cellBlue}0.708 & 0.000 & \cellcolor{cellBlueLight}0.614 \\
& & F1  & 0.472 & \cellcolor{cellBlue}0.679 & 0.000 & 0.176 & \cellcolor{cellBlueLight}0.651 & 0.000 & 0.507 \\
& & AUC & 0.487 & \cellcolor{cellBlue}0.625 & 0.499 & 0.484 & 0.506 & 0.500 & \cellcolor{cellBlueLight}0.512 \\
\midrule

\multirow{10}{*}{\textbf{IPAD}} & \multirow{5}{*}{video}
  & Acc & 0.555 & 0.545 & 0.531 & \cellcolor{cellRedLight}0.607 & 0.582 & \cellcolor{cellRed}0.658 & 0.582 \\
& & P   & \cellcolor{cellRedLight}0.643 & \cellcolor{cellRed}0.836 & 0.581 & 0.375 & 0.590 & 0.238 & 0.588 \\
& & R   & 0.529 & 0.271 & 0.700 & 0.386 & \cellcolor{cellRedLight}0.929 & 0.089 & \cellcolor{cellRed}\textbf{0.954} \\
& & F1  & 0.581 & 0.409 & 0.635 & 0.388 & \cellcolor{cellRed}0.721 & 0.130 & \cellcolor{cellRedLight}\textbf{0.714} \\
& & AUC & 0.522 & \cellcolor{cellRedLight}0.539 & 0.498 & 0.519 & 0.417 & 0.519 & \cellcolor{cellRed}\textbf{0.565} \\
\cline{2-10}
& \multirow{5}{*}{frame}
  & Acc & 0.716 & 0.661 & 0.703 & \cellcolor{cellBlueLight}0.815 & 0.671 & \cellcolor{cellBlue}0.852 & 0.515 \\
& & P   & \cellcolor{cellBlue}0.410 & 0.309 & 0.114 & 0.273 & \cellcolor{cellBlueLight}0.331 & 0.000 & 0.291 \\
& & R   & 0.006 & \cellcolor{cellBlueLight}0.166 & 0.003 & 0.119 & 0.164 & 0.000 & \cellcolor{cellBlue}\textbf{0.477} \\
& & F1  & 0.012 & 0.216 & 0.006 & 0.165 & \cellcolor{cellBlueLight}0.220 & 0.000 & \cellcolor{cellBlue}\textbf{0.338} \\
& & AUC & 0.430 & 0.495 & 0.497 & \cellcolor{cellBlue}0.532 & 0.512 & 0.490 & \cellcolor{cellBlueLight}0.518 \\
\bottomrule
\multicolumn{10}{l}{\scriptsize ``$\dagger$'' = trained; ``$\ast$'' = training-free. Colored cells = best; lighter ones = 2nd best.}\\
\end{tabular}%
}
\end{table}

\begin{table}[t]\centering
\caption{\textbf{Video-level AUROC ($\uparrow$) on Phys-AD~\cite{li2025towards}
across 22 object categories.}
``$\dagger$''~=~methods requiring training data.
Per category, the best / 2nd-best across all methods is
\colorbox{cellRed}{\rule{0pt}{1.1ex}\rule{1.1ex}{0pt}} / \colorbox{cellRedLight}{\rule{0pt}{1.1ex}\rule{1.1ex}{0pt}};
our best results are additionally in \textbf{bold}.}
\label{tab:physad}
\setlength{\tabcolsep}{2pt}\renewcommand{\arraystretch}{1.3}
\resizebox{\columnwidth}{!}{%
\tiny
\begin{tabular}{@{}l cc cc cc c@{}}
\toprule
& \multicolumn{2}{c}{\textbf{Trad.}}
& \multicolumn{2}{c}{\textbf{Direct Prompting}}
& \multicolumn{2}{c}{\textbf{Workflow}}
& \multicolumn{1}{c}{\textbf{O-VAD (Ours)}} \\
\cmidrule(lr){2-3}\cmidrule(lr){4-5}\cmidrule(lr){6-7}\cmidrule(lr){8-8}
\textbf{Category}
& MNAD.p$^\dagger$ & S3R$^\dagger$
& Qwen3$^\ast$ & GPT-5$^\ast$
& URF$^\ast$ & VERA$^\dagger$
& \textbf{O-VAD}$^\ast$ \\
\midrule
{Ball} & \cellcolor{cellRedLight}0.589 & 0.155 & 0.528 & \cellcolor{cellRed}0.744 & 0.461 & 0.548 & 0.579 \\
{Button} & \cellcolor{cellRedLight}0.526 & 0.108 & 0.506 & 0.456 & \cellcolor{cellRed}0.717 & 0.469 & 0.393 \\
{Car} & \cellcolor{cellRedLight}0.599 & 0.527 & 0.492 & \cellcolor{cellRed}0.617 & 0.497 & 0.383 & 0.575 \\
{Caster Wheel} & 0.468 & \cellcolor{cellRed}0.830 & 0.533 & 0.543 & 0.410 & \cellcolor{cellRedLight}0.609 & 0.301 \\
{Clip} & \cellcolor{cellRed}0.628 & \cellcolor{cellRedLight}0.537 & 0.442 & 0.518 & 0.275 & 0.180 & 0.423 \\
{Clock} & 0.513 & 0.529 & 0.581 & \cellcolor{cellRed}0.766 & 0.483 & 0.500 & \cellcolor{cellRedLight}0.669 \\
{Fan} & 0.409 & \cellcolor{cellRedLight}0.602 & 0.566 & 0.472 & 0.523 & 0.389 & \cellcolor{cellRed}\textbf{0.669} \\
{Gear} & 0.418 & 0.358 & \cellcolor{cellRedLight}0.644 & 0.397 & 0.451 & 0.512 & \cellcolor{cellRed}\textbf{0.879} \\
{Hinge} & 0.497 & \cellcolor{cellRed}0.716 & \cellcolor{cellRedLight}0.589 & 0.516 & 0.544 & 0.041 & 0.526 \\
{Liquid} & 0.453 & 0.733 & \cellcolor{cellRed}0.833 & 0.512 & 0.691 & \cellcolor{cellRedLight}0.827 & 0.776 \\
{Lock} & 0.442 & 0.323 & 0.388 & 0.231 & 0.029 & \cellcolor{cellRedLight}0.557 & \cellcolor{cellRed}\textbf{0.780} \\
{Magnet} & 0.512 & \cellcolor{cellRedLight}0.724 & 0.592 & 0.364 & 0.518 & \cellcolor{cellRed}0.892 & 0.576 \\
{Roll. Bear.} & 0.437 & 0.016 & 0.418 & 0.451 & 0.304 & \cellcolor{cellRedLight}0.500 & \cellcolor{cellRed}\textbf{0.650} \\
{Rub. Band} & 0.561 & 0.463 & \cellcolor{cellRedLight}0.767 & 0.279 & \cellcolor{cellRed}0.783 & 0.276 & 0.721 \\
{Screw} & \cellcolor{cellRedLight}0.638 & 0.620 & 0.450 & 0.441 & 0.298 & 0.400 & \cellcolor{cellRed}\textbf{0.704} \\
{Servo} & 0.479 & 0.452 & \cellcolor{cellRedLight}0.492 & 0.330 & 0.342 & 0.341 & \cellcolor{cellRed}\textbf{0.495} \\
{Slide} & \cellcolor{cellRed}0.534 & \cellcolor{cellRedLight}0.533 & 0.529 & 0.528 & 0.345 & 0.225 & 0.415 \\
{Sph. Bear.} & 0.547 & \cellcolor{cellRed}0.894 & 0.483 & 0.440 & 0.629 & 0.114 & \cellcolor{cellRedLight}0.641 \\
{Sticky Roller} & 0.517 & \cellcolor{cellRedLight}0.778 & 0.533 & 0.523 & 0.713 & 0.474 & \cellcolor{cellRed}\textbf{0.811} \\
{Toothpaste} & 0.257 & 0.535 & 0.578 & \cellcolor{cellRed}0.832 & 0.528 & 0.579 & \cellcolor{cellRedLight}0.689 \\
{U Disk} & 0.513 & \cellcolor{cellRedLight}0.585 & 0.578 & 0.437 & 0.404 & 0.311 & \cellcolor{cellRed}\textbf{0.606} \\
{Zipper} & 0.262 & \cellcolor{cellRed}0.791 & \cellcolor{cellRedLight}0.517 & 0.477 & 0.273 & 0.224 & 0.388 \\
\midrule
\textbf{Avg.} & 0.495 & \cellcolor{cellRedLight}0.555 & 0.513 & 0.503 & 0.426 & 0.456 & \cellcolor{cellRed}\textbf{0.584} \\
\bottomrule
\end{tabular}}
\end{table}
\subsection{Quantitative Results}
\paragraph{\textbf{Overall comparison.}}

Table~\ref{tab:main} summarizes aggregate Acc/P/R/F1/AUROC across the three benchmarks. Despite using no training data, domain knowledge, or predefined taxonomies, O-VAD attains the best average video-level AUROC among training-free methods on Phys-AD (0.584), beating the direct-prompting baselines Qwen3-VL-32B (0.513) and GPT-5 (0.503) and the agentic workflows URF-ZS-HVAA (0.426) and VERA (0.456), and even exceeding both \emph{trained} baselines MNAD.p (0.495) and S3R (0.555). It also yields the most semantically faithful descriptions, with the best type-level BERTScore (0.803) among training-free methods. The same pattern holds on LiquidAD and IPAD. This shows the bottleneck in VLM-based IVAD lies not in language reasoning but in object-level evidence construction: O-VAD's ``ground → track → reason'' pipeline avoids the false-negative collapse seen in caption-level baselines.

\paragraph{\textbf{Per-dataset analysis.}}
Table~\ref{tab:physad} reports per-category video-level AUROC on \textbf{Phys-AD} across all 22 object categories. O-VAD claims the best or second-best AUROC on 16 of 22 categories, with the largest gains on categories involving complex multi-step interactions (\eg, Gear 0.879, Lock 0.780, Screw 0.704) where temporal state tracking is most critical, and on those with subtle surface-level anomalies (\eg, Sticky Roller 0.811, U Disk 0.606) where object-centric analysis reveals details invisible to frame-level approaches. The other two benchmarks stress complementary capabilities. \textbf{LiquidAD} requires per-instance precision in a cluttered scene with up to eight simultaneous pipettes: zero-shot VLMs collapse to near-zero recall and a strong bias toward ``normal,'' whereas O-VAD tracks per-pipette state trajectories to attain the best training-free video-level AUROC (0.692) and the strongest frame-level recall and F1. \textbf{IPAD} shifts from physics-grounded reasoning to reference-based deviation across 16 industrial scenarios, where O-VAD again achieves the highest average video-level (0.565) and frame-level (0.518) AUROC and ranks first on 12 of 16 scenarios at the video level. Together these results show that object-centric state tracking generalizes across multi-instance laboratory scenes and periodic industrial processes alike. Per-category breakdowns reporting all five metrics (Acc/P/R/F1/AUROC) for every category and scenario on all three datasets are provided in the supplementary material (\S\ref{sec:supp_physad} and \S\ref{sec:res-IPAD}).

\begin{figure}[!t]
    \centering
    \includegraphics[width=1\linewidth]{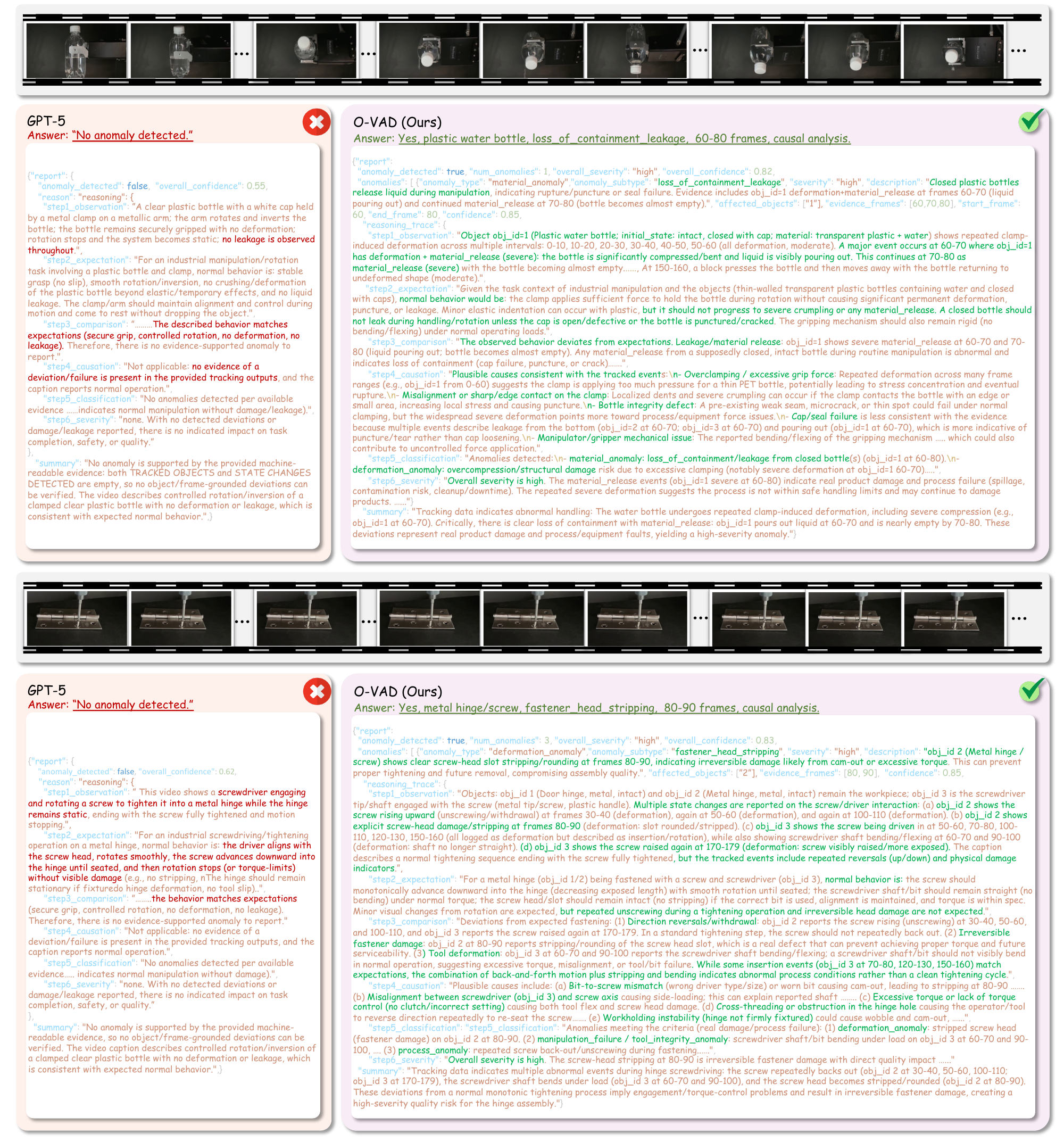}
    \caption{Qualitative results. Each row shows a different industrial manipulation task---plastic bottle rotation (top), hinge screw fastening (bottom)---with GPT-5's output (left) and O-VAD’s output (right).}
    \label{fig:qualitative}
\end{figure}

\subsection{Qualitative Results}

To illustrate how object-centric evidence shapes reasoning quality, we compare O-VAD against GPT-5~\cite{singh2025openai}---prompted with the same cascaded chain-of-thought but \emph{without} object grounding or state tracking---on three representative Phys-AD scenarios (Figure~\ref{fig:qualitative}). Additional reasoning traces, failure-case analyses, and human and LLM-as-judge studies are provided in the appendix (\S\ref{sec:supp_qualitative}).

\paragraph{\textbf{Case Analysis.}}
Figure~\ref{fig:qualitative} visualizes full reasoning traces on representative Phys-AD examples. In Case~1 (plastic bottle rotation), a clamped bottle leaks severely at frames~60--80. GPT-5 reports ``no leakage,'' while O-VAD tracks progressive deformation (frames~0--60), detects material release at frame~60, and classifies a high-severity \textit{loss-of-containment} anomaly. In Case~2 (hinge screw fastening), three defects co-occur: repeated screw back-out (frames~30--110), head-slot stripping (frames~80--90), and screwdriver shaft bending (frames~60--100). GPT-5 concludes normal tightening; O-VAD captures all three streams and attributes them to torque-control or bit-mismatch failure. 
Across all cases, GPT-5 defaults to ``no anomaly'' without object state evidence, whereas O-VAD produces grounded reports with localized objects, frame ranges, severity, and causal explanations.

\paragraph{\textbf{Why object state trajectories matter.}}
Several patterns emerge. First, GPT-5's failures stem not from poor language reasoning but from impoverished evidence: without object-wise state trajectories, it receives caption-level inputs and judges only whether ``the behavior matches expectations,'' confirming that the bottleneck lies in \textit{fine-grained evidence construction}. Second, O-VAD's multi-object tracking enables \textit{cross-object diagnostic reasoning} (e.g., correlating shaft bending with screw stripping in Case~2), which is unavailable to frame-level approaches. Third, open-ended state-change events provide \emph{spatiotemporally grounded evidence} anchoring every claim to specific objects and frame ranges, making reports directly actionable for root-cause diagnosis.

\paragraph{\textbf{Open-ended anomaly explanation.}}
O-VAD can detect free-form anomalies grounded in tracked object states rather than a fixed taxonomy. In Case~1, it attributes the leakage to progressive stress from repeated clamp-induced deformation, citing the spatial relationship between gripper and breach. In Case~2, it synthesizes three independent anomaly streams (direction reversals, head stripping, shaft bending) into a coherent torque-control failure hypothesis. Such structured causal reasoning goes beyond binary labels to provide the diagnostic detail operators need for corrective action.

\subsection{Ablation Studies}

We validate the contribution of each component in O-VAD by systematically removing one module at a time on four representative task subsets from Phys-AD (sticky roller, liquid, screw, and rubber band) that span diverse anomaly characteristics. Results are reported in Table~\ref{tab:ablation}.

\paragraph{\textbf{State Tracking is indispensable.}} Removing state tracking causes catastrophic failure: precision, recall, and F1 drop to zero on three of four categories (sticky roller, liquid, screw) as the system collapses to predicting all samples normal, and recall falls to 0.033 (F1 0.065) on rubber band. Without an explicit mechanism to model state transitions across frames, the system cannot distinguish anomalous deviations from normal progressions. BERT scores stay paradoxically high (0.867–0.947) as the model emits fluent but content-agnostic ``normal'' descriptions, underscoring that text-quality metrics alone are insufficient for evaluating anomaly detection.

\paragraph{\textbf{Video-level captioning provides a global semantic prior.}}  Removing video-level captioning has a category-dependent effect, revealing its role as a scene-level contextual anchor. For sticky roller and rubber band, where anomalies are holistic behavioral deviations (e.g., irregular rolling, band detachment), the caption supplies scene semantics that guide reasoning, and its removal causes AUC drops of 0.247 and 0.131. For liquid, classification metrics are unchanged and AUC decreases marginally (0.034), so state tracking alone suffices for anomalies such as overflow. On screw, removing the caption slightly improves F1 (0.700$\to$0.733): screw anomalies are localized and instantaneous (e.g., misalignment, incomplete tightening), so a coarse video-level summary introduces task-irrelevant context---motivating future multi-granularity captioning adapted to the anomaly's spatio-temporal scale.

\paragraph{\textbf{Structured cognitive reasoning improves calibration.}} Replacing our cascaded six-step reasoning module (Sec.~\ref{sec:cot}) with a generic ``think step by step'' prompt still yields reasonable results, but the structured pipeline consistently improves AUC across all four categories (by 0.038--0.091) and raises BERT scores. Decomposing reasoning into process understanding, observation, expectation, comparison, causation, and classification thus yields better-calibrated confidence and more faithful explanations grounded in tracking evidence.

\paragraph{\textbf{Post-verification serves as an effective safeguard.}} 
Removing visual verification (Sec.~\ref{sec:verify}) preserves most hard classification metrics but consistently degrades AUC (up to 0.178 on sticky roller) and BERT scores, indicating it mainly benefits borderline cases rather than flipping high-confidence predictions. Through multiplicative confidence gating (Eq.~\ref{eq:conf_gate}), the verifier re-examines evidence frames against the caption to downweight spurious detections, sharpening the separation between true anomalies and ambiguous cases and improving explanation faithfulness.

\begin{table*}[t]
\centering
\caption{\textbf{Ablation study on Phys-AD subset.}
Per-category Acc/P/R/F1/AUC/BERT scores across ablation variants.
The result per category with full design is \textbf{bold}; \colorbox{degraded}{orange} marks degradation from the full model.}
\label{tab:ablation}
\setlength{\tabcolsep}{4pt}
\small
\resizebox{\textwidth}{!}{%
\begin{tabular}{ll cccccc l cccccc}
\toprule
 & & \multicolumn{6}{c}{\textbf{sticky roller}} & & \multicolumn{6}{c}{\textbf{liquid}} \\
\cmidrule(lr){3-8}\cmidrule(lr){10-15}
\textbf{Variant} & & Acc & P & R & F1 & AUC & BERT & & Acc & P & R & F1 & AUC & BERT \\
\midrule
\textbf{\textit{O-VAD (full)}}
  & & \textbf{0.689} & \textbf{0.691} & \textbf{0.967} & \textbf{0.806} & \textbf{0.811} & \textbf{0.689}
  & & \textbf{0.667} & \textbf{0.667} & \textbf{1.000} & \textbf{0.800} & \textbf{0.776} & \textbf{0.666} \\
(a) \textit{w/o Caption}
  & & \cc 0.644 & \cc 0.659 & 0.967 & \cc 0.784 & \cc 0.564 & \cc 0.645
  & & 0.667 & 0.667 & 1.000 & 0.800 & \cc 0.742 & \cc 0.650 \\
(b) \textit{w/o State Track.}
  & & \cc 0.311 & \cc 0.000 & \cc 0.000 & \cc 0.000 & \cc 0.484 & 0.915
  & & \cc 0.333 & \cc 0.000 & \cc 0.000 & \cc 0.000 & \cc 0.667 & 0.867 \\
(c) \textit{w/o CoT Reas.}
  & & \cc 0.667 & \cc 0.667 & 1.000 & \cc 0.800 & \cc 0.720 & \cc 0.648
  & & 0.667 & 0.667 & 1.000 & 0.800 & \cc 0.738 & \cc 0.632 \\
(d) \textit{w/o Post-verifier}
  & & \cc 0.667 & \cc 0.667 & 1.000 & \cc 0.800 & \cc 0.633 & \cc 0.645
  & & 0.667 & 0.667 & 1.000 & 0.800 & \cc 0.718 & \cc 0.648 \\
\midrule
 & & \multicolumn{6}{c}{\textbf{screw}} & & \multicolumn{6}{c}{\textbf{rubber band}} \\
\cmidrule(lr){3-8}\cmidrule(lr){10-15}
\textbf{Variant} & & Acc & P & R & F1 & AUC & BERT & & Acc & P & R & F1 & AUC & BERT \\
\midrule
\textbf{\textit{O-VAD (full)}}
  & & \textbf{0.600} & \textbf{0.700} & \textbf{0.700} & \textbf{0.700} & \textbf{0.704} & \textbf{0.784}
  & & \textbf{0.633} & \textbf{0.611} & \textbf{0.733} & \textbf{0.667} & \textbf{0.721} & \textbf{0.784} \\
(a) \textit{w/o Caption}
  & & 0.644 & 0.733 & 0.733 & 0.733 & 0.731 & 0.788
  & & \cc 0.500 & \cc 0.500 & \cc 0.667 & \cc 0.571 & \cc 0.590 & \cc 0.776 \\
(b) \textit{w/o State Track.}
  & & \cc 0.333 & \cc 0.000 & \cc 0.000 & \cc 0.000 & \cc 0.400 & 0.936
  & & \cc 0.517 & 1.000 & \cc 0.033 & \cc 0.065 & \cc 0.587 & 0.947 \\
(c) \textit{w/o CoT Reas.}
  & & 0.644 & 0.733 & 0.733 & 0.733 & \cc 0.664 & \cc 0.781
  & & \cc 0.617 & \cc 0.595 & 0.733 & \cc 0.657 & \cc 0.681 & \cc 0.766 \\
(d) \textit{w/o Post-verifier}
  & & 0.622 & 0.710 & 0.733 & 0.721 & \cc 0.622 & \cc 0.758
  & & \cc 0.583 & \cc 0.571 & \cc 0.667 & \cc 0.615 & \cc 0.714 & 0.799 \\
\bottomrule
\end{tabular}
}
\end{table*}

\section{Conclusion}

Without labor-intensive annotations, our agentic framework delivers state-of-the-art performance on any level anomaly detection, surpassing frontier VLMs on quantitative and qualitative evaluations. This demonstrates a new paradigm for industrial video anomaly detection that bridges the gap between low-level pattern recognition and high-level cognitive reasoning, paving the way for more reliable, explainable, and generalizable industrial inspection systems.

\paragraph{\textbf{Limitations.}}

First, the multi-stage agentic pipeline incurs substantially higher latency than single-forward-pass VAD models, scaling with the number of tracked objects per video.
Second, O-VAD relies on the VLM's internalized commonsense rather than domain-specific expert priors, so its performance declines on anomalies defined by invisible physical properties or precise specifications 
that cannot be inferred from visual appearance alone---few-shot in-context learning with specification examples is a promising direction to bridge this gap.
Third, static or perception-ambiguous defects (\eg, stuck buttons, degaussed magnets) that produce visually plausible behavior remain a fundamental boundary of perception-driven reasoning, motivating future integration of lightweight domain priors and process specification references.

\section*{Acknowledgements}

This work was supported in part by U.S. NSF grants DBI-2238093, DBI-2422619, IIS-2211597, and MCB-2205148. Yizhou Zhao was supported in part by the SoftBank Group–ARM Fellowship.

\bibliographystyle{splncs04}
\bibliography{main}

\begin{thebibliography}{10}
\providecommand{\url}[1]{\texttt{#1}}
\providecommand{\urlprefix}{URL }
\providecommand{\doi}[1]{https://doi.org/#1}

\bibitem{bai2025qwen3}
Bai, S., Cai, Y., Chen, R., Chen, K., Chen, X., Cheng, Z., Deng, L., Ding, W., Gao, C., Ge, C., et~al.: Qwen3-vl technical report. arXiv preprint arXiv:2511.21631  (2025)

\bibitem{carion2025sam}
Carion, N., Gustafson, L., Hu, Y.T., Debnath, S., Hu, R., Suris, D., Ryali, C., Alwala, K.V., Khedr, H., Huang, A., et~al.: Sam 3: Segment anything with concepts. arXiv preprint arXiv:2511.16719  (2025)

\bibitem{chen2025m}
Chen, Z., Li, J., Liang, J., Tan, L., Guo, Y., Lu, C., Li, Y.L.: M\^{} 3-vos: Multi-phase, multi-transition, and multi-scenery video object segmentation. In: Proceedings of the Computer Vision and Pattern Recognition Conference. pp. 29193--29202 (2025)

\bibitem{cheng2024putting}
Cheng, H.K., Oh, S.W., Price, B., Lee, J.Y., Schwing, A.: Putting the object back into video object segmentation. In: Proceedings of the IEEE/CVF Conference on Computer Vision and Pattern Recognition. pp. 3151--3161 (2024)

\bibitem{dabouei2025deep}
Dabouei, A., Shibu, J.P., Dalal, V., Cao, C., MacWilliams, A., Kangas, J., Xu, M.: Deep video anomaly detection in automated laboratory setting. Expert Systems with Applications  \textbf{271},  126581 (2025)

\bibitem{defard2021padim}
Defard, T., Setkov, A., Loesch, A., Audigier, R.: Padim: a patch distribution modeling framework for anomaly detection and localization. In: International conference on pattern recognition. pp. 475--489. Springer (2021)

\bibitem{devlin2019bert}
Devlin, J., Chang, M.W., Lee, K., Toutanova, K.: Bert: Pre-training of deep bidirectional transformers for language understanding. In: Proceedings of the 2019 conference of the North American chapter of the association for computational linguistics: human language technologies, volume 1 (long and short papers). pp. 4171--4186 (2019)

\bibitem{fan2024revitalizing}
Fan, L., Huang, J., Di, D., Su, A., Pagnucco, M., Song, Y.: Revitalizing reconstruction models for multi-class anomaly detection via class-aware contrastive learning. arXiv preprint arXiv:2412.04769  (2024)

\bibitem{fang2023fastrecon}
Fang, Z., Wang, X., Li, H., Liu, J., Hu, Q., Xiao, J.: Fastrecon: Few-shot industrial anomaly detection via fast feature reconstruction. In: Proceedings of the IEEE/CVF International Conference on Computer Vision. pp. 17481--17490 (2023)

\bibitem{gu2024anomalygpt}
Gu, Z., Zhu, B., Zhu, G., Chen, Y., Tang, M., Wang, J.: Anomalygpt: Detecting industrial anomalies using large vision-language models. In: Proceedings of the AAAI conference on artificial intelligence. vol.~38, pp. 1932--1940 (2024)

\bibitem{guo2025dinomaly}
Guo, J., Lu, S., Zhang, W., Chen, F., Li, H., Liao, H.: Dinomaly: The less is more philosophy in multi-class unsupervised anomaly detection. In: Proceedings of the Computer Vision and Pattern Recognition Conference. pp. 20405--20415 (2025)

\bibitem{he2024mambaad}
He, H., Bai, Y., Zhang, J., He, Q., Chen, H., Gan, Z., Wang, C., Li, X., Tian, G., Xie, L.: Mambaad: Exploring state space models for multi-class unsupervised anomaly detection. Advances in Neural Information Processing Systems  \textbf{37},  71162--71187 (2024)

\bibitem{huang2022hierarchical}
Huang, C., Liu, Y., Zhang, Z., Liu, C., Wen, J., Xu, Y., Wang, Y.: Hierarchical graph embedded pose regularity learning via spatio-temporal transformer for abnormal behavior detection. In: Proceedings of the 30th ACM international conference on multimedia. pp. 307--315 (2022)

\bibitem{huang2025vad}
Huang, C., Wang, B., Wen, J., Liu, C., Wang, W., Shen, L., Cao, X.: Vad-r1: Towards video anomaly reasoning via perception-to-cognition chain-of-thought. arXiv preprint arXiv:2505.19877  (2025)

\bibitem{hyun2024reconpatch}
Hyun, J., Kim, S., Jeon, G., Kim, S.H., Bae, K., Kang, B.J.: Reconpatch: Contrastive patch representation learning for industrial anomaly detection. In: Proceedings of the IEEE/CVF winter conference on applications of computer vision. pp. 2052--2061 (2024)

\bibitem{jiang2024mmad}
Jiang, X., Li, J., Deng, H., Liu, Y., Gao, B.B., Zhou, Y., Li, J., Wang, C., Zheng, F.: Mmad: A comprehensive benchmark for multimodal large language models in industrial anomaly detection. arXiv preprint arXiv:2410.09453  (2024)

\bibitem{jiang2022softpatch}
Jiang, X., Liu, J., Wang, J., Nie, Q., Wu, K., Liu, Y., Wang, C., Zheng, F.: Softpatch: Unsupervised anomaly detection with noisy data. Advances in Neural Information Processing Systems  \textbf{35},  15433--15445 (2022)

\bibitem{kangjudo}
Kang, H., Lee, W., Kim, J., Park, H.: Judo: A juxtaposed domain-oriented multimodal reasoner for industrial anomaly qa. In: The Fourteenth International Conference on Learning Representations (2026)

\bibitem{kirillov2023segment}
Kirillov, A., Mintun, E., Ravi, N., Mao, H., Rolland, C., Gustafson, L., Xiao, T., Whitehead, S., Berg, A.C., Lo, W.Y., et~al.: Segment anything. In: Proceedings of the IEEE/CVF international conference on computer vision. pp. 4015--4026 (2023)

\bibitem{li2022scale}
Li, G., Cai, G., Zeng, X., Zhao, R.: Scale-aware spatio-temporal relation learning for video anomaly detection. In: European Conference on Computer Vision. pp. 333--350. Springer (2022)

\bibitem{li2025video}
Li, J., Dang, L., Xiao, Q., Shang, S., Cheng, J., Wu, H., Hao, Y., Wu, Q.: Video anomaly detection with semantics-aware information bottleneck. arXiv preprint arXiv:2506.02535  (2025)

\bibitem{li2026vadtree}
Li, W., Xu, Y., Rao, Y., Wang, Z., Deng, S.: Vadtree: Explainable training-free video anomaly detection via hierarchical granularity-aware tree. Advances in Neural Information Processing Systems  \textbf{38},  148372--148404 (2026)

\bibitem{li2025towards}
Li, W., Gu, Y., Chen, X., Xu, X., Hu, M., Huang, X., Wu, Y.: Towards visual discrimination and reasoning of real-world physical dynamics: Physics-grounded anomaly detection. In: Proceedings of the Computer Vision and Pattern Recognition Conference. pp. 30409--30419 (2025)

\bibitem{lin2025unified}
Lin, D., Qu, M., Han, K., Jiao, J., Jin, X., Wei, Y.: A unified reasoning framework for holistic zero-shot video anomaly analysis. arXiv preprint arXiv:2511.00962  (2025)

\bibitem{lin2025vlm}
Lin, S., Wang, C., Ding, X., Wang, Y., Du, B., Song, L., Wang, C., Liu, H.: A vlm-based method for visual anomaly detection in robotic scientific laboratories. In: 2025 International Conference on Advanced Robotics and Mechatronics (ICARM). pp. 34--39. IEEE (2025)

\bibitem{liu2024ipad}
Liu, J., Yan, Y., Li, J., Zhao, W., Chu, P., Sheng, X., Liu, Y., Yang, X.: Ipad: Industrial process anomaly detection dataset. IEEE Transactions on Circuits and Systems for Video Technology  \textbf{35}(1),  380--393 (2024)

\bibitem{liu2019exploring}
Liu, K., Ma, H.: Exploring background-bias for anomaly detection in surveillance videos. In: Proceedings of the 27th ACM International Conference on Multimedia. pp. 1490--1499 (2019)

\bibitem{park2020learning}
Park, H., Noh, J., Ham, B.: Learning memory-guided normality for anomaly detection. In: Proceedings of the IEEE/CVF conference on computer vision and pattern recognition. pp. 14372--14381 (2020)

\bibitem{qi2022high}
Qi, L., Kuen, J., Guo, W., Shen, T., Gu, J., Jia, J., Lin, Z., Yang, M.H.: High-quality entity segmentation. arXiv preprint arXiv:2211.05776  (2022)

\bibitem{radford2021clip}
Radford, A., Kim, J.W., Hallacy, C., Ramesh, A., Goh, G., Agarwal, S., Sastry, G., Askell, A., Mishkin, P., Clark, J., et~al.: Learning transferable visual models from natural language supervision. In: International conference on machine learning. pp. 8748--8763. PmLR (2021)

\bibitem{ravi2024sam2}
Ravi, N., Gabeur, V., Hu, Y.T., Hu, R., Ryali, C., Ma, T., Khedr, H., R{\"a}dle, R., Rolland, C., Gustafson, L., et~al.: Sam 2: Segment anything in images and videos. arXiv preprint arXiv:2408.00714  (2024)

\bibitem{roth2022towards}
Roth, K., Pemula, L., Zepeda, J., Sch{\"o}lkopf, B., Brox, T., Gehler, P.: Towards total recall in industrial anomaly detection. In: Proceedings of the IEEE/CVF conference on computer vision and pattern recognition. pp. 14318--14328 (2022)

\bibitem{singh2025openai}
Singh, A., Fry, A., Perelman, A., Tart, A., Ganesh, A., El-Kishky, A., McLaughlin, A., Low, A., Ostrow, A., Ananthram, A., et~al.: Openai gpt-5 system card. arXiv preprint arXiv:2601.03267  (2025)

\bibitem{sun2023long}
Sun, S., Gong, X.: Long-short temporal co-teaching for weakly supervised video anomaly detection. In: 2023 IEEE International Conference on Multimedia and Expo (ICME). pp. 2711--2716. IEEE (2023)

\bibitem{sun2025tracking}
Sun, Y., Yang, X., Sun, J.J., Hariharan, B.: Tracking and understanding object transformations. arXiv preprint arXiv:2511.04678  (2025)

\bibitem{tokmakov2023breaking}
Tokmakov, P., Li, J., Gaidon, A.: Breaking the" object" in video object segmentation. In: Proceedings of the IEEE/CVF Conference on Computer Vision and Pattern Recognition. pp. 22836--22845 (2023)

\bibitem{wu2022self}
Wu, J.C., Hsieh, H.Y., Chen, D.J., Fuh, C.S., Liu, T.L.: Self-supervised sparse representation for video anomaly detection. In: European Conference on Computer Vision. pp. 729--745. Springer (2022)

\bibitem{wu2021weakly}
Wu, J., Zhang, W., Li, G., Wu, W., Tan, X., Li, Y., Ding, E., Lin, L.: Weakly-supervised spatio-temporal anomaly detection in surveillance video. arXiv preprint arXiv:2108.03825  (2021)

\bibitem{wu2024weakly}
Wu, P., Zhou, X., Pang, G., Yang, Z., Yan, Q., Wang, P., Zhang, Y.: Weakly supervised video anomaly detection and localization with spatio-temporal prompts. In: Proceedings of the 32nd ACM International Conference on Multimedia. pp. 9301--9310 (2024)

\bibitem{ye2025vera}
Ye, M., Liu, W., He, P.: Vera: Explainable video anomaly detection via verbalized learning of vision-language models. In: Proceedings of the Computer Vision and Pattern Recognition Conference. pp. 8679--8688 (2025)

\bibitem{yu2023video}
Yu, J., Li, X., Zhao, X., Zhang, H., Wang, Y.X.: Video state-changing object segmentation. In: Proceedings of the IEEE/CVF international conference on computer vision. pp. 20439--20448 (2023)

\bibitem{zavrtanik2021draem}
Zavrtanik, V., Kristan, M., Sko{\v{c}}aj, D.: Draem-a discriminatively trained reconstruction embedding for surface anomaly detection. In: Proceedings of the IEEE/CVF international conference on computer vision. pp. 8330--8339 (2021)

\bibitem{zhangdiffusionad}
Zhang, H., Wang, Z., Wu, Z., Jiang, Y.: Diffusionad: norm-guided one-step denoising diffusion for anomaly detection (2023)

\bibitem{zhang2024realnet}
Zhang, X., Xu, M., Zhou, X.: Realnet: A feature selection network with realistic synthetic anomaly for anomaly detection. In: Proceedings of the IEEE/CVF conference on computer vision and pattern recognition. pp. 16699--16708 (2024)

\bibitem{zhao2025omniad}
Zhao, S., Lin, Y., Han, L., Zhao, Y., Wei, Y.: Omniad: Detect and understand industrial anomaly via multimodal reasoning. arXiv preprint arXiv:2505.22039  (2025)

\bibitem{zheng2026iad}
Zheng, H., Lin, T., Wang, W., Wang, Z., Zhang, W., Zhu, J., Shao, F.: Iad-unify: A region-grounded unified model for industrial anomaly segmentation, understanding, and generation. arXiv preprint arXiv:2604.12440  (2026)

\bibitem{zheng2023judging}
Zheng, L., Chiang, W.L., Sheng, Y., Zhuang, S., Wu, Z., Zhuang, Y., Lin, Z., Li, Z., Li, D., Xing, E., et~al.: Judging llm-as-a-judge with mt-bench and chatbot arena. Advances in neural information processing systems  \textbf{36},  46595--46623 (2023)

\bibitem{zhu2025vau}
Zhu, L., Chen, Q., Shen, X., Cun, X.: Vau-r1: Advancing video anomaly understanding via reinforcement fine-tuning. arXiv preprint arXiv:2505.23504  (2025)

\bibitem{zou2026unlocking}
Zou, S., Tian, X., Wesemann, L., Waschkowski, F., Yang, Z., Zhang, J.: Unlocking vision-language models for video anomaly detection via fine-grained prompting. In: Proceedings of the IEEE/CVF Winter Conference on Applications of Computer Vision. pp. 4223--4233 (2026)

\end{thebibliography}

\clearpage

\appendix
\section*{Appendix}

\section{More Details about Experiment Settings}

\subsection{Details of Datasets.}
\label{sec:supp_datasets}

We evaluate O-VAD on Phys-AD~\cite{li2025towards} and
LiquidAD~\cite{dabouei2025deep}, and additionally discuss IPAD~\cite{liu2024ipad} to contextualize our design choices. All three target industrial video anomaly detection but differ in scale, interaction
modality, and anomaly characteristics. Table~\ref{tab:dataset-comparison} summarizes their key properties.
 
\begin{table*}[h]
\centering
\caption{\textbf{Comparison of industrial video anomaly detection datasets.}}
\label{tab:dataset-comparison}
\vspace{-2pt}
\resizebox{\textwidth}{!}{
\setlength{\tabcolsep}{8pt}
\begin{tabular}{lccccccccc}
\toprule
\textbf{Dataset} & \textbf{Domain} & \textbf{Data Type} & \textbf{\#Videos} & \textbf{\#Obj. Categories} & \textbf{\#Anomaly Types} & \textbf{FPS} & 
\textbf{Annotation Granularity} & \textbf{Periodicity} & \textbf{Text Labels} \\
\midrule
Phys-AD~\cite{li2025towards} & Robot manipulation & Real & 6,434 & 22 (49 obj.) & 47 & 60 
& Video, Object & \ding{55} & \ding{51} \\
LiquidAD$^\ddag$~\cite{dabouei2025deep} & Laboratory automation & Real & 2,251 & 1 & 8 (pipette-level) & 30 & 
Video, Frame, Object & \ding{55} & \ding{55} \\
IPAD~\cite{liu2024ipad} & Factory equipment & Syn.+Real & $\sim$2,000 & 16 & 39 & 25 & 
Video, Frame & \ding{51} & \ding{55} \\
\bottomrule
\multicolumn{10}{l}{\footnotesize $^\ddag$ Due to the difficulty of reproducing the frame- and pipette-level data splits of LiquidAD, we use only video-level annotations in this work.}
\end{tabular}}
\end{table*}
 
\paragraph{(i) Phys-AD~\cite{li2025towards}} is a large-scale, physics-grounded video dataset for industrial anomaly detection. It is a real-world dataset focused on robotic arm manipulation, comprising 6,434 videos that cover 47 anomaly types across 22 object categories spanning metals, plastics, fluids, and articulated assemblies. Each object is paired with category-specific mechanical interactions, including pressing, rotating, and stretching, performed by a UR5 robotic arm or servo motors. Ground-truth annotations are provided at both the video level and object level, covering normal and anomalous conditions. Detecting anomalies requires reasoning about how an object's physical state evolves under mechanical interaction, \eg~whether a rubber band develops micro-cracks or a screw backs out during tightening. The dataset further distinguishes persistent anomalies (visible throughout the entire sequence) from intermittent ones (emerging only at specific temporal points), demanding fine-grained temporal understanding. These properties make Phys-AD an ideal testbed for evaluating O-VAD's object-centric state tracking capability.

\paragraph{(ii) LiquidAD~\cite{dabouei2025deep}} targets anomaly detection in automated laboratory liquid transfer. It comprises 2,251 videos (1,792 training / 459 test) at 30~FPS, each capturing a complete transfer process in which 8 parallel pipettes dispense liquid into 8 test tubes. Anomalies include pipette clogging, incomplete dispensing, and volume deviations, with ground-truth annotations provided at the video, frame, and individual pipette levels. Although LiquidAD covers a single procedural scenario, it presents substantial intra-scene complexity: up to 8 simultaneous dispensing events occur per frame, yet only one or two pipettes may be anomalous at any given time. Visual differences between normal and anomalous transfers are often subtle, such as a marginally reduced liquid column or a momentary dispensing hesitation, requiring temporally precise and spatially localized detection. O-VAD addresses this challenge by constructing per-pipette state trajectories that identify which pipette deviates from expected behavior and at which frames the deviation occurs.

\paragraph{(iii) IPAD~\cite{liu2024ipad}} is a video anomaly detection dataset designed for industrial manufacturing processes, comprising 597,979 frames across around 2,000 video clips. The dataset covers a diverse set of industrial devices, including conveyor belts, lifters, cutters, and grippers, encompassing both synthetic data (12 devices) and real on-site data (4 devices). A distinctive characteristic of IPAD is its explicit annotation of periodicity, a hallmark of industrial equipment in which actions repeat in fixed cycles. Anomaly types span appearance changes, positional deviations, motion irregularities, and logical errors, amounting to 39 distinct anomaly types in total. Importantly, environmental variations such as lighting changes and camera jitter are labeled as normal conditions, with the explicit intent of distinguishing genuine anomalies from benign scene perturbations.
We evaluate O-VAD on IPAD (see \S\ref{sec:res-IPAD}), though we note that this dataset emphasizes periodic structure and cross-domain transfer rather than physics-grounded physical interactions. Unlike Phys-AD~\cite{li2025towards} and LiquidAD~\cite{dabouei2025deep}, where anomalies are grounded in physical commonsense and mechanical interactions, IPAD defines anomalies primarily through deviations in object appearance (color, shape, size) and spatial location relative to reference normal samples. To adapt O-VAD to this reference-based evaluation protocol, we incorporate reference normal scenes as part of the object-level common sense representation, while continuing to exclude known anomaly types during evaluation (see \S\ref{sec:res-IPAD} for details).

\paragraph{Summary.} These three datasets represent a progression of complementary challenges: Phys-AD requires physics-grounded reasoning over diverse mechanical interactions, LiquidAD demands per-object temporal localization within cluttered and visually homogeneous scenes, and IPAD emphasizes periodic temporal structure with synthetic-to-real generalization. O-VAD's three-stage pipeline, consisting of object discovery, state tracking, and chain-of-thought reasoning, is principally designed to address the first two settings, where anomalies are rooted in physical commonsense and object-level state evolution. Notably, the temporal reasoning capacity inherent in state tracking, which models how object states evolve across frames, generalizes naturally to periodic structures, making O-VAD's training-free design readily extensible to the reference-based evaluation protocol of IPAD.

\subsection{Hyperparameter Details.}
\label{sec:hyperparams}

O-VAD inherits and extends the threshold-based filtering mechanism of TubeletGraph~\cite{sun2025tracking}.
Table~\ref{tab:hyperparams} summarizes every threshold used across the three stages,
together with its default value and the stage in which it operates.
Below we discuss each threshold, its design rationale, and sensitivity.


\paragraph{VLM detection confidence.} We set $\tau_{\text{conf,1}}$ $0.1$.
During VLM-grounded object discovery (\S3.2), SAM3~\cite{carion2025sam} segments objects identified by the VLM, and we retain detections with confidence above 0.1. This threshold is intentionally very permissive to maximize recall at the discovery stage: missing a task-relevant object early in the pipeline is irrecoverable, whereas false-positive objects can be pruned during downstream state tracking and anomaly reasoning.

\begin{table}[h]
\centering
\caption{Summary of threshold hyperparameters used in O-VAD.}
\label{tab:hyperparams}
\footnotesize
\resizebox{\linewidth}{!}{%
\begin{tabular}{lllccl}
\toprule
\textbf{Symbol} & \textbf{Name} & \textbf{Definition} & \textbf{Value} & \textbf{Stage} & \textbf{Origin} \\
\midrule
$\tau_{\text{conf,1}}$   & VLM detection confidence & Confidence threshold for retaining VLM-grounded object detections during discovery    & 0.1  & 1 & Ours \\
\midrule
$\tau_{\text{coverage}}$ & Coverage threshold & Threshold for initiating new tubelets to avoid redundant spatial tracking   & 0.25 & 2 & ~\cite{sun2025tracking} \\
$\tau_{\text{prox}}$    & Spatial proximity gate           & Threshold ensuring new tracks emerge spatially near existing objects         & 0.3  & 2 & ~\cite{sun2025tracking} \\
$\tau_{\text{sem}}$    & Semantic consistency gate        & Threshold filtering tracks by CLIP semantic similarity to the target object           & 0.7  & 2 & ~\cite{sun2025tracking} \\
\midrule
$\tau_{\text{hi}}$     & High-confidence bypass  & Threshold skipping verification for already overwhelmingly confident anomaly detections       & 0.8  & 3 & Ours \\
$\tau_{\text{lo}}$      & Low-confidence discard & Threshold discarding low-confidence anomaly candidates as likely noise                   & 0.2  & 3 & Ours \\
$\tau_{\text{conf}}$   & Post-verification retention  & Threshold retaining verified anomalies after multiplicative confidence recalibration         & 0.3  & 3 & Ours \\
\bottomrule
\end{tabular}%
}
\end{table}



\paragraph{Coverage threshold.} We set $\tau_{\text{coverage}}$ $0.25$.
When constructing the spatiotemporal partition (Eq.(4)), a new tubelet is initiated at frame $t>1$ only when less than $\tau_{\text{coverage}}$ of a newly detected entity's area is already covered by existing tubelets.
This value ensures completeness without spawning redundant tracks for regions that are already well-covered. It ensures that any region that is mostly untracked receives its own tubelet.
In the original TubeletGraph work, this threshold was set without further tuning and shown to be robust across VOST~\cite{tokmakov2023breaking},
M$^3$-VOS~\cite{chen2025m}, and VSCOS~\cite{yu2023video}, so we adopt the same value.


\paragraph{Spatial proximity threshold.} We choose $\tau_{\text{prox}}=0.3$.
A candidate track $C$ emerging at frame~$s$ is retained only if its initial mask overlaps with at least one of SAM2~\cite{ravi2024sam2}'s three multi-mask candidates at that frame by more than 30\% (Eq.~5).
The physical motivation is that a transformed or fragmented object (e.g.\ a piece breaking off, liquid being released) initially appears near the
original object. The threshold is deliberately permissive because SAM2's multi-mask output already expands the effective search region to account for segmentation ambiguity during transformation.
A grid search over $\tau_{\text{prox}} \in \{0.1, 0.2, 0.3, 0.4, 0.5\}$ on the VOST train split shows that tracking performance ($\mathcal{J}$) varies by at
most 1.5 points across this range \cite{sun2025tracking}, confirming low sensitivity.

\paragraph{Semantic consistency threshold.} We use $\tau_{\text{sem}}=0.7$.
A candidate track must exhibit a maximum masked CLIP~\cite{radford2021clip} cosine similarity
exceeding 0.7 with the prompt object across all valid frame pairs (Eq.~6).
This threshold is stricter than $\tau_{\text{prox}}$ because semantic filtering is the primary defence against false positives:
spatially proximal but semantically unrelated entities
(e.g.\ an operator's hand entering the scene) must be rejected.
At the same time, the threshold cannot be too aggressive (\eg~{0.9}),
because genuinely related post-transformation objects (\eg~{foil box $\to$ foil sheet, intact bottle $\to$ deformed bottle with liquid}) often have only moderate CLIP similarity due to appearance change.
The same grid search shows that $\tau_{\text{sem}} = 0.7$ is near-optimal and performance is stable across the $[0.5, 0.8]$ range, dropping noticeably only at 0.9.



\paragraph{High-confidence bypass.} We use $\tau_{\text{hi}}=0.8$.
Anomalies with $c_{\text{orig}} > \tau_{\text{hi}}$ bypass visual verification entirely given the original confidence $c_{\text{orig}}$.
These represent cases where the accumulated state-change evidence is already overwhelming (\eg~clearly visible leakage or material release);
re-verifying would add latency without altering the decision.

\paragraph{Low-confidence discard.} We use $\tau_{\text{lo}}=0.2$.
Anomalies with $c_{\text{orig}} < \tau_{\text{lo}}$ are discarded without verification.
The reasoning chain itself is not confident in these candidates, and empirically they correspond to noise in the state-change events rather than genuine anomalies.

\paragraph{Post-verification retention.} We use $\tau_{\text{conf}}=0.3$.
For the intermediate band $\tau_{\text{lo}} \le c_{\text{orig}} \le \tau_{\text{hi}}$, the final confidence after multiplicative gating exceeds 0.3 for the anomaly to be retained.
The ablation study (Table~3) shows that removing the post-verifier degrades AUC by up to 0.178 (\eg{sticky roller}), confirming that this recalibration sharpens the separation between true anomalies and ambiguous
cases without flipping high-confidence predictions.

\subsection{Stage-wise Sampling and Grounding Details}
\label{sec:supp_sampling}

\paragraph{Frame sampling across stages.}
O-VAD applies a distinct, mostly deterministic sampling rule at each stage, summarized in Table~\ref{tab:sampling}. Stage~1 queries the VLM for the object inventory once on frame~$0$ and selects the SAM mask-prompt frame from 5 uniformly sampled candidates (\texttt{step}\,$=\!N/5$), retaining all frames for downstream use. Stage~2 subsamples at a content-adaptive frame rate set by an inter-stage VLM call, clamped to $[2,10]$ FPS (defaulting to $3$ if the VLM is unavailable) to preserve key state transitions while reducing redundancy. Stage~3 loads frames at a fixed dataset-specific stride (60 / 30 / 25 for Phys-AD / LiquidAD / IPAD) for visual verification. All rules are deterministic except the Stage-2 rate, which is content-adaptive.

\begin{table}[h]
\centering
\caption{Frame sampling policy across O-VAD stages.}
\label{tab:sampling}
\setlength{\tabcolsep}{5pt}
\renewcommand{\arraystretch}{1.15}
\resizebox{\columnwidth}{!}{%
\begin{tabular}{l l l l}
\toprule
\textbf{Stage} & \textbf{Sampling rule} & \textbf{Rate / count} & \textbf{Determinism} \\
\midrule
1. Grounding & Uniform scan + max-confidence selection & 5 frames for SAM prompt; all frames retained & deterministic \\
2. Tracking  & Inter-stage VLM, content-adaptive FPS    & FPS\,$\in\![2,10]$ (default 3 if VLM unavailable) & content-adaptive \\
3. Reasoning & Dataset-specific stride for verification & 60\,/\,30\,/\,25 (Phys-AD\,/\,LiquidAD\,/\,IPAD) & deterministic \\
\bottomrule
\end{tabular}%
}
\end{table}

\paragraph{Object deduplication (whole vs.\ parts).}
Stage~1 issues a \emph{single} open-vocabulary VLM query requesting all distinct objects, and SAM is invoked with the full name list as a multi-class query that produces \emph{one fused binary mask}. Whole and part labels (\eg, ``screw'' and ``screw thread'') therefore collapse into a single connected mask region rather than spawning two tubelets. The textual labels are retained verbatim only as prompts for downstream per-tubelet identification, not as identity counts. Because naming is performed exactly once (on frame~$0$), no inter-frame label fusion is required in Stage~1; persistent identity is established later by SAM2 tracking in Stage~2.

\paragraph{Spatial cues.}
Stage~1 persists two artifacts: a binary segmentation mask over the selected frame and an object-name list. Richer per-object spatial descriptors (bounding box, refined state, material) are produced downstream by the Stage~2 per-tubelet identification prompt, which returns structured trajectories with \texttt{OBJECT}/\texttt{STATE}/\texttt{MATERIAL}/\texttt{BBOX} fields.

\paragraph{Spatial proximity and deformation.}
The Stage-2 proximity and semantic-consistency formulation is inherited from TubeletGraph~\cite{sun2025tracking}. Two properties make it robust under large deformation: (i) proximity is computed against SAM2's \emph{three} multi-mask candidates, not the single current-frame mask, explicitly accounting for segmentation ambiguity during transformation; and (ii) semantic consistency uses CLIP mask-pooled features rather than raw appearance, tolerating visual change while preserving identity.

\subsection{Anomaly Type Evaluation.}
\label{sec:bert-type-eval}

\paragraph{BERT similarity.}
Since O-VAD produces open-ended anomaly type labels, exact-match accuracy is inapplicable. We instead measure semantic alignment using BERT-base-uncased~\cite{devlin2019bert} cosine similarity.
For each video, a \emph{reference} string is the ground-truth anomaly-type label (or \texttt{``normal''} for non-anomalous videos), and a \emph{candidate} string concatenates the predicted \texttt{anomaly\_type} and \texttt{anomaly\_subtype} fields across all
detected anomalies (semicolon-delimited), defaulting to
\texttt{``normal''} when none is detected.
Both strings are encoded via the \texttt{[CLS]} embedding from BERT's final hidden layer. The reported BERT score is the mean cosine similarity over all matched video--report pairs (covering both normal and abnormal ground truths). This metric naturally accommodates synonyms and paraphrases
(e.g.\ ``fastener head stripping'' vs.\ ``screw stripping''), and it should be read jointly with classification metrics.

\subsection{Full prompts of proprietary models.}


\paragraph{\textbf{Prompts in Stage~1.}}During object grounding, the VLM is prompted to enumerate all task-relevant objects in each sampled frame and return their descriptions as a structured JSON array for downstream segmentation with SAM.

\begin{promptbox}[Prompt for $p_{\mathrm{detect}}$ \textnormal{(Stage~1: Object Detection)}]
\footnotesize\ttfamily
"role": "user",\\
"content": (\\
\quad f"\{$I_{t_k}$\}"\\
\quad \hlp{"Analyze this image and list ALL distinct objects/items visible."}\\[2pt]
\quad "For each object, provide a SHORT, SPECIFIC description\\
\quad that could be used to identify it.\\
\quad Focus on:\\
\quad - Main objects (not background elements)\\
\quad - Objects that could be tracked in a video\\
\quad - Use specific descriptions (e.g., `red apple' not just `fruit')"\\[2pt]
\quad \hlp{"Return ONLY a JSON array of object descriptions, nothing else."}\\
\quad "Example format: [\textbackslash"red apple being cut\textbackslash",\\
\quad \textbackslash"kitchen knife with black handle\textbackslash", \textbackslash"green cutting board\textbackslash"]"\\
)
\end{promptbox}


\paragraph{\textbf{Prompts in Stage~2.}}Before tracking state changes, the VLM is queried once per object on its first visible frame to establish a semantic baseline comprising its name, physical state, and material.

\begin{promptbox}[Prompt for $p_{\mathrm{id}}$ \textnormal{(Stage~2: Object Identification)}]
\footnotesize\ttfamily
"role": "user",\\
"content": (\\
\quad f"\{$\tilde{I}_1^i$\}"  \textnormal{\color{gray}\% highlighted first frame}\\
\quad "Look at this image from a video.\\
\quad The object of interest is highlighted with a\\
\quad \{init\_c\_name\} contour."\\[2pt]
\quad \hlp{"Please describe:"}\\
\quad \hlp{"1. Object name (be specific)"}\\
\quad \hlp{"2. Current state (intact, squeezed, opened, etc.)"}\\
\quad \hlp{"3. Material type (plastic, metal, liquid inside, etc.)"}\\
\quad "4. Bounding Box (spatial information)."\\[2pt]
\quad "Answer format:\\
\quad OBJECT: [name]\\
\quad STATE: [current state]\\
\quad MATERIAL: [material]\\
\quad BBOX: [bbox]"\\
)
\end{promptbox}


\noindent The VLM compares color-highlighted visualizations of the same object across two time steps and reports any obvious state change with an open-ended type, description, and severity label.

\begin{promptbox}[Prompt for $p_{\mathrm{state}}$ \textnormal{(Stage~2: State Change Detection)}]
\footnotesize\ttfamily
"role": "system",\\
"content": "You are an expert video analysis assistant\\
\quad specialized in detecting object state changes and interactions.\\
\quad Your tasks include: 1. Identifying objects and their current states;\\
\quad 2. Detecting OBVIOUS change in object appearance, shape, material\\
\quad or moving state (slight changes should be ignored);\\
\quad 3. Recognizing interactions between objects;\\
\quad 4. Check for dynamics like material flow, deformation, or leaking."\\[4pt]
"role": "user",\\
"content": (\\
\quad f"\{$\tilde{I}_{t_k}^i$\}"\quad f"\{$\tilde{I}_{t_{k+1}}^i$\}"\\
\quad \hlp{"Compare these two frames from a video."}\\
\quad "The same object is highlighted with \{init\_c\_name\}\\
\quad contour in both."\\[2pt]
\quad "First image: Earlier frame\\
\quad Second image: Later frame"\\[2pt]
\quad \hlp{"Carefully check for OBVIOUS changes:"}\\
\quad \hlp{"- DEFORMATION: Is the object's shape different?"}\\
\quad \hlp{"- MATERIAL\_RELEASE: Is anything coming out of the object?"}\\
\quad \hlp{"- SURFACE\_CHANGE: Any cracks, tears, openings, damage?"}\\
\quad \hlp{"- TEXTURE\_CHANGE: Has the surface appearance changed?"}\\
\quad \hlp{"- MOVEMENT\_CHANGE: Has the object changed between static and moving?"}\\
\quad \hlp{"- INTERACTION\_CHANGE: Are the objects interactively move together?"}\\[2pt]
\quad "Answer format:\\
\quad STATE\_CHANGED: [yes/no]\\
\quad CHANGE\_TYPE: [deformation/material\_release/surface\_change/\\
\quad\quad size\_change/none/...]\\
\quad CHANGE\_DESCRIPTION: [describe what changed in detail]\\
\quad CHANGE\_SEVERITY: [none/slight/moderate/severe]"\\
)
\end{promptbox}


\noindent When multiple objects coexist in a frame, the VLM is additionally queried to characterize inter-object interactions, identifying the contact type, force direction, and visible effect on the manipulated object.

\begin{promptbox}[Prompt for $p_{\mathrm{inter}}$ \textnormal{(Stage~2: Interaction Analysis)}]
\footnotesize\ttfamily
"role": "user",\\
"content": (\\
\quad f"\{$\tilde{I}_{t}$\}"  \textnormal{\color{gray}\% both objects highlighted}\\
\quad "Analyze the interaction between objects in this image.\\
\quad Object 1 (being manipulated): \{init\_c\_name\} contour\\
\quad Object 2 (manipulating): \{query\_c\_name\} contour"\\[2pt]
\quad \hlp{"Describe the interaction:"}\\
\quad \hlp{"1. CONTACT\_TYPE: How are they touching?"}\\
\quad \hlp{"2. FORCE\_DIRECTION: Where is force being applied?"}\\
\quad \hlp{"3. VISIBLE\_EFFECT: What effect is visible on Object 1?"}\\
\quad \hlp{"4. ACTION\_VERB: What action is happening?"}\\[2pt]
\quad "Answer format:\\
\quad CONTACT\_TYPE: [type]\\
\quad FORCE\_DIRECTION: [direction]\\
\quad VISIBLE\_EFFECT: [effect description]\\
\quad ACTION\_VERB: [single verb describing the action]"\\
)
\end{promptbox}


\paragraph{\textbf{Prompts in Stage~3.}}The core anomaly reasoning prompt supplies the VLM with all accumulated evidence---object metadata, state change events, video caption, and sampled frames---and instructs it to follow a six-step chain-of-thought from observation through severity assessment, producing free-form anomaly classifications.

\begin{promptbox}[Prompt for $p_{\mathrm{CoT}}$ \textnormal{(Stage~3: Anomaly Reasoning)}]
\footnotesize\ttfamily
"role": "system",\\
"content": "You are an expert anomaly detection system for industrial\\
\quad processes. You have access to object tracking data including object\\
\quad metadata (descriptions, materials, initial states), video caption,\\
\quad and fine-grained state change events detected across video frames.\\
\quad ...\\
\quad Be thorough but avoid false positives. Consider physical plausibility\\
\quad and context. Always reference specific object IDs, frame ranges,\\
\quad and state change events in your reasoning."\\[4pt]
"role": "user",\\
"content": (\\
\quad f"\{sampled frames\}"\\
\quad "TASK CONTEXT: \{task\_context\}\\
\quad VIDEO CAPTION: \{caption\}\\
\quad TRACKED OBJECTS: \{object\_info\}\\
\quad STATE CHANGES DETECTED: \{state\_changes\}"\\[2pt]
\quad \hlp{"Anomalies should cause or reflect real flaw in object,"}\\
\quad \hlp{"real flaw in pipeline or real damage to object."}\\[2pt]
\quad \hlp{"Follow this 6-step reasoning chain:"}\\
\quad \hlp{"STEP 1 - OBSERVATION: What specific changes and events"}\\
\quad \hlp{"\quad occurred? Reference tracked objects by their IDs."}\\
\quad \hlp{"STEP 2 - EXPECTATION: What should have happened given"}\\
\quad \hlp{"\quad the task context and object materials?"}\\
\quad \hlp{"STEP 3 - COMPARISON: How do observed state changes"}\\
\quad \hlp{"\quad differ from expectations?"}\\
\quad \hlp{"STEP 4 - CAUSATION: What could cause these deviations?"}\\
\quad \hlp{"\quad Reason freely -- do NOT limit to predefined categories."}\\
\quad \hlp{"STEP 5 - CLASSIFICATION: Classify anomalies using your"}\\
\quad \hlp{"\quad own judgment. You are free to name the anomaly type."}\\
\quad \hlp{"STEP 6 - SEVERITY: Rate severity (none/low/medium/"}\\
\quad \hlp{"\quad high/critical) based on impact, safety, reversibility."}\\[2pt]
\quad "Output your analysis as JSON: \{reasoning, anomalies,\\
\quad is\_anomalous, overall\_severity, summary\}"\\
)
\end{promptbox}


\noindent Each candidate anomaly with intermediate confidence is verified against evidence frames and the video caption, explicitly checking whether the claimed anomaly is a genuine failure or a normal process behavior.

\begin{promptbox}[Prompt for $p_{\mathrm{verify}}$ \textnormal{(Stage~3: Visual Verification)}]
\footnotesize\ttfamily
"role": "user",\\
"content": (\\
\quad f"\{evidence frames\}"\\[2pt]
\quad "Video Caption: \{caption\}"\\[2pt]
\quad "CLAIMED ANOMALY:\\
\quad Type: \{anomaly\_type\}\\
\quad Description: \{anomaly\_description\}\\
\quad Affected objects: \{affected\_objects\}"\\[2pt]
\quad \hlp{"Based on the video caption and frames, verify if this"}\\
\quad \hlp{"CLAIMED ANOMALY really exists following these steps:"}\\
\quad \hlp{"1. Check Video Caption, do you find evidence of the"}\\
\quad \hlp{"\quad claimed anomaly in caption or frames?"}\\
\quad \hlp{"2. Is the description accurate according to either"}\\
\quad \hlp{"\quad caption or frames?"}\\
\quad \hlp{"3. What is your confidence level?"}\\[2pt]
\quad "Output JSON: \{verified: true/false,\\
\quad confidence: 0.0--1.0.\}"\\
)
\end{promptbox}

\subsection{Segmentation and Tracking Model Selection.}
\label{sec:model-selection}

O-VAD employs three distinct segmentation and tracking models---CropFormer,
SAM2, and SAM3---at different pipeline stages.
Each choice is motivated by the specific input-output requirements of that
stage, as summarised in Table~\ref{tab:models} and discussed below.

\begin{table}[h]
\centering
\caption{Segmentation and tracking models used in O-VAD and the
requirements that motivate each choice.}
\label{tab:models}
\resizebox{\linewidth}{!}{%
\begin{tabular}{lccp{5.8cm}}
\toprule
\textbf{Model} & \textbf{Stage} & \textbf{Task} & \textbf{Key Requirement} \\
\midrule
SAM3~\cite{carion2025sam}   & 1 & Object grounding
    & Concept-aware: bridge VLM text output $\to$ pixel masks \\
CropFormer~\cite{qi2022high} & 2 & Entity segmentation
    & Exhaustive class-agnostic dense partition of every frame \\
SAM2~\cite{ravi2024sam2}  & 2 & Tubelet propagation
    & Memory-based temporal tracking with multi-mask output \\
\bottomrule
\end{tabular}%
}
\end{table}

\paragraph{\textbf{Stage~1: SAM3 for Concept-Aware Segmentation.}}

Stage~1 (\S3.2) requires converting VLM-generated object inventories, including names, natural-language descriptions, and spatial cues, into precise pixel-level masks. This is a \emph{semantic-to-mask} task where the input is a textual concept, not a geometric prompt (point, box, or scribble). 
SAM3 (Segment Anything with Concepts)~\cite{carion2025sam} extends the SAM family with concept-level understanding, enabling it to segment objects directly from semantic descriptions.
Neither the original SAM~\cite{kirillov2023segment} nor SAM2~\cite{ravi2024sam2} supports text-conditioned segmentation; using them would require an additional step to convert VLM descriptions into point or box prompts, introducing error and complexity.
SAM3's concept-aware capability thus provides the most direct bridge between the VLM's structured output and the per-object masks
$M_1^i = \text{SAM3}(I_1, o_i)$ required by downstream stages.

\paragraph{\textbf{Stage~2: CropFormer for Entity Segmentation.}}

The spatiotemporal partitioning step (\S3.3) requires an \emph{exhaustive, class-agnostic} spatial partition of each frame, where every visually distinct
region must be segmented, not just the target object.
This is critical because the partition forms the search space for recovering post-transformation objects since any region that is not segmented can never be recovered as a candidate tubelet.

CropFormer~\cite{qi2022high} is an entity segmentation model specifically designed for high-quality, class-agnostic segmentation of all ``things'' in an image.
Its crop-based architecture handles objects across a wide range of scales, which is important in industrial settings where anomaly-relevant details (\eg~a small leak or a stripped screw head) can be very small relative to the frame.
The ablations in~\cite{sun2025tracking} shows that
replacing CropFormer with SAM2 automasks degrades $\mathcal{J}$ by 1.7 points, primarily due to SAM's reduced reliability for small objects.

Critically, CropFormer is a per-frame model with no temporal component. This is by design: each frame's partition is computed independently so that newly appearing entities (post-transformation fragments, released material) are detected as soon as they become visible, without requiring any temporal prior that might suppress them.

\paragraph{\textbf{Stage~2: SAM2 for Tubelet Propagation.}}

Once per-frame entities are detected by CropFormer, each entity is tracked forward in time via SAM2~\cite{ravi2024sam2} to form tubelets (Eq.~4).
SAM2 is designed for video-native object segmentation with a memory-attention mechanism that maintains temporal consistency.
Two additional properties make it particularly well-suited for this role:
%
(i)\textit{Multi-mask output.} SAM2 produces three candidate masks per frame, capturing segmentation ambiguity. O-VAD's spatial proximity prior ($\tau_{\text{prox}}$, Eq.~5) explicitly exploits these candidate masks to estimate where post-transformation objects might appear, effectively expanding the search region during state changes.
(ii)\textit{Object confidence scores.} SAM2 outputs per-frame object score logits that indicate tracking reliability. These scores serve as an additional signal for detecting when a track is being lost, which is a potential indicator of an ongoing state transformation.

Meanwhile, the ablation in \cite{sun2025tracking} also shows that replacing SAM2 with Cutie~\cite{cheng2024putting} for tubelet propagation causes a more significant degradation ($3.3$ in $\mathcal{J}$ and $9.3$ in temporal recall $\mathcal{T}_R$), confirming that SAM2's tracking quality is critical for the partition's completeness.

\paragraph{\textbf{Why Not a Single Model?}}

Each stage has fundamentally different input-output requirements:
Stage~1 needs \emph{text $\to$ mask} (concept-aware segmentation),
Stage~2's partitioning needs \emph{frame $\to$ all masks} (exhaustive entity segmentation),
and Stage~2's tracking needs \emph{mask $\to$ mask sequence}
(temporal propagation with memory).
No single existing model satisfies all three.
SAM3 can segment named objects but is not designed for exhaustive spatial
partitioning;
CropFormer produces dense per-frame segments but does not perform temporal
tracking;
and SAM2 excels at temporal propagation from a given mask but requires
geometric prompts and does not perform class-agnostic partitioning.
The modular design allows each component to operate at its respective
strength, and the clean interfaces between stages (masks, tubelets,
state-change events) make the pipeline straightforward to update as
better models become available.

\section{Detailed Per-Category Quantitative Results}
\label{sec:supp_detailed_results}
\subsection{Detailed Results on Phys-AD}
\label{sec:supp_physad}

\begin{table*}[t]
\centering
\caption{\textbf{Video- and type-level results on Phys-AD~\cite{li2025towards} across 22 object categories.}
``$\dagger$'' denotes methods requiring training data; ``$\ast$'' denotes training-free methods.
``BERT'' is the type-level BERTScore, reported only for methods that emit free-form anomaly-type descriptions.
For each row, the best / 2nd-best across all methods is highlighted: video-level metrics in
\colorbox{cellRed}{\rule{0pt}{1ex}\rule{1ex}{0pt}} / \colorbox{cellRedLight}{\rule{0pt}{1ex}\rule{1ex}{0pt}},
type-level BERT in \colorbox{cellPurple}{\rule{0pt}{1ex}\rule{1ex}{0pt}} / \colorbox{cellPurpleLight}{\rule{0pt}{1ex}\rule{1ex}{0pt}};
our best results are additionally in \textbf{bold}. Blank cells indicate results not available.
The 22 categories are split into two side-by-side blocks; ``Avg.'' reports the average over all categories.}
\label{tab:physad-full}
\setlength{\tabcolsep}{2.5pt}
\renewcommand{\arraystretch}{1.0}
\resizebox{\textwidth}{!}{%
\begin{tabular}{@{}ll ccccccc @{\hspace{5pt}}|@{\hspace{5pt}} ll ccccccc@{}}
\toprule
& & \multicolumn{2}{c}{\textbf{Trad.}} & \multicolumn{2}{c}{\textbf{Direct}} & \multicolumn{2}{c}{\textbf{Wkf.}} & \multicolumn{1}{c}{\textbf{Ours}} & & & \multicolumn{2}{c}{\textbf{Trad.}} & \multicolumn{2}{c}{\textbf{Direct}} & \multicolumn{2}{c}{\textbf{Wkf.}} & \multicolumn{1}{c}{\textbf{Ours}}\\
\cmidrule(lr){3-4}\cmidrule(lr){5-6}\cmidrule(lr){7-8}\cmidrule(lr){9-9}\cmidrule(lr){12-13}\cmidrule(lr){14-15}\cmidrule(lr){16-17}\cmidrule(lr){18-18}
\textbf{Cat.} & \textbf{Met.} & MNAD$^\dagger$ & S3R$^\dagger$ & Qwen3$^\ast$ & GPT5$^\ast$ & URF$^\ast$ & VERA$^\dagger$ & \textbf{O-VAD}$^\ast$ & \textbf{Cat.} & \textbf{Met.} & MNAD$^\dagger$ & S3R$^\dagger$ & Qwen3$^\ast$ & GPT5$^\ast$ & URF$^\ast$ & VERA$^\dagger$ & \textbf{O-VAD}$^\ast$\\
\midrule
\multirow{6}{*}{\textbf{Ball}} & Acc & 0.637 & 0.326 & 0.504 & \cellcolor{cellRedLight}0.677 & \cellcolor{cellRed}0.704 & 0.477 & 0.570 & \multirow{6}{*}{\textbf{Magnet}} & Acc & 0.533 & \cellcolor{cellRedLight}0.767 & 0.544 & \cellcolor{cellRed}0.908 & 0.589 & 0.631 & 0.689 \\
 & P & 0.681 & 0.000 & \cellcolor{cellRedLight}0.695 & 0.667 & 0.692 & 0.667 & \cellcolor{cellRed}\textbf{0.742} &  & P & 0.680 & 0.800 & 0.771 & \cellcolor{cellRedLight}0.968 & 0.717 & \cellcolor{cellRed}1.000 & 0.700 \\
 & R & \cellcolor{cellRedLight}0.856 & 0.000 & 0.456 & 0.800 & \cellcolor{cellRed}1.000 & 0.057 & 0.544 &  & R & 0.567 & \cellcolor{cellRedLight}0.867 & 0.450 & 0.857 & 0.633 & 0.314 & \cellcolor{cellRed}\textbf{0.933} \\
 & F1 & \cellcolor{cellRedLight}0.759 & 0.000 & 0.550 & 0.727 & \cellcolor{cellRed}0.818 & 0.105 & 0.628 &  & F1 & 0.618 & \cellcolor{cellRedLight}0.832 & 0.568 & \cellcolor{cellRed}0.909 & 0.673 & 0.478 & 0.800 \\
 & AUC & \cellcolor{cellRedLight}0.589 & 0.155 & 0.528 & \cellcolor{cellRed}0.744 & 0.461 & 0.548 & 0.579 &  & AUC & 0.512 & \cellcolor{cellRedLight}0.724 & 0.592 & 0.364 & 0.518 & \cellcolor{cellRed}0.892 & 0.576 \\
 & BERT &  &  & \cellcolor{cellPurpleLight}0.821 &  & 0.642 &  & \cellcolor{cellPurple}\textbf{0.862} &  & BERT &  &  & \cellcolor{cellPurple}0.781 &  & 0.643 &  & \cellcolor{cellPurpleLight}0.699 \\
\cmidrule(lr){2-9}\cmidrule(lr){11-18}
\multirow{6}{*}{\textbf{Button}} & Acc & \cellcolor{cellRed}0.697 & 0.203 & 0.330 & 0.400 & \cellcolor{cellRedLight}0.580 & 0.462 & 0.380 & \multirow{6}{*}{\textbf{Roll. Bear.}} & Acc & 0.433 & 0.483 & 0.429 & \cellcolor{cellRed}0.625 & 0.517 & \cellcolor{cellRedLight}0.536 & 0.536 \\
 & P & 0.807 & \cellcolor{cellRed}1.000 & 0.809 & 0.375 & \cellcolor{cellRedLight}0.967 & 0.500 & 0.781 &  & P & 0.462 & 0.000 & 0.350 & \cellcolor{cellRed}0.857 & \cellcolor{cellRedLight}0.509 & 0.000 & 0.500 \\
 & R & \cellcolor{cellRed}0.817 & 0.004 & 0.212 & 0.171 & \cellcolor{cellRedLight}0.492 & 0.029 & 0.312 &  & R & \cellcolor{cellRedLight}0.800 & 0.000 & 0.269 & 0.231 & \cellcolor{cellRed}0.967 & 0.000 & 0.346 \\
 & F1 & \cellcolor{cellRed}0.812 & 0.008 & 0.337 & 0.235 & \cellcolor{cellRedLight}0.652 & 0.054 & 0.446 &  & F1 & \cellcolor{cellRedLight}0.585 & 0.000 & 0.304 & 0.364 & \cellcolor{cellRed}0.667 & 0.000 & 0.409 \\
 & AUC & \cellcolor{cellRedLight}0.526 & 0.108 & 0.506 & 0.456 & \cellcolor{cellRed}0.717 & 0.469 & 0.393 &  & AUC & 0.437 & 0.016 & 0.418 & 0.451 & 0.304 & \cellcolor{cellRedLight}0.500 & \cellcolor{cellRed}\textbf{0.650} \\
 & BERT &  &  & \cellcolor{cellPurpleLight}0.780 &  & 0.666 &  & \cellcolor{cellPurple}\textbf{0.861} &  & BERT &  &  & \cellcolor{cellPurpleLight}0.812 &  & 0.626 &  & \cellcolor{cellPurple}\textbf{0.850} \\
\cmidrule(lr){2-9}\cmidrule(lr){11-18}
\multirow{6}{*}{\textbf{Car}} & Acc & \cellcolor{cellRed}0.758 & 0.497 & 0.332 & 0.677 & 0.642 & 0.462 & \cellcolor{cellRedLight}0.742 & \multirow{6}{*}{\textbf{Rub. Band}} & Acc & 0.567 & 0.583 & \cellcolor{cellRedLight}0.767 & \cellcolor{cellRed}0.817 & 0.733 & 0.500 & 0.633 \\
 & P & 0.760 & \cellcolor{cellRed}0.801 & 0.733 & 0.659 & \cellcolor{cellRedLight}0.761 & 0.000 & 0.748 &  & P & 0.567 & 0.600 & \cellcolor{cellRedLight}0.944 & \cellcolor{cellRed}1.000 & 0.719 & 0.000 & 0.611 \\
 & R & \cellcolor{cellRed}0.991 & 0.438 & 0.171 & 0.829 & 0.762 & 0.000 & \cellcolor{cellRedLight}0.989 &  & R & 0.567 & 0.500 & 0.567 & 0.633 & \cellcolor{cellRed}0.767 & 0.000 & \cellcolor{cellRedLight}0.733 \\
 & F1 & \cellcolor{cellRed}0.860 & 0.566 & 0.278 & 0.734 & 0.761 & 0.000 & \cellcolor{cellRedLight}0.852 &  & F1 & 0.567 & 0.545 & 0.708 & \cellcolor{cellRed}0.775 & \cellcolor{cellRedLight}0.742 & 0.000 & 0.667 \\
 & AUC & \cellcolor{cellRedLight}0.599 & 0.527 & 0.492 & \cellcolor{cellRed}0.617 & 0.497 & 0.383 & 0.575 &  & AUC & 0.561 & 0.463 & \cellcolor{cellRedLight}0.767 & 0.279 & \cellcolor{cellRed}0.783 & 0.276 & 0.721 \\
 & BERT &  &  & \cellcolor{cellPurple}0.809 &  & 0.603 &  & \cellcolor{cellPurpleLight}0.737 &  & BERT &  &  & \cellcolor{cellPurple}0.803 &  & 0.639 &  & \cellcolor{cellPurpleLight}0.784 \\
\cmidrule(lr){2-9}\cmidrule(lr){11-18}
\multirow{6}{*}{\textbf{Caster Wheel}} & Acc & 0.400 & \cellcolor{cellRed}0.833 & 0.300 & \cellcolor{cellRedLight}0.785 & 0.367 & 0.462 & 0.583 & \multirow{6}{*}{\textbf{Screw}} & Acc & 0.622 & 0.600 & 0.444 & \cellcolor{cellRedLight}0.626 & \cellcolor{cellRed}0.689 & 0.460 & 0.600 \\
 & P & 0.737 & \cellcolor{cellRedLight}0.973 & \cellcolor{cellRed}1.000 & 0.957 & 0.818 & 0.000 & 0.717 &  & P & \cellcolor{cellRedLight}0.760 & \cellcolor{cellRed}0.929 & 0.619 & 0.596 & 0.682 & 0.432 & 0.700 \\
 & R & 0.311 & \cellcolor{cellRed}0.800 & 0.067 & 0.629 & 0.200 & 0.000 & \cellcolor{cellRedLight}0.733 &  & R & 0.633 & 0.433 & 0.433 & 0.683 & \cellcolor{cellRed}1.000 & 0.059 & \cellcolor{cellRedLight}0.700 \\
 & F1 & 0.438 & \cellcolor{cellRed}0.878 & 0.125 & \cellcolor{cellRedLight}0.759 & 0.321 & 0.000 & 0.725 &  & F1 & 0.691 & 0.591 & 0.510 & 0.661 & \cellcolor{cellRed}0.811 & 0.096 & \cellcolor{cellRedLight}0.700 \\
 & AUC & 0.468 & \cellcolor{cellRed}0.830 & 0.533 & 0.543 & 0.410 & \cellcolor{cellRedLight}0.609 & 0.301 &  & AUC & \cellcolor{cellRedLight}0.638 & 0.620 & 0.450 & 0.441 & 0.298 & 0.400 & \cellcolor{cellRed}\textbf{0.704} \\
 & BERT &  &  & \cellcolor{cellPurpleLight}0.754 &  & 0.626 &  & \cellcolor{cellPurple}\textbf{0.796} &  & BERT &  &  & \cellcolor{cellPurpleLight}0.779 &  & 0.657 &  & \cellcolor{cellPurple}\textbf{0.784} \\
\cmidrule(lr){2-9}\cmidrule(lr){11-18}
\multirow{6}{*}{\textbf{Clip}} & Acc & \cellcolor{cellRed}0.642 & \cellcolor{cellRedLight}0.622 & 0.461 & 0.585 & 0.344 & 0.154 & 0.403 & \multirow{6}{*}{\textbf{Servo}} & Acc & \cellcolor{cellRedLight}0.671 & 0.333 & 0.362 & 0.508 & \cellcolor{cellRed}0.750 & 0.462 & 0.447 \\
 & P & \cellcolor{cellRedLight}0.817 & 0.703 & 0.619 & 0.565 & \cellcolor{cellRed}1.000 & 0.000 & 0.634 &  & P & \cellcolor{cellRedLight}0.773 & \cellcolor{cellRed}0.833 & 0.737 & 0.527 & 0.752 & 0.000 & 0.726 \\
 & R & 0.596 & \cellcolor{cellRedLight}0.750 & 0.500 & \cellcolor{cellRed}1.000 & 0.017 & 0.000 & 0.246 &  & R & 0.794 & 0.139 & 0.233 & \cellcolor{cellRedLight}0.829 & \cellcolor{cellRed}0.994 & 0.000 & 0.418 \\
 & F1 & 0.689 & \cellcolor{cellRed}0.726 & 0.553 & \cellcolor{cellRedLight}0.722 & 0.033 & 0.000 & 0.354 &  & F1 & \cellcolor{cellRedLight}0.784 & 0.238 & 0.354 & 0.644 & \cellcolor{cellRed}0.856 & 0.000 & 0.530 \\
 & AUC & \cellcolor{cellRed}0.628 & \cellcolor{cellRedLight}0.537 & 0.442 & 0.518 & 0.275 & 0.180 & 0.423 &  & AUC & 0.479 & 0.452 & \cellcolor{cellRedLight}0.492 & 0.330 & 0.342 & 0.341 & \cellcolor{cellRed}\textbf{0.495} \\
 & BERT &  &  & \cellcolor{cellPurpleLight}0.798 &  & 0.640 &  & \cellcolor{cellPurple}\textbf{0.876} &  & BERT &  &  & \cellcolor{cellPurpleLight}0.797 &  & 0.638 &  & \cellcolor{cellPurple}\textbf{0.851} \\
\cmidrule(lr){2-9}\cmidrule(lr){11-18}
\multirow{6}{*}{\textbf{Clock}} & Acc & 0.568 & 0.532 & \cellcolor{cellRedLight}0.644 & \cellcolor{cellRed}0.723 & 0.356 & 0.462 & 0.458 & \multirow{6}{*}{\textbf{Slide}} & Acc & \cellcolor{cellRed}0.647 & 0.320 & 0.387 & \cellcolor{cellRedLight}0.523 & 0.200 & 0.462 & 0.493 \\
 & P & 0.669 & 0.734 & 0.717 & 0.698 & \cellcolor{cellRedLight}0.857 & 0.000 & \cellcolor{cellRed}\textbf{1.000} &  & P & 0.802 & \cellcolor{cellRed}1.000 & \cellcolor{cellRedLight}0.833 & 0.542 & 0.000 & 0.000 & 0.766 \\
 & R & 0.696 & 0.466 & \cellcolor{cellRedLight}0.770 & \cellcolor{cellRed}0.857 & 0.041 & 0.000 & 0.155 &  & R & \cellcolor{cellRedLight}0.742 & 0.150 & 0.292 & \cellcolor{cellRed}0.743 & 0.000 & 0.000 & 0.536 \\
 & F1 & 0.682 & 0.570 & \cellcolor{cellRedLight}0.743 & \cellcolor{cellRed}0.769 & 0.077 & 0.000 & 0.269 &  & F1 & \cellcolor{cellRed}0.771 & 0.261 & 0.432 & 0.626 & 0.000 & 0.000 & \cellcolor{cellRedLight}0.631 \\
 & AUC & 0.513 & 0.529 & 0.581 & \cellcolor{cellRed}0.766 & 0.483 & 0.500 & \cellcolor{cellRedLight}0.669 &  & AUC & \cellcolor{cellRed}0.534 & \cellcolor{cellRedLight}0.533 & 0.529 & 0.528 & 0.345 & 0.225 & 0.415 \\
 & BERT &  &  & \cellcolor{cellPurpleLight}0.792 &  & 0.597 &  & \cellcolor{cellPurple}\textbf{0.900} &  & BERT &  &  & \cellcolor{cellPurpleLight}0.803 &  & 0.639 &  & \cellcolor{cellPurple}\textbf{0.812} \\
\cmidrule(lr){2-9}\cmidrule(lr){11-18}
\multirow{6}{*}{\textbf{Fan}} & Acc & \cellcolor{cellRedLight}0.642 & 0.570 & 0.427 & 0.569 & 0.383 & 0.462 & \cellcolor{cellRed}\textbf{0.771} & \multirow{6}{*}{\textbf{Sph. Bear.}} & Acc & 0.500 & \cellcolor{cellRed}0.867 & 0.483 & 0.467 & \cellcolor{cellRedLight}0.633 & 0.500 & 0.517 \\
 & P & 0.748 & 0.828 & \cellcolor{cellRed}0.846 & 0.667 & \cellcolor{cellRedLight}0.831 & 0.000 & 0.792 &  & P & 0.500 & \cellcolor{cellRed}0.893 & 0.471 & 0.417 & \cellcolor{cellRedLight}0.595 & 0.000 & 0.509 \\
 & R & \cellcolor{cellRedLight}0.787 & 0.537 & 0.287 & 0.400 & 0.220 & 0.000 & \cellcolor{cellRed}\textbf{0.940} &  & R & \cellcolor{cellRed}0.900 & 0.833 & 0.267 & 0.167 & 0.833 & 0.000 & \cellcolor{cellRedLight}0.900 \\
 & F1 & \cellcolor{cellRedLight}0.767 & 0.652 & 0.429 & 0.500 & 0.348 & 0.000 & \cellcolor{cellRed}\textbf{0.860} &  & F1 & 0.643 & \cellcolor{cellRed}0.862 & 0.340 & 0.238 & \cellcolor{cellRedLight}0.694 & 0.000 & 0.651 \\
 & AUC & 0.409 & \cellcolor{cellRed}0.602 & \cellcolor{cellRedLight}0.566 & 0.472 & 0.523 & 0.389 & 0.532 &  & AUC & 0.547 & \cellcolor{cellRed}0.894 & 0.483 & 0.440 & 0.629 & 0.114 & \cellcolor{cellRedLight}0.641 \\
 & BERT &  &  & \cellcolor{cellPurpleLight}0.797 &  & 0.630 &  & \cellcolor{cellPurple}\textbf{0.890} &  & BERT &  &  & \cellcolor{cellPurple}0.826 &  & 0.662 &  & \cellcolor{cellPurpleLight}0.726 \\
\cmidrule(lr){2-9}\cmidrule(lr){11-18}
\multirow{6}{*}{\textbf{Gear}} & Acc & \cellcolor{cellRedLight}0.740 & 0.198 & 0.638 & 0.739 & 0.513 & 0.462 & \cellcolor{cellRed}\textbf{0.819} & \multirow{6}{*}{\textbf{Sticky Roller}} & Acc & 0.511 & 0.689 & 0.511 & \cellcolor{cellRedLight}0.720 & \cellcolor{cellRed}0.756 & 0.529 & 0.689 \\
 & P & 0.801 & 0.000 & \cellcolor{cellRedLight}0.880 & \cellcolor{cellRed}1.000 & 0.808 & 0.000 & 0.797 &  & P & 0.633 & \cellcolor{cellRed}0.944 & 0.700 & 0.674 & \cellcolor{cellRedLight}0.788 & 0.488 & 0.691 \\
 & R & \cellcolor{cellRedLight}0.897 & 0.000 & 0.633 & 0.514 & 0.514 & 0.000 & \cellcolor{cellRed}\textbf{0.953} &  & R & 0.633 & 0.567 & 0.467 & 0.736 & \cellcolor{cellRedLight}0.867 & 0.064 & \cellcolor{cellRed}\textbf{0.967} \\
 & F1 & \cellcolor{cellRedLight}0.847 & 0.000 & 0.737 & 0.679 & 0.628 & 0.000 & \cellcolor{cellRed}\textbf{0.868} &  & F1 & 0.633 & 0.708 & 0.560 & 0.726 & \cellcolor{cellRed}0.825 & 0.105 & \cellcolor{cellRedLight}0.806 \\
 & AUC & 0.418 & 0.358 & \cellcolor{cellRedLight}0.644 & 0.397 & 0.451 & 0.512 & \cellcolor{cellRed}\textbf{0.879} &  & AUC & 0.517 & \cellcolor{cellRedLight}0.778 & 0.533 & 0.523 & 0.713 & 0.474 & \cellcolor{cellRed}\textbf{0.811} \\
 & BERT &  &  & \cellcolor{cellPurpleLight}0.815 &  & 0.643 &  & \cellcolor{cellPurple}\textbf{0.855} &  & BERT &  &  & \cellcolor{cellPurple}0.814 &  & 0.623 &  & \cellcolor{cellPurpleLight}0.689 \\
\cmidrule(lr){2-9}\cmidrule(lr){11-18}
\multirow{6}{*}{\textbf{Hinge}} & Acc & \cellcolor{cellRedLight}0.683 & \cellcolor{cellRed}0.700 & 0.417 & 0.339 & 0.550 & 0.462 & 0.617 & \multirow{6}{*}{\textbf{Toothpaste}} & Acc & 0.244 & 0.589 & 0.578 & \cellcolor{cellRed}0.908 & 0.567 & 0.431 & \cellcolor{cellRedLight}0.656 \\
 & P & 0.760 & \cellcolor{cellRed}1.000 & \cellcolor{cellRedLight}0.917 & 0.409 & 0.821 & 0.000 & 0.750 &  & P & 0.233 & \cellcolor{cellRedLight}0.700 & 0.684 & \cellcolor{cellRed}0.914 & 0.583 & 0.417 & 0.667 \\
 & R & \cellcolor{cellRed}0.844 & 0.600 & 0.244 & 0.514 & 0.511 & 0.000 & \cellcolor{cellRedLight}0.733 &  & R & 0.222 & 0.311 & 0.289 & \cellcolor{cellRed}0.914 & 0.467 & 0.143 & \cellcolor{cellRedLight}0.622 \\
 & F1 & \cellcolor{cellRed}0.800 & \cellcolor{cellRedLight}0.750 & 0.386 & 0.456 & 0.630 & 0.000 & 0.742 &  & F1 & 0.227 & 0.431 & 0.406 & \cellcolor{cellRed}0.914 & 0.519 & 0.213 & \cellcolor{cellRedLight}0.644 \\
 & AUC & 0.497 & \cellcolor{cellRed}0.716 & \cellcolor{cellRedLight}0.589 & 0.516 & 0.544 & 0.041 & 0.526 &  & AUC & 0.257 & 0.535 & 0.578 & \cellcolor{cellRed}0.832 & 0.528 & 0.579 & \cellcolor{cellRedLight}0.689 \\
 & BERT &  &  & \cellcolor{cellPurple}0.807 &  & 0.643 &  & \cellcolor{cellPurpleLight}0.790 &  & BERT &  &  & \cellcolor{cellPurpleLight}0.788 &  & 0.619 &  & \cellcolor{cellPurple}\textbf{0.870} \\
\cmidrule(lr){2-9}\cmidrule(lr){11-18}
\multirow{6}{*}{\textbf{Liquid}} & Acc & 0.578 & \cellcolor{cellRedLight}0.756 & \cellcolor{cellRed}0.822 & 0.750 & 0.644 & 0.683 & 0.667 & \multirow{6}{*}{\textbf{U Disk}} & Acc & 0.504 & \cellcolor{cellRed}0.613 & 0.442 & 0.462 & 0.500 & 0.446 & \cellcolor{cellRedLight}0.512 \\
 & P & 0.641 & 0.828 & 0.923 & \cellcolor{cellRedLight}0.941 & 0.938 & \cellcolor{cellRed}1.000 & 0.667 &  & P & 0.502 & \cellcolor{cellRed}0.599 & 0.445 & 0.000 & 0.000 & 0.444 & \cellcolor{cellRedLight}0.533 \\
 & R & \cellcolor{cellRedLight}0.833 & 0.800 & 0.800 & 0.533 & 0.500 & 0.367 & \cellcolor{cellRed}\textbf{1.000} &  & R & \cellcolor{cellRed}0.975 & \cellcolor{cellRedLight}0.683 & 0.475 & 0.000 & 0.000 & 0.114 & 0.200 \\
 & F1 & 0.725 & \cellcolor{cellRedLight}0.814 & \cellcolor{cellRed}0.857 & 0.681 & 0.652 & 0.537 & 0.800 &  & F1 & \cellcolor{cellRed}0.663 & \cellcolor{cellRedLight}0.638 & 0.460 & 0.000 & 0.000 & 0.182 & 0.291 \\
 & AUC & 0.453 & 0.733 & \cellcolor{cellRed}0.833 & 0.512 & 0.691 & \cellcolor{cellRedLight}0.827 & 0.776 &  & AUC & 0.513 & \cellcolor{cellRedLight}0.585 & 0.442 & 0.437 & 0.404 & 0.311 & \cellcolor{cellRed}\textbf{0.606} \\
 & BERT &  &  & \cellcolor{cellPurple}0.817 &  & 0.640 &  & \cellcolor{cellPurpleLight}0.666 &  & BERT &  &  & \cellcolor{cellPurpleLight}0.785 &  & 0.650 &  & \cellcolor{cellPurple}\textbf{0.906} \\
\cmidrule(lr){2-9}\cmidrule(lr){11-18}
\multirow{6}{*}{\textbf{Lock}} & Acc & \cellcolor{cellRedLight}0.678 & 0.328 & 0.367 & \cellcolor{cellRed}0.785 & 0.333 & 0.462 & 0.361 & \multirow{6}{*}{\textbf{Zipper}} & Acc & 0.472 & \cellcolor{cellRed}0.706 & 0.406 & 0.554 & 0.350 & 0.462 & \cellcolor{cellRedLight}0.600 \\
 & P & \cellcolor{cellRedLight}0.682 & 0.000 & 0.542 & \cellcolor{cellRed}0.957 & 0.000 & 0.000 & 0.593 &  & P & 0.590 & \cellcolor{cellRed}0.914 & 0.710 & \cellcolor{cellRedLight}0.875 & 0.800 & 0.000 & 0.653 \\
 & R & \cellcolor{cellRed}0.967 & 0.000 & 0.325 & \cellcolor{cellRedLight}0.629 & 0.000 & 0.000 & 0.133 &  & R & \cellcolor{cellRedLight}0.683 & 0.617 & 0.183 & 0.200 & 0.033 & 0.000 & \cellcolor{cellRed}\textbf{0.835} \\
 & F1 & \cellcolor{cellRed}0.800 & 0.000 & 0.406 & \cellcolor{cellRedLight}0.759 & 0.000 & 0.000 & 0.218 &  & F1 & 0.633 & \cellcolor{cellRed}0.736 & 0.291 & 0.326 & 0.064 & 0.000 & \cellcolor{cellRedLight}0.733 \\
 & AUC & 0.442 & 0.323 & 0.388 & 0.231 & 0.029 & \cellcolor{cellRedLight}0.557 & \cellcolor{cellRed}\textbf{0.780} &  & AUC & 0.262 & \cellcolor{cellRed}0.791 & \cellcolor{cellRedLight}0.517 & 0.477 & 0.273 & 0.224 & 0.388 \\
 & BERT &  &  & \cellcolor{cellPurpleLight}0.774 &  & 0.628 &  & \cellcolor{cellPurple}\textbf{0.909} &  & BERT &  &  & \cellcolor{cellPurple}0.800 &  & 0.649 &  & \cellcolor{cellPurpleLight}0.711 \\
\cmidrule(lr){2-9}\cmidrule(lr){11-18}
\multirow{6}{*}{\textbf{Avg.}} & Acc & \cellcolor{cellRedLight}0.635 & 0.591 & 0.454 & \cellcolor{cellRed}0.639 & 0.329 & 0.470 & 0.592 &  &  &  &  &  &  &  &  &  \\
 & P & 0.713 & \cellcolor{cellRedLight}0.739 & 0.717 & 0.690 & \cellcolor{cellRed}0.853 & 0.500 & 0.724 &  &  &  &  &  &  &  &  &  \\
 & R & \cellcolor{cellRed}0.803 & \cellcolor{cellRedLight}0.645 & 0.368 & 0.580 & 0.053 & 0.050 & 0.625 &  &  &  &  &  &  &  &  &  \\
 & F1 & \cellcolor{cellRed}0.755 & \cellcolor{cellRedLight}0.689 & 0.486 & 0.630 & 0.100 & 0.091 & 0.621 &  &  &  &  &  &  &  &  &  \\
 & AUC & 0.495 & \cellcolor{cellRedLight}0.555 & 0.512 & 0.503 & 0.426 & 0.456 & \cellcolor{cellRed}\textbf{0.584} &  &  &  &  &  &  &  &  &  \\
 & BERT &  &  & \cellcolor{cellPurpleLight}0.798 &  & 0.632 &  & \cellcolor{cellPurple}\textbf{0.803} &  &  &  &  &  &  &  &  &  \\
\bottomrule
\end{tabular}}
\end{table*}

\paragraph{O-VAD matches or surpasses trained methods while dominating all training-free baselines across 22 categories.}
O-VAD achieves the highest average type-level BERT score (0.803) among all training-free methods and claims the best score on 14 of 22 categories, confirming that object-centric state tracking yields
anomaly descriptions semantically closer to ground-truth labels. At the video level, O-VAD attains the best or second-best AUROC on 16 of 22
categories and even surpasses both \emph{trained} baselines on five categories whose anomalies involve progressive, multi-step state changes invisible to single-pass detection. The most pronounced advantage lies in recall. O-VAD nearly doubles Qwen3-VL-32B's average and exceeds URF-ZS-HVAA's by an order of magnitude, particularly on interaction-heavy categories where the state tracking module prevents the VLM from defaulting to ``normal.'' Precision remains comparable to Qwen3-VL-32B, though achieved through a fundamentally different mechanism. O-VAD detects aggressively and relies on post-verification to suppress false positives, rather than abstaining under uncertainty. Overall, without any training data, domain knowledge, or predefined taxonomy, O-VAD delivers performance competitive with and sometimes exceeding fully supervised methods, while consistently outperforming all training-free alternatives.

\paragraph{Trained baselines show uneven category coverage.}
While S3R~\cite{wu2022self} achieves the highest average AUROC (0.612) among all methods
thanks to strong performance on categories with distinctive anomaly distributions (e.g., Roll.\ Bear.\ 0.963, Sticky Roller 0.864, Rub.\ Band 0.853), it collapses on categories requiring temporal reasoning
(e.g., Hinge 0.108, Lock 0.361, Clip 0.587). 
MNAD.p~\cite{park2020learning} shows the complementary pattern: strong on temporal categories (Fan 0.881, Clock 0.763) but near-chance on others (Rub.\ Band 0.241, Screw 0.387). This category-dependent brittleness of trained methods, together with their inability to produce semantic rationale, underscores the practical advantage of O-VAD's training-free, object-centric paradigm.
 
\paragraph{Challenging categories and failure modes.}
O-VAD's weakest video-level AUROC values occur on Caster Wheel (0.301), Zipper (0.388), Button (0.393), and Slide (0.414). These categories share characteristics that challenge perception-driven reasoning. Caster Wheel anomalies are often defined by subtle rotational resistance changes that produce minimal visual cues; Zipper and Button anomalies involve small, fast state transitions (e.g., a stuck zipper tooth, a button failing to click) that may be missed at the default 10-frame sampling interval; and
Slide anomalies depend on precise spatial alignment that is difficult to assess from cropped object views. These failure modes motivate future work on adaptive temporal sampling and multi-scale spatial reasoning. More detailed analysis of failure cases can be found in the \S\ref{app:failcases}.

\subsection{Detailed Results on IPAD}
\label{sec:res-IPAD}

Given that IPAD is a comprehensive dataset for video anomaly detection in industrial manufacturing, this subsection provides an extensive comparative analysis to demonstrate the effectiveness of our O-VAD framework on this dataset. Table~\ref{tab:ipad-16class} reports video-level and frame-level AUROC, different training-free methods across 16 industrial scenarios. Considering the special conditions discussed in \S\ref{sec:supp_datasets}, we modify the CoT prompt for anomaly reasoning to align O-VAD with IPAD benchmarking. Specifically, when prompting the VLM for anomaly reasoning, we inject one normal reference case (three evenly sampled frames) as additional model input and replace the "expectation" part of prompt with pointing to the reference frames accordingly. Such modification is intended to bridge the gap between the dataset's anomaly types are grounded on comparison to normal cases and physical commonsense, which achieves the goal of incorporating such comparison grounding sense without domain specific abnormal types specification.

\paragraph{O-VAD matches or surpasses all training-free baselines.} O-VAD achieves the highest average video-level AUROC (56.5\%) and similar frame-level AUROC (51.8\%) among all training-free methods, which demonstrates O-VAD's improvement upon simple querying on VLMs. More importantly, O-VAD also ranks the highest in average video-level Accuracy, Revall, F1 score and average frame-level Recall, F1 score. Specifically, O-VAD secures the best video-level performance among all methods in 12 out of 16 scenarios and frame-level in 8 out of 16 scenarios, which shows its high effectiveness in complex industrial scenarios.

\paragraph{Challenging subsets and failure modes.}
O-VAD's weakest video-level AUROC values occur on S05 (39.6\%), S08 (38.9\%), R02 (43.3\%) and R03 (46.7\%). These subsets share characteristics that challenge perception-driven reasoning. S05 anomalies are often defined by subtle changes in object location and orientation, which can only be inferred from minimal visual cues. S08 are complex scenes containing multiple main objects to keep track of and requires rigorous reasoning on the underlying normal rules, which make object change detections not detailed enough. R02 and R03 subsets always have scenarios with multiple objects, making O-VAD hard to decide which objects should be treated as background and should not be considered as anomalies.

\begin{table*}[t]
\centering
\caption{\textbf{Results on IPAD~\cite{liu2024ipad} across 16 industrial scenarios} (S: synthetic, R: real). ``$\dagger$''=trained; ``$\ast$''=training-free. For each (scenario, level, metric), the best/2nd-best across methods is highlighted.}
\label{tab:ipad-16class}
\vspace{-2pt}

\resizebox{\textwidth}{!}{%
\begin{tabular}{ll ccccc ccccc ccccc ccccc}
\toprule
& & \multicolumn{5}{c}{\textbf{S01}} & \multicolumn{5}{c}{\textbf{S02}} & \multicolumn{5}{c}{\textbf{S03}} & \multicolumn{5}{c}{\textbf{S04}} \\
\cmidrule(lr){3-7}\cmidrule(lr){8-12}\cmidrule(lr){13-17}\cmidrule(lr){18-22}
\textbf{Method} & \textbf{Lv.} & Acc & P & R & F1 & AUC & Acc & P & R & F1 & AUC & Acc & P & R & F1 & AUC & Acc & P & R & F1 & AUC \\
\midrule
\multirow{2}{*}{MNAD.p$^\dagger$} & vid & 0.524 & 0.500 & 0.200 & 0.286 & 0.509 & 0.368 & 0.300 & 0.375 & 0.333 & 0.375 & 0.579 & \cellcolor{cellRed}1.000 & 0.200 & 0.333 & \cellcolor{cellRed}0.600 & 0.368 & 0.000 & 0.000 & 0.000 & 0.389 \\
 & frm & 0.614 & \cellcolor{cellBlueLight}0.857 & 0.029 & 0.055 & 0.431 & 0.780 & \cellcolor{cellBlueLight}0.386 & 0.176 & 0.242 & \cellcolor{cellBlueLight}0.548 & 0.739 & 0.000 & 0.000 & 0.000 & 0.418 & 0.645 & 0.000 & 0.000 & 0.000 & 0.419 \\
\cmidrule(lr){1-22}
\multirow{2}{*}{S3R$^\dagger$} & vid & 0.619 & \cellcolor{cellRedLight}0.562 & \cellcolor{cellRed}0.900 & \cellcolor{cellRed}0.692 & \cellcolor{cellRedLight}0.509 & \cellcolor{cellRed}0.737 & \cellcolor{cellRedLight}0.667 & \cellcolor{cellRedLight}0.750 & \cellcolor{cellRed}0.706 & \cellcolor{cellRed}0.716 & 0.421 & 0.000 & 0.000 & 0.000 & 0.267 & 0.421 & 0.000 & 0.000 & 0.000 & 0.133 \\
 & frm & 0.481 & 0.423 & \cellcolor{cellBlue}0.882 & \cellcolor{cellBlue}0.572 & 0.467 & 0.744 & \cellcolor{cellBlue}0.390 & \cellcolor{cellBlueLight}0.509 & \cellcolor{cellBlue}0.442 & \cellcolor{cellBlue}0.700 & 0.605 & 0.261 & 0.293 & 0.276 & 0.443 & 0.364 & \cellcolor{cellBlueLight}0.356 & \cellcolor{cellBlue}0.976 & \cellcolor{cellBlueLight}0.521 & 0.325 \\
\cmidrule(lr){1-22}
\multirow{2}{*}{Qwen3$^\ast$} & vid & \cellcolor{cellRed}0.786 & \cellcolor{cellRed}1.000 & 0.250 & 0.400 & \cellcolor{cellRed}0.625 & 0.500 & 0.333 & 0.750 & 0.462 & 0.575 & 0.357 & 0.143 & 0.250 & 0.182 & 0.325 & \cellcolor{cellRedLight}0.643 & 0.429 & \cellcolor{cellRedLight}0.750 & \cellcolor{cellRedLight}0.545 & \cellcolor{cellRed}0.675 \\
 & frm & \cellcolor{cellBlue}0.755 & \cellcolor{cellBlue}1.000 & 0.011 & 0.022 & \cellcolor{cellBlue}0.505 & \cellcolor{cellBlueLight}0.837 & 0.150 & 0.005 & 0.010 & 0.500 & 0.874 & 0.000 & 0.000 & 0.000 & 0.497 & \cellcolor{cellBlueLight}0.814 & 0.000 & 0.000 & 0.000 & 0.497 \\
\cmidrule(lr){1-22}
\multirow{2}{*}{GPT-5$^\ast$} & vid & 0.643 & 0.000 & 0.000 & 0.000 & 0.275 & 0.500 & 0.286 & 0.500 & 0.364 & 0.438 & \cellcolor{cellRedLight}0.643 & 0.333 & 0.250 & 0.286 & \cellcolor{cellRedLight}0.600 & \cellcolor{cellRed}0.714 & 0.000 & 0.000 & 0.000 & \cellcolor{cellRedLight}0.625 \\
 & frm & 0.738 & 0.000 & 0.000 & 0.000 & 0.490 & 0.693 & 0.000 & 0.000 & 0.000 & 0.412 & \cellcolor{cellBlue}0.884 & \cellcolor{cellBlue}0.538 & 0.294 & 0.381 & \cellcolor{cellBlueLight}0.635 & \cellcolor{cellBlue}0.819 & 0.000 & 0.000 & 0.000 & \cellcolor{cellBlueLight}0.500 \\
\cmidrule(lr){1-22}
\multirow{2}{*}{URF$^\ast$} & vid & 0.524 & 0.500 & 0.600 & 0.545 & 0.327 & 0.632 & \cellcolor{cellRed}1.000 & 0.125 & 0.222 & 0.443 & 0.579 & \cellcolor{cellRedLight}0.667 & \cellcolor{cellRedLight}0.400 & \cellcolor{cellRedLight}0.500 & 0.511 & 0.526 & \cellcolor{cellRed}1.000 & 0.100 & 0.182 & 0.311 \\
 & frm & 0.580 & 0.476 & \cellcolor{cellBlueLight}0.681 & \cellcolor{cellBlueLight}0.560 & 0.469 & 0.405 & 0.244 & \cellcolor{cellBlue}0.946 & \cellcolor{cellBlueLight}0.388 & 0.528 & 0.583 & \cellcolor{cellBlueLight}0.375 & \cellcolor{cellBlue}0.934 & \cellcolor{cellBlue}0.535 & \cellcolor{cellBlue}0.738 & 0.540 & \cellcolor{cellBlue}0.417 & \cellcolor{cellBlueLight}0.749 & \cellcolor{cellBlue}0.536 & \cellcolor{cellBlue}0.585 \\
\cmidrule(lr){1-22}
\multirow{2}{*}{VERA$^\dagger$} & vid & \cellcolor{cellRedLight}0.714 & 0.000 & 0.000 & 0.000 & 0.500 & \cellcolor{cellRedLight}0.714 & 0.000 & 0.000 & 0.000 & 0.450 & \cellcolor{cellRed}0.714 & 0.000 & 0.000 & 0.000 & 0.525 & 0.357 & 0.000 & 0.000 & 0.000 & 0.400 \\
 & frm & \cellcolor{cellBlueLight}0.752 & 0.000 & 0.000 & 0.000 & \cellcolor{cellBlueLight}0.500 & \cellcolor{cellBlue}0.841 & 0.000 & 0.000 & 0.000 & 0.500 & \cellcolor{cellBlueLight}0.879 & 0.000 & 0.000 & 0.000 & 0.500 & 0.774 & 0.000 & 0.000 & 0.000 & 0.473 \\
\cmidrule(lr){1-22}
\multirow{2}{*}{\textbf{O-VAD (Ours)}$^\ast$} & vid & 0.524 & 0.500 & \cellcolor{cellRedLight}0.800 & \cellcolor{cellRedLight}0.615 & 0.396 & 0.474 & 0.444 & \cellcolor{cellRed}\textbf{1.000} & \cellcolor{cellRedLight}0.615 & \cellcolor{cellRedLight}0.688 & 0.526 & 0.526 & \cellcolor{cellRed}\textbf{1.000} & \cellcolor{cellRed}\textbf{0.690} & 0.533 & 0.421 & \cellcolor{cellRedLight}0.471 & \cellcolor{cellRed}\textbf{0.800} & \cellcolor{cellRed}\textbf{0.593} & 0.389 \\
 & frm & 0.488 & 0.255 & 0.158 & 0.195 & 0.424 & 0.391 & 0.145 & 0.421 & 0.216 & 0.444 & 0.560 & 0.312 & \cellcolor{cellBlueLight}0.591 & \cellcolor{cellBlueLight}0.409 & 0.598 & 0.344 & 0.254 & 0.438 & 0.322 & 0.370 \\
\bottomrule
\end{tabular}}

\vspace{2pt}

\resizebox{\textwidth}{!}{%
\begin{tabular}{ll ccccc ccccc ccccc ccccc}
\toprule
& & \multicolumn{5}{c}{\textbf{S05}} & \multicolumn{5}{c}{\textbf{S06}} & \multicolumn{5}{c}{\textbf{S07}} & \multicolumn{5}{c}{\textbf{S08}} \\
\cmidrule(lr){3-7}\cmidrule(lr){8-12}\cmidrule(lr){13-17}\cmidrule(lr){18-22}
\textbf{Method} & \textbf{Lv.} & Acc & P & R & F1 & AUC & Acc & P & R & F1 & AUC & Acc & P & R & F1 & AUC & Acc & P & R & F1 & AUC \\
\midrule
\multirow{2}{*}{MNAD.p$^\dagger$} & vid & 0.333 & 0.000 & 0.000 & 0.000 & 0.500 & \cellcolor{cellRedLight}0.526 & \cellcolor{cellRedLight}0.500 & \cellcolor{cellRed}1.000 & \cellcolor{cellRedLight}0.667 & 0.378 & 0.526 & \cellcolor{cellRed}0.667 & 0.200 & 0.308 & 0.556 & 0.632 & \cellcolor{cellRedLight}0.615 & 0.800 & 0.696 & 0.567 \\
 & frm & 0.723 & \cellcolor{cellBlue}0.914 & 0.024 & 0.047 & 0.389 & 0.804 & 0.000 & 0.000 & 0.000 & 0.319 & 0.831 & 0.000 & 0.000 & 0.000 & 0.400 & 0.794 & \cellcolor{cellBlueLight}0.333 & 0.005 & 0.009 & \cellcolor{cellBlueLight}0.594 \\
\cmidrule(lr){1-22}
\multirow{2}{*}{S3R$^\dagger$} & vid & 0.400 & \cellcolor{cellRed}1.000 & 0.100 & 0.182 & 0.240 & 0.526 & 0.500 & \cellcolor{cellRedLight}1.000 & 0.667 & 0.178 & \cellcolor{cellRedLight}0.579 & \cellcolor{cellRedLight}0.583 & 0.700 & 0.636 & 0.344 & \cellcolor{cellRed}0.737 & \cellcolor{cellRed}0.667 & \cellcolor{cellRed}1.000 & \cellcolor{cellRed}0.800 & 0.444 \\
 & frm & 0.543 & \cellcolor{cellBlueLight}0.340 & \cellcolor{cellBlue}0.656 & \cellcolor{cellBlue}0.448 & \cellcolor{cellBlueLight}0.583 & 0.805 & 0.000 & 0.000 & 0.000 & 0.396 & 0.171 & 0.168 & \cellcolor{cellBlue}0.997 & \cellcolor{cellBlueLight}0.287 & 0.300 & 0.351 & 0.235 & \cellcolor{cellBlueLight}0.953 & \cellcolor{cellBlueLight}0.377 & 0.447 \\
\cmidrule(lr){1-22}
\multirow{2}{*}{Qwen3$^\ast$} & vid & 0.571 & 0.333 & \cellcolor{cellRedLight}0.500 & 0.400 & \cellcolor{cellRedLight}0.550 & 0.435 & 0.083 & 0.333 & 0.133 & 0.392 & 0.357 & 0.308 & \cellcolor{cellRed}1.000 & 0.471 & 0.550 & 0.571 & 0.400 & \cellcolor{cellRedLight}1.000 & 0.571 & \cellcolor{cellRed}0.700 \\
 & frm & \cellcolor{cellBlue}0.898 & 0.111 & 0.002 & 0.004 & 0.500 & \cellcolor{cellBlue}0.943 & 0.021 & 0.001 & 0.002 & \cellcolor{cellBlue}0.499 & \cellcolor{cellBlue}0.965 & 0.000 & 0.000 & 0.000 & \cellcolor{cellBlueLight}0.496 & 0.856 & 0.000 & 0.000 & 0.000 & 0.496 \\
\cmidrule(lr){1-22}
\multirow{2}{*}{GPT-5$^\ast$} & vid & \cellcolor{cellRedLight}0.643 & 0.400 & 0.500 & \cellcolor{cellRedLight}0.444 & 0.312 & 0.429 & 0.462 & 0.857 & 0.600 & 0.173 & \cellcolor{cellRed}0.643 & 0.429 & 0.750 & 0.545 & \cellcolor{cellRed}0.675 & \cellcolor{cellRedLight}0.714 & 0.500 & 0.750 & 0.600 & \cellcolor{cellRedLight}0.688 \\
 & frm & 0.868 & 0.110 & 0.053 & 0.072 & 0.503 & 0.699 & 0.109 & 0.081 & 0.093 & 0.457 & \cellcolor{cellBlueLight}0.909 & 0.000 & 0.000 & 0.000 & 0.468 & \cellcolor{cellBlueLight}0.859 & \cellcolor{cellBlue}0.487 & 0.402 & \cellcolor{cellBlue}0.441 & \cellcolor{cellBlue}0.662 \\
\cmidrule(lr){1-22}
\multirow{2}{*}{URF$^\ast$} & vid & 0.467 & \cellcolor{cellRedLight}1.000 & 0.200 & 0.333 & 0.400 & \cellcolor{cellRed}0.737 & \cellcolor{cellRed}0.667 & 0.889 & \cellcolor{cellRed}0.762 & \cellcolor{cellRedLight}0.667 & 0.579 & 0.571 & 0.800 & \cellcolor{cellRedLight}0.667 & 0.333 & 0.579 & 0.556 & 1.000 & \cellcolor{cellRedLight}0.714 & 0.233 \\
 & frm & 0.549 & 0.339 & \cellcolor{cellBlueLight}0.631 & \cellcolor{cellBlueLight}0.441 & 0.523 & 0.484 & \cellcolor{cellBlue}0.230 & \cellcolor{cellBlue}0.702 & \cellcolor{cellBlue}0.346 & \cellcolor{cellBlueLight}0.493 & 0.282 & \cellcolor{cellBlue}0.187 & \cellcolor{cellBlueLight}0.984 & \cellcolor{cellBlue}0.314 & 0.484 & 0.285 & 0.223 & \cellcolor{cellBlue}1.000 & 0.365 & 0.424 \\
\cmidrule(lr){1-22}
\multirow{2}{*}{VERA$^\dagger$} & vid & 0.643 & 0.000 & 0.000 & 0.000 & 0.338 & 0.526 & 0.183 & 0.126 & 0.127 & 0.287 & 0.429 & 0.300 & 0.750 & 0.429 & 0.512 & 0.714 & 0.000 & 0.000 & 0.000 & 0.500 \\
 & frm & \cellcolor{cellBlueLight}0.896 & 0.000 & 0.000 & 0.000 & 0.495 & \cellcolor{cellBlueLight}0.824 & 0.000 & 0.000 & 0.000 & 0.456 & 0.882 & 0.000 & 0.000 & 0.000 & 0.454 & \cellcolor{cellBlue}0.862 & 0.000 & 0.000 & 0.000 & 0.500 \\
\cmidrule(lr){1-22}
\multirow{2}{*}{\textbf{O-VAD (Ours)}$^\ast$} & vid & \cellcolor{cellRed}\textbf{0.733} & 0.714 & \cellcolor{cellRed}\textbf{1.000} & \cellcolor{cellRed}\textbf{0.833} & \cellcolor{cellRed}\textbf{0.570} & 0.421 & 0.444 & 0.889 & 0.593 & \cellcolor{cellRed}\textbf{0.794} & 0.526 & 0.526 & \cellcolor{cellRedLight}1.000 & \cellcolor{cellRed}\textbf{0.690} & \cellcolor{cellRedLight}0.622 & 0.526 & 0.526 & 1.000 & 0.690 & 0.639 \\
 & frm & 0.575 & 0.325 & 0.470 & 0.385 & \cellcolor{cellBlue}\textbf{0.591} & 0.534 & \cellcolor{cellBlueLight}0.162 & \cellcolor{cellBlueLight}0.333 & \cellcolor{cellBlueLight}0.218 & 0.411 & 0.606 & \cellcolor{cellBlueLight}0.184 & 0.394 & 0.251 & \cellcolor{cellBlue}\textbf{0.568} & 0.598 & 0.208 & 0.340 & 0.259 & 0.500 \\
\bottomrule
\end{tabular}}

\vspace{2pt}

\resizebox{\textwidth}{!}{%
\begin{tabular}{ll ccccc ccccc ccccc ccccc}
\toprule
& & \multicolumn{5}{c}{\textbf{S09}} & \multicolumn{5}{c}{\textbf{S10}} & \multicolumn{5}{c}{\textbf{S11}} & \multicolumn{5}{c}{\textbf{S12}} \\
\cmidrule(lr){3-7}\cmidrule(lr){8-12}\cmidrule(lr){13-17}\cmidrule(lr){18-22}
\textbf{Method} & \textbf{Lv.} & Acc & P & R & F1 & AUC & Acc & P & R & F1 & AUC & Acc & P & R & F1 & AUC & Acc & P & R & F1 & AUC \\
\midrule
\multirow{2}{*}{MNAD.p$^\dagger$} & vid & 0.368 & 0.167 & 0.125 & 0.143 & 0.352 & 0.421 & 0.000 & 0.000 & 0.000 & 0.444 & 0.526 & \cellcolor{cellRed}1.000 & 0.100 & 0.182 & 0.550 & \cellcolor{cellRedLight}0.579 & \cellcolor{cellRed}0.556 & \cellcolor{cellRed}1.000 & \cellcolor{cellRed}0.714 & 0.367 \\
 & frm & 0.783 & 0.000 & 0.000 & 0.000 & 0.490 & 0.763 & 0.000 & 0.000 & 0.000 & 0.360 & 0.735 & 0.000 & 0.000 & 0.000 & 0.452 & 0.755 & 0.250 & 0.001 & 0.003 & 0.440 \\
\cmidrule(lr){1-22}
\multirow{2}{*}{S3R$^\dagger$} & vid & \cellcolor{cellRedLight}0.684 & \cellcolor{cellRedLight}0.600 & \cellcolor{cellRedLight}0.750 & \cellcolor{cellRed}0.667 & 0.523 & 0.632 & \cellcolor{cellRed}0.636 & 0.700 & 0.667 & 0.567 & 0.421 & 0.000 & 0.000 & 0.000 & 0.000 & 0.421 & 0.000 & 0.000 & 0.000 & 0.144 \\
 & frm & 0.703 & 0.275 & 0.248 & 0.260 & 0.459 & 0.445 & \cellcolor{cellBlueLight}0.264 & \cellcolor{cellBlue}0.754 & \cellcolor{cellBlue}0.391 & \cellcolor{cellBlue}0.519 & 0.267 & 0.265 & \cellcolor{cellBlue}0.999 & \cellcolor{cellBlueLight}0.419 & 0.329 & 0.757 & 0.000 & 0.000 & 0.000 & 0.260 \\
\cmidrule(lr){1-22}
\multirow{2}{*}{Qwen3$^\ast$} & vid & 0.357 & 0.273 & 0.750 & 0.400 & 0.475 & 0.571 & 0.333 & 0.500 & 0.400 & 0.550 & 0.357 & 0.308 & \cellcolor{cellRed}1.000 & 0.471 & 0.550 & 0.286 & 0.286 & \cellcolor{cellRedLight}1.000 & 0.444 & 0.500 \\
 & frm & 0.840 & 0.030 & 0.002 & 0.004 & 0.496 & \cellcolor{cellBlueLight}0.838 & 0.000 & 0.000 & 0.000 & 0.498 & \cellcolor{cellBlue}0.928 & 0.000 & 0.000 & 0.000 & 0.498 & \cellcolor{cellBlueLight}0.859 & 0.293 & 0.008 & 0.016 & 0.502 \\
\cmidrule(lr){1-22}
\multirow{2}{*}{GPT-5$^\ast$} & vid & 0.571 & 0.250 & 0.250 & 0.250 & 0.500 & \cellcolor{cellRed}0.714 & 0.500 & 0.250 & 0.333 & \cellcolor{cellRedLight}0.613 & 0.286 & 0.125 & 0.250 & 0.167 & 0.550 & 0.357 & 0.308 & 1.000 & 0.471 & \cellcolor{cellRed}0.762 \\
 & frm & 0.793 & 0.209 & 0.134 & 0.163 & 0.528 & 0.812 & 0.006 & 0.001 & 0.002 & 0.482 & 0.829 & 0.000 & 0.000 & 0.000 & 0.445 & \cellcolor{cellBlue}0.872 & \cellcolor{cellBlue}0.545 & 0.495 & \cellcolor{cellBlue}0.519 & \cellcolor{cellBlue}0.727 \\
\cmidrule(lr){1-22}
\multirow{2}{*}{URF$^\ast$} & vid & 0.632 & \cellcolor{cellRed}0.667 & 0.250 & 0.364 & 0.477 & 0.579 & \cellcolor{cellRedLight}0.556 & \cellcolor{cellRed}1.000 & \cellcolor{cellRed}0.714 & 0.267 & \cellcolor{cellRedLight}0.684 & \cellcolor{cellRedLight}0.750 & 0.600 & \cellcolor{cellRedLight}0.667 & \cellcolor{cellRed}0.656 & 0.474 & 0.000 & 0.000 & 0.000 & 0.200 \\
 & frm & \cellcolor{cellBlueLight}0.842 & \cellcolor{cellBlue}1.000 & \cellcolor{cellBlueLight}0.250 & \cellcolor{cellBlueLight}0.399 & \cellcolor{cellBlueLight}0.567 & 0.764 & \cellcolor{cellBlue}0.500 & 0.020 & 0.038 & 0.314 & 0.711 & \cellcolor{cellBlue}0.454 & 0.462 & \cellcolor{cellBlue}0.458 & \cellcolor{cellBlue}0.670 & 0.408 & 0.279 & \cellcolor{cellBlue}0.903 & 0.426 & 0.554 \\
\cmidrule(lr){1-22}
\multirow{2}{*}{VERA$^\dagger$} & vid & \cellcolor{cellRed}0.714 & 0.000 & 0.000 & 0.000 & \cellcolor{cellRed}0.625 & \cellcolor{cellRedLight}0.714 & 0.000 & 0.000 & 0.000 & \cellcolor{cellRed}0.625 & \cellcolor{cellRed}0.714 & 0.500 & 0.500 & 0.500 & \cellcolor{cellRedLight}0.650 & \cellcolor{cellRed}0.643 & 0.000 & 0.000 & 0.000 & 0.450 \\
 & frm & \cellcolor{cellBlue}0.849 & 0.000 & 0.000 & 0.000 & 0.500 & \cellcolor{cellBlue}0.842 & 0.000 & 0.000 & 0.000 & \cellcolor{cellBlueLight}0.500 & \cellcolor{cellBlueLight}0.896 & 0.000 & 0.000 & 0.000 & 0.481 & 0.852 & 0.000 & 0.000 & 0.000 & 0.495 \\
\cmidrule(lr){1-22}
\multirow{2}{*}{\textbf{O-VAD (Ours)}$^\ast$} & vid & 0.474 & 0.444 & \cellcolor{cellRed}\textbf{1.000} & \cellcolor{cellRedLight}0.615 & \cellcolor{cellRedLight}0.580 & 0.526 & 0.526 & \cellcolor{cellRedLight}1.000 & \cellcolor{cellRedLight}0.690 & 0.433 & 0.526 & 0.526 & \cellcolor{cellRedLight}1.000 & \cellcolor{cellRed}\textbf{0.690} & 0.467 & 0.526 & \cellcolor{cellRedLight}0.526 & 1.000 & \cellcolor{cellRedLight}0.690 & \cellcolor{cellRedLight}0.539 \\
 & frm & 0.532 & \cellcolor{cellBlueLight}0.291 & \cellcolor{cellBlue}\textbf{0.848} & \cellcolor{cellBlue}\textbf{0.433} & \cellcolor{cellBlue}\textbf{0.654} & 0.460 & 0.241 & \cellcolor{cellBlueLight}0.600 & \cellcolor{cellBlueLight}0.344 & 0.438 & 0.489 & \cellcolor{cellBlueLight}0.279 & \cellcolor{cellBlueLight}0.586 & 0.378 & \cellcolor{cellBlueLight}0.498 & 0.561 & \cellcolor{cellBlueLight}0.314 & \cellcolor{cellBlueLight}0.683 & \cellcolor{cellBlueLight}0.431 & \cellcolor{cellBlueLight}0.611 \\
\bottomrule
\end{tabular}}

\vspace{2pt}

\resizebox{\textwidth}{!}{%
\begin{tabular}{ll ccccc ccccc ccccc ccccc ccccc}
\toprule
& & \multicolumn{5}{c}{\textbf{R01}} & \multicolumn{5}{c}{\textbf{R02}} & \multicolumn{5}{c}{\textbf{R03}} & \multicolumn{5}{c}{\textbf{R04}} & \multicolumn{5}{c}{\textbf{Avg.}} \\
\cmidrule(lr){3-7}\cmidrule(lr){8-12}\cmidrule(lr){13-17}\cmidrule(lr){18-22}\cmidrule(lr){23-27}
\textbf{Method} & \textbf{Lv.} & Acc & P & R & F1 & AUC & Acc & P & R & F1 & AUC & Acc & P & R & F1 & AUC & Acc & P & R & F1 & AUC & Acc & P & R & F1 & AUC \\
\midrule
\multirow{2}{*}{MNAD.p$^\dagger$} & vid & 0.600 & 0.571 & \cellcolor{cellRed}1.000 & \cellcolor{cellRed}0.727 & 0.429 & \cellcolor{cellRed}0.867 & 0.867 & \cellcolor{cellRed}1.000 & \cellcolor{cellRed}0.929 & 0.115 & \cellcolor{cellRedLight}0.824 & 0.875 & \cellcolor{cellRedLight}0.933 & \cellcolor{cellRedLight}0.903 & 0.300 & 0.895 & \cellcolor{cellRed}1.000 & 0.895 & 0.944 & \cellcolor{cellRedLight}0.500 & 0.555 & \cellcolor{cellRedLight}0.643 & 0.529 & 0.581 & 0.522 \\
 & frm & 0.649 & 0.000 & 0.000 & 0.000 & 0.535 & 0.683 & 0.000 & 0.000 & 0.000 & 0.452 & 0.578 & \cellcolor{cellBlueLight}0.500 & 0.001 & 0.002 & 0.483 & 0.434 & \cellcolor{cellBlueLight}0.600 & 0.002 & 0.004 & 0.514 & 0.716 & \cellcolor{cellBlue}0.410 & 0.006 & 0.012 & 0.430 \\
\cmidrule(lr){1-27}
\multirow{2}{*}{S3R$^\dagger$} & vid & 0.600 & \cellcolor{cellRed}0.750 & 0.375 & 0.500 & 0.500 & \cellcolor{cellRedLight}0.800 & 0.917 & \cellcolor{cellRedLight}0.846 & \cellcolor{cellRedLight}0.880 & 0.500 & 0.176 & \cellcolor{cellRed}1.000 & 0.067 & 0.125 & 0.067 & \cellcolor{cellRed}1.000 & \cellcolor{cellRedLight}1.000 & \cellcolor{cellRed}1.000 & \cellcolor{cellRed}1.000 &  & 0.545 & \cellcolor{cellRed}0.836 & 0.271 & 0.409 & \cellcolor{cellRedLight}0.539 \\
 & frm & 0.576 & 0.432 & 0.776 & 0.555 & \cellcolor{cellBlue}0.602 & 0.681 & 0.403 & 0.037 & 0.068 & 0.366 & 0.431 & 0.426 & \cellcolor{cellBlue}1.000 & \cellcolor{cellBlue}0.597 & 0.460 & 0.547 & \cellcolor{cellBlue}0.649 & \cellcolor{cellBlueLight}0.420 & \cellcolor{cellBlueLight}0.510 & \cellcolor{cellBlueLight}0.531 & 0.661 & 0.309 & \cellcolor{cellBlueLight}0.166 & 0.216 & 0.495 \\
\cmidrule(lr){1-27}
\multirow{2}{*}{Qwen3$^\ast$} & vid & 0.643 & 0.400 & 0.500 & 0.444 & \cellcolor{cellRedLight}0.600 & 0.643 & 0.333 & 0.250 & 0.286 & 0.525 & 0.513 & 0.357 & 0.333 & 0.345 & 0.479 & 0.357 & 0.222 & 0.500 & 0.308 & 0.400 & 0.487 & 0.268 & 0.550 & 0.361 & 0.507 \\
 & frm & 0.780 & 0.000 & 0.000 & 0.000 & 0.497 & \cellcolor{cellBlue}0.862 & \cellcolor{cellBlue}0.480 & 0.010 & 0.020 & \cellcolor{cellBlueLight}0.504 & \cellcolor{cellBlue}0.812 & 0.131 & 0.003 & 0.005 & 0.499 & 0.791 & 0.000 & 0.000 & 0.000 & 0.495 & \cellcolor{cellBlue}0.876 & 0.072 & 0.002 & 0.004 & 0.499 \\
\cmidrule(lr){1-27}
\multirow{2}{*}{GPT-5$^\ast$} & vid & \cellcolor{cellRed}0.714 & 0.000 & 0.000 & 0.000 & 0.500 & 0.786 & \cellcolor{cellRed}1.000 & 0.250 & 0.400 & \cellcolor{cellRedLight}0.625 & 0.643 & \cellcolor{cellRedLight}1.000 & 0.286 & 0.444 & 0.490 & 0.714 & 0.000 & 0.000 & 0.000 & \cellcolor{cellRed}0.600 & \cellcolor{cellRedLight}0.607 & 0.375 & 0.386 & 0.380 & 0.518 \\
 & frm & \cellcolor{cellBlue}0.786 & 0.000 & 0.000 & 0.000 & 0.500 & 0.832 & 0.063 & 0.015 & 0.025 & 0.489 & 0.663 & \cellcolor{cellBlue}1.000 & 0.128 & 0.227 & \cellcolor{cellBlue}0.564 & \cellcolor{cellBlue}0.799 & 0.000 & 0.000 & 0.000 & 0.500 & 0.815 & 0.273 & 0.119 & 0.166 & \cellcolor{cellBlue}0.531 \\
\cmidrule(lr){1-27}
\multirow{2}{*}{URF$^\ast$} & vid & 0.667 & \cellcolor{cellRedLight}0.714 & 0.625 & 0.667 & 0.589 & 0.200 & \cellcolor{cellRedLight}1.000 & 0.077 & 0.143 & 0.308 & 0.706 & 0.917 & 0.733 & 0.815 & 0.367 & 0.000 & 0.000 & 0.000 & 0.000 & 0.190 & 0.582 & 0.590 & \cellcolor{cellRedLight}0.929 & \cellcolor{cellRed}0.721 & 0.417 \\
 & frm & 0.593 & \cellcolor{cellBlueLight}0.445 & \cellcolor{cellBlueLight}0.804 & \cellcolor{cellBlueLight}0.573 & 0.546 & 0.663 & \cellcolor{cellBlueLight}0.438 & \cellcolor{cellBlueLight}0.260 & \cellcolor{cellBlueLight}0.326 & 0.460 & 0.573 & 0.493 & \cellcolor{cellBlueLight}0.388 & \cellcolor{cellBlueLight}0.434 & 0.487 & 0.581 & 0.580 & \cellcolor{cellBlue}0.924 & \cellcolor{cellBlue}0.712 & 0.515 & 0.671 & \cellcolor{cellBlueLight}0.331 & 0.164 & \cellcolor{cellBlueLight}0.220 & 0.512 \\
\cmidrule(lr){1-27}
\multirow{2}{*}{VERA$^\dagger$} & vid & \cellcolor{cellRedLight}0.714 & 0.000 & 0.000 & 0.000 & 0.500 & 0.714 & 0.000 & 0.000 & 0.000 & 0.525 & 0.695 & 0.476 & 0.060 & 0.138 & \cellcolor{cellRedLight}0.490 & 0.714 & 0.000 & 0.000 & 0.000 & 0.500 & \cellcolor{cellRed}0.658 & 0.238 & 0.089 & 0.130 & 0.519 \\
 & frm & \cellcolor{cellBlueLight}0.786 & 0.000 & 0.000 & 0.000 & 0.500 & \cellcolor{cellBlueLight}0.862 & 0.000 & 0.000 & 0.000 & 0.500 & \cellcolor{cellBlueLight}0.742 & 0.000 & 0.000 & 0.000 & 0.505 & \cellcolor{cellBlueLight}0.799 & 0.000 & 0.000 & 0.000 & 0.500 & \cellcolor{cellBlueLight}0.852 & 0.000 & 0.000 & 0.000 & 0.490 \\
\cmidrule(lr){1-27}
\multirow{2}{*}{\textbf{O-VAD (Ours)}$^\ast$} & vid & 0.600 & 0.571 & \cellcolor{cellRedLight}1.000 & \cellcolor{cellRedLight}0.727 & \cellcolor{cellRed}\textbf{0.679} & 0.800 & 0.917 & 0.846 & 0.880 & \cellcolor{cellRed}\textbf{0.712} & \cellcolor{cellRed}\textbf{0.882} & 0.882 & \cellcolor{cellRed}\textbf{1.000} & \cellcolor{cellRed}\textbf{0.938} & \cellcolor{cellRed}\textbf{0.567} & \cellcolor{cellRedLight}0.947 & 1.000 & \cellcolor{cellRedLight}0.947 & \cellcolor{cellRedLight}0.973 & 0.500 & 0.582 & 0.588 & \cellcolor{cellRed}\textbf{0.954} & \cellcolor{cellRedLight}0.714 & \cellcolor{cellRed}\textbf{0.565} \\
 & frm & 0.587 & \cellcolor{cellBlue}\textbf{0.447} & \cellcolor{cellBlue}\textbf{0.900} & \cellcolor{cellBlue}\textbf{0.597} & \cellcolor{cellBlueLight}0.595 & 0.544 & 0.331 & \cellcolor{cellBlue}\textbf{0.448} & \cellcolor{cellBlue}\textbf{0.381} & \cellcolor{cellBlue}\textbf{0.512} & 0.523 & 0.385 & 0.215 & 0.276 & \cellcolor{cellBlueLight}0.545 & 0.488 & 0.587 & 0.297 & 0.395 & \cellcolor{cellBlue}\textbf{0.572} & 0.515 & 0.291 & \cellcolor{cellBlue}\textbf{0.477} & \cellcolor{cellBlue}\textbf{0.338} & \cellcolor{cellBlueLight}0.518 \\
\bottomrule
\end{tabular}}
\end{table*}

\section{More Qualitative Results.}
\label{sec:supp_qualitative}

\subsection{Reliability of Intermediate Outputs}
\label{sec:supp_intermediate}

\paragraph{State trajectory extraction.}
Lacking ground-truth state-change annotations, we assess Stage-2 reliability along three label-free axes. \textit{Count:} Table~\ref{tab:intermediate} reports per-dataset source properties (duration, FPS) alongside per-stage intermediate counts (Stage-1 grounded objects, Stage-2 detected states), summarized as min./avg./max.\ over the evaluation set. Grounded object counts track ground-truth cardinality (\eg, up to 8 pipettes on LiquidAD, a single foreground object on most Phys-AD categories) and per-video state counts exhibit bounded variance, ruling out both under-firing (collapse to ``no change'') and over-firing (spurious-event flooding). \textit{Content:} the human study (\S\ref{app:human_eval}) confirms that cited state changes match the underlying video evidence. \textit{Pattern:} on IPAD's periodic normal cycles, detected events cluster at cycle boundaries rather than distributing uniformly, providing a label-free signal of temporal fidelity. The ``w/o State Tracking'' ablation (Table~\ref{tab:ablation}) supplies a fourth axis via downstream impact, where recall collapses once the signal is removed.

\begin{table}[h]
\centering
\caption{\textbf{Intermediate outputs of O-VAD across the three benchmarks.} For each dataset we report source video properties (duration, FPS) and per-stage intermediate outputs: Stage-1 grounded object count and Stage-2 detected state count, each summarized as min.\,/\,avg.\,/\,max.\ over the evaluation set.}
\label{tab:intermediate}
\setlength{\tabcolsep}{4pt}
\renewcommand{\arraystretch}{1.15}
\resizebox{\columnwidth}{!}{%
\begin{tabular}{l cc cc}
\toprule
& \multicolumn{2}{c}{\textbf{Video Info}}
& \multicolumn{2}{c}{\textbf{Intermediate Results (min.\,/\,avg.\,/\,max.)}} \\
\cmidrule(lr){2-3} \cmidrule(lr){4-5}
\textbf{Dataset}
& Duration (s) & FPS
& Stage 1: Grounding $\rightarrow$ \#Objects
& Stage 2: Tracking $\rightarrow$ \#States \\
\midrule
\textbf{Phys-AD}  & 0.98\,/\,2.10\,/\,4.00   & 60 & 1\,/\,2.31\,/\,12 & 1\,/\,10.61\,/\,73  \\
\textbf{LiquidAD} & 8.20\,/\,13.77\,/\,204.50 & 30 & 1\,/\,5.44\,/\,16 & 13\,/\,104.56\,/\,416 \\
\textbf{IPAD}     & 1.04\,/\,22.40\,/\,40.00  & 25 & 1\,/\,3.27\,/\,10 & 1\,/\,29.88\,/\,134  \\
\bottomrule
\end{tabular}%
}
\end{table}

\paragraph{Post-verification threshold sensitivity.}
($\tau_{\text{hi}},\tau_{\text{lo}},\tau_{\text{conf}}$), the post-verifier's three thresholds, encode a principled skip/verify/discard partition of the confidence space rather than three independent free parameters, leaving one effective hyperparameter (the verification band). The ``w/o Post-verifier'' ablation (Table~\ref{tab:ablation}) already shows AUROC drops up to 0.178 on sticky roller, confirming its role in recalibrating borderline cases; a $\pm0.1$ sensitivity sweep over each threshold leaves the retained anomaly set stable, indicating the pipeline is not sensitive to the exact cut points.

\subsection{Multi-Instance Re-Identification}
\label{sec:supp_reid}

\begin{figure}[h]
    \centering
    \includegraphics[width=0.4\linewidth]{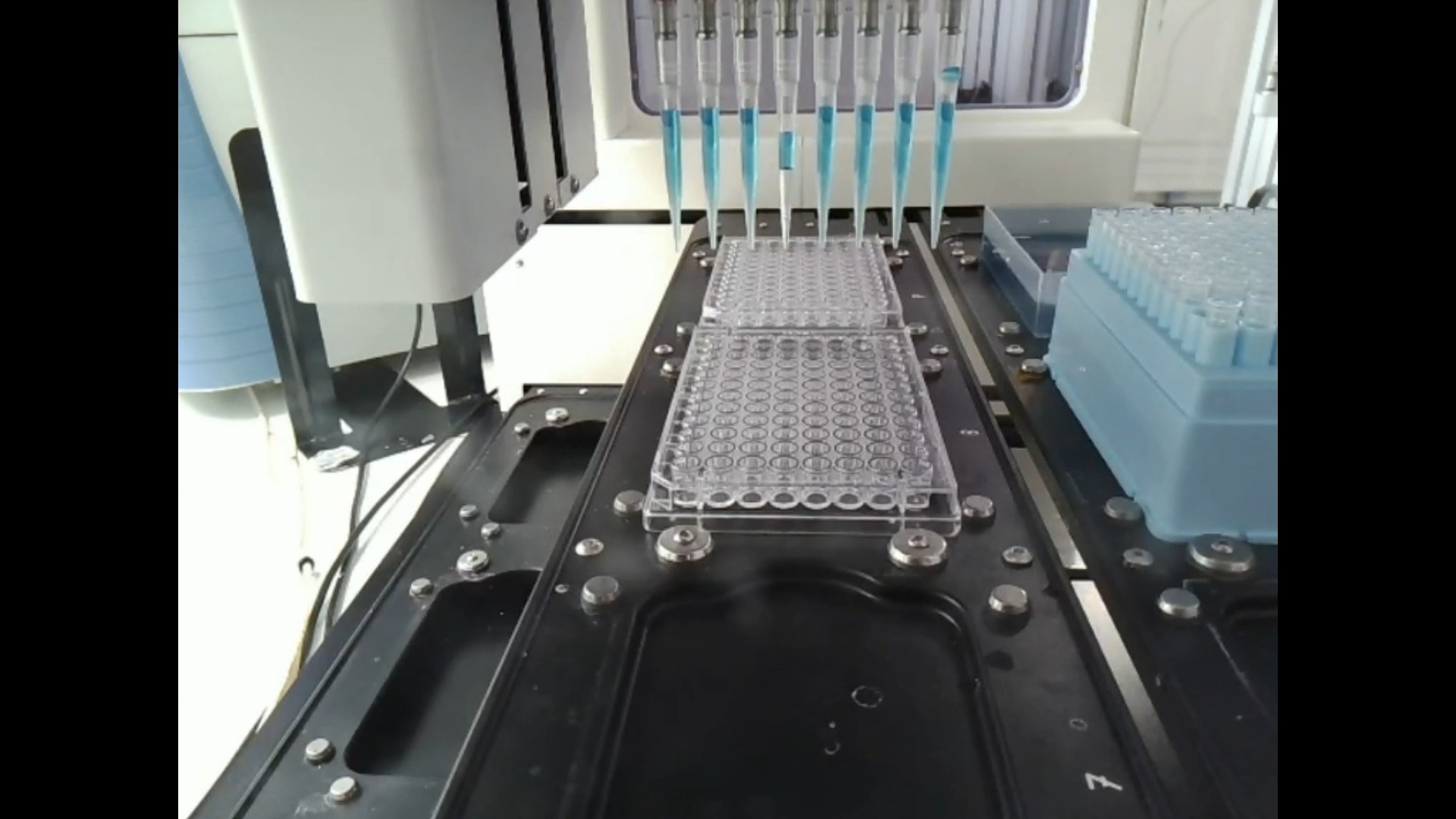}
    \includegraphics[width=0.4\linewidth]{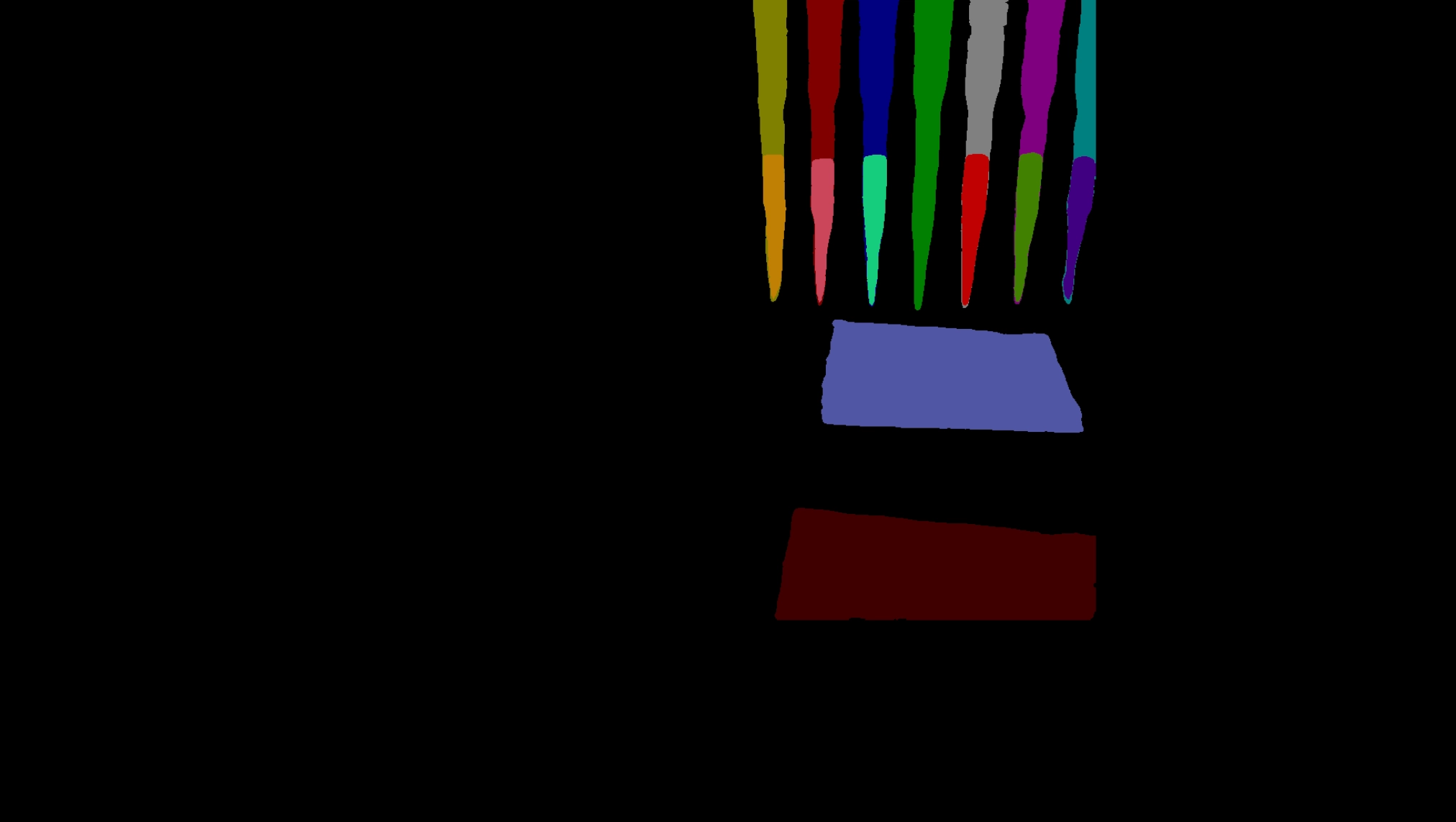} \\
    \includegraphics[width=0.4\linewidth]{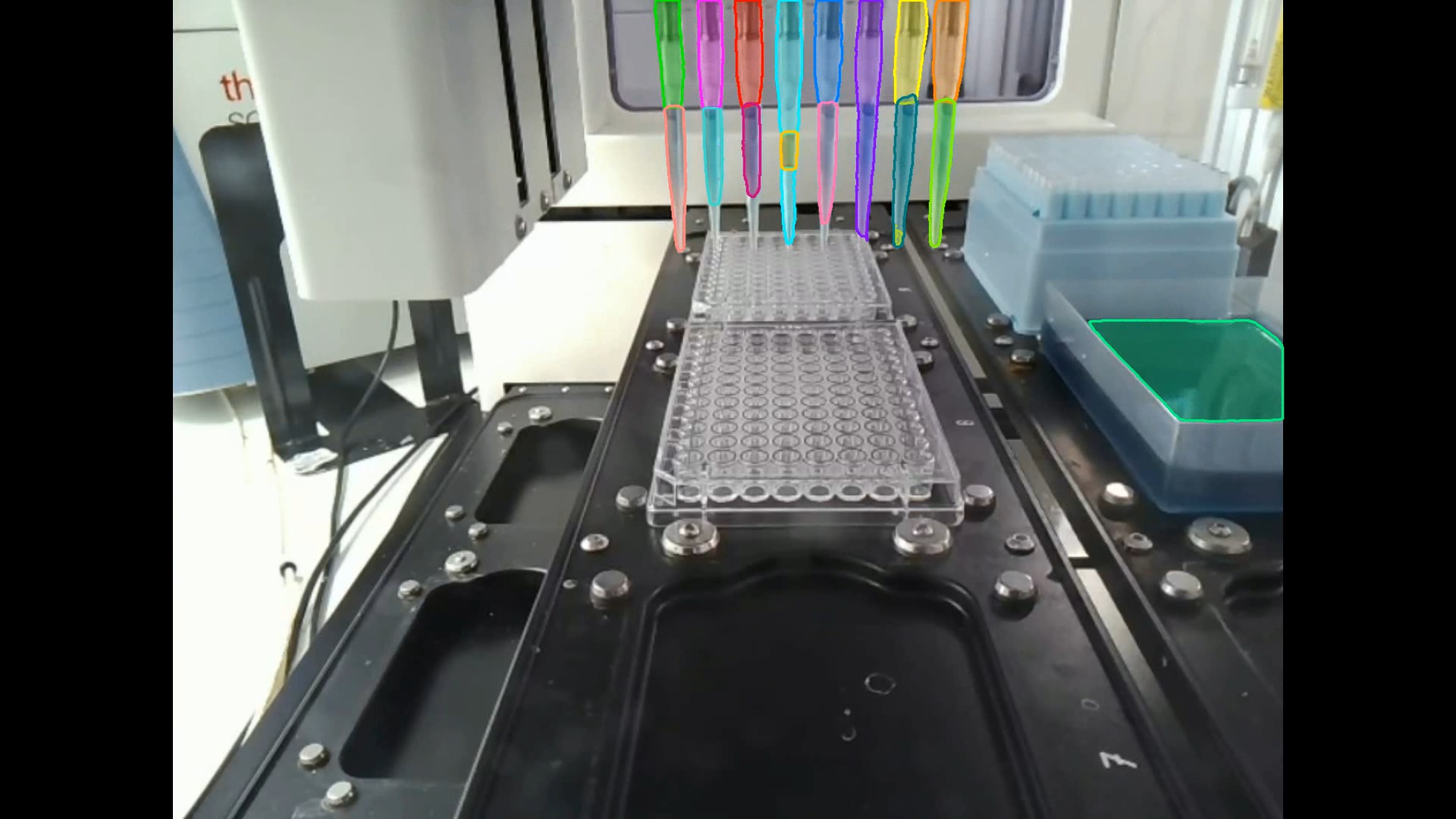}
    \includegraphics[width=0.4\linewidth]{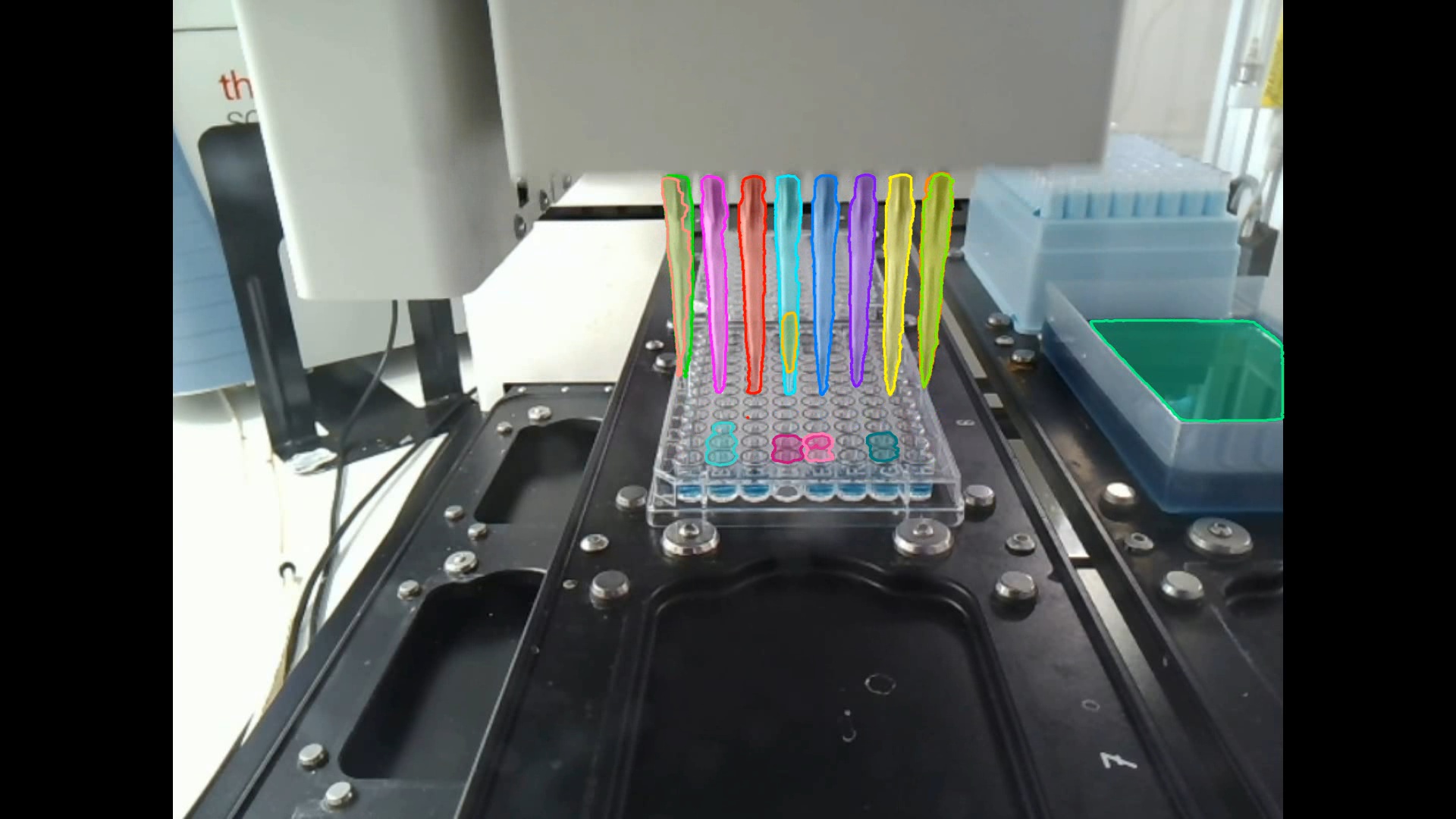}
    \caption{\textbf{Multi-instance re-ID on LiquidAD.} O-VAD's iterative detect--segment--mask-out scheme assigns one stable tubelet per pipette across all 8 visually-identical instances, including under partial occlusion by the dispensing head and during liquid transfer.}
    \label{fig:reid_pipettes}
\end{figure}

Production scenes with many look-alike instances (\eg, LiquidAD's 8 identical pipettes) raise two distinct sub-problems, each handled by a separate mechanism. \textit{Enumeration (Stage~1):} we extend the open-vocabulary detector with an iterative scheme---the VLM emits a class-level bounding box, the highest-confidence instance inside is segmented and its pixels are masked out, and detection is re-invoked on the masked image until no further instance is returned---yielding one mask per look-alike instance \emph{by construction}. \textit{Persistent identity (Stage~2):} each enumerated instance is propagated by SAM2's mask-memory attention, which maintains identity positionally rather than through appearance descriptors that would collapse on near-identical objects. Table~\ref{tab:intermediate} confirms that grounded counts match ground-truth cardinality on LiquidAD (8 pipettes per video), and Figure~\ref{fig:reid_pipettes} illustrates both stages under partial occlusion by the dispensing head and during liquid transfer.

\subsection{Comparative Reasoning Traces}
\label{app:traces}

\begin{figure}[t]
    \centering
    \includegraphics[width=1\linewidth]{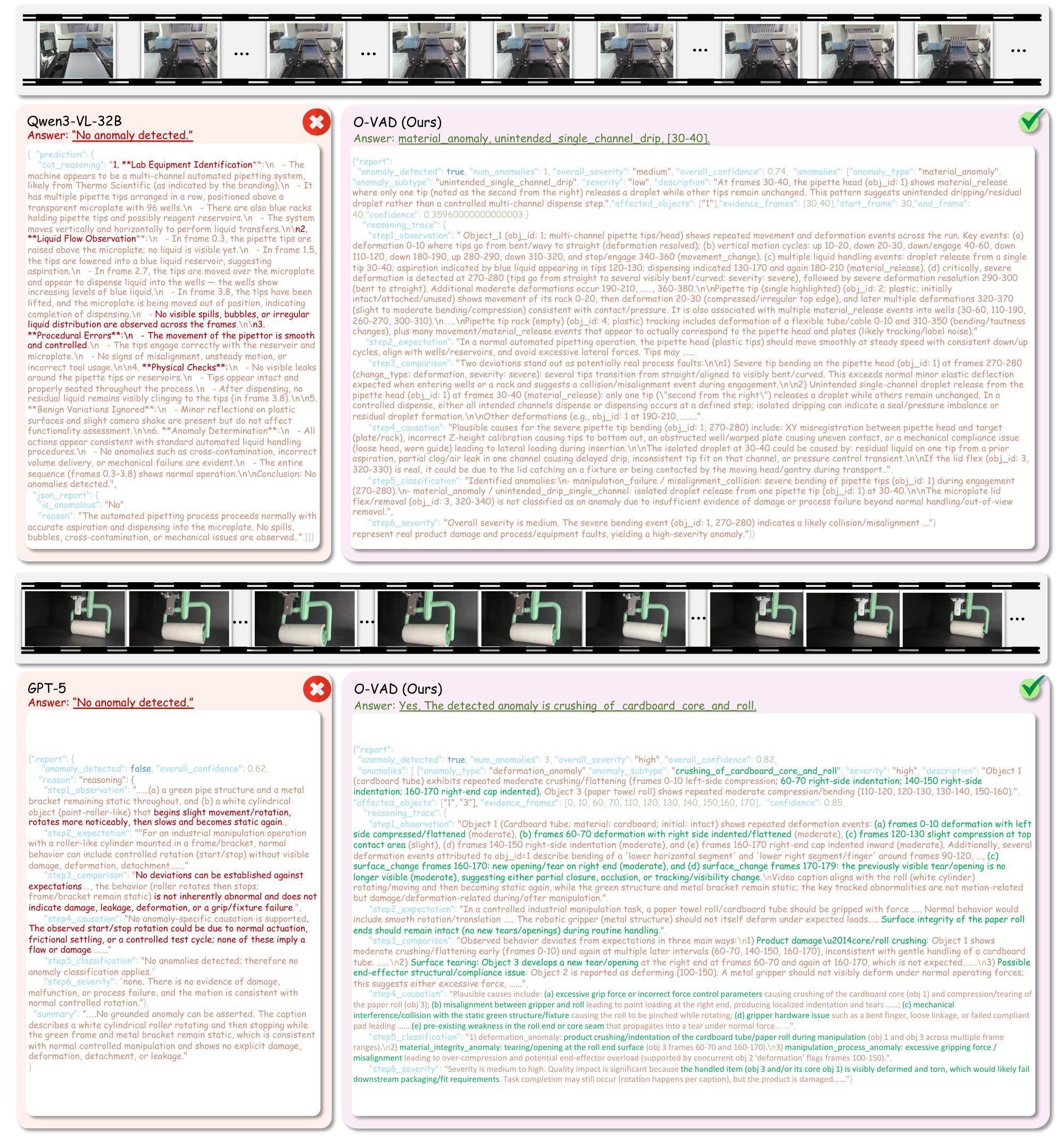}
    \caption{\textbf{Comparative reasoning traces on LiquidAD and Phys-AD.} \textit{Top:} An unintended single-channel drip during automated pipetting. Qwen3-VL-32B observes the full sequence but concludes normal operation, while O-VAD tracks per-tip state changes and identifies the isolated droplet release at frames 30--40. \textit{Bottom:} A sticky roller undergoes crushing during robotic manipulation. GPT-5 describes the rotation as normal and finds no deviation, while O-VAD tracks progressive deformation of the cardboard core and paper roll across multiple frame intervals, detecting crushing, surface tearing, and possible end-effector overload.}
    \label{fig:liquid-stickyroller}
\end{figure}

We compare the full reasoning traces of O-VAD against baseline VLMs on two representative success cases from LiquidAD and Phys-AD (Fig.~\ref{fig:liquid-stickyroller}).

\paragraph{LiquidAD: unintended single-channel drip.}
In this case, a multi-channel pipette head releases a droplet from only one tip at frames 30--40, while the remaining tips stay unchanged. Qwen3-VL-32B processes the full video and reports normal operation, noting that the movement is smooth, tips are properly seated, and no spills or bubbles are observed. Its frame-level descriptions confirm correct aspiration and dispensing but miss the isolated single-tip drip entirely. This is because the anomaly is localized to one channel among many, making it invisible to holistic frame-level inspection.
O-VAD, by contrast, tracks the pipette head as a single object with per-frame state annotations. At frames 30--40, the state tracker flags a material release event from only one tip (the second from the right). The reasoning chain then compares this against the expected multi-channel dispense pattern and concludes that an isolated drip is inconsistent with a controlled dispense step. It attributes the cause to a possible seal or pressure imbalance in that channel. This case demonstrates how object-centric tracking surfaces fine-grained per-component deviations that frame-level captioning overlooks.

\paragraph{Phys-AD: sticky roller crushing.}
Here, a robotic gripper manipulates a paper towel roll mounted on a cardboard core. GPT-5 observes the roller rotating and stopping, describes the behavior as consistent with normal controlled manipulation, and reports no anomaly. Its reasoning explicitly states that no deviations can be established against expectations.
O-VAD tracks three objects independently: the cardboard core (obj\_id 1), a metal gripper component (obj\_id 2), and the paper towel roll (obj\_id 3). It detects repeated crushing and flattening of the cardboard core at frames 0--10, 60--70, 140--150, and 160--170, along with compression and bending of the paper roll at frames 110--140. It further identifies a surface tear on the roll end at frames 60--70 and 160--170. The reasoning chain synthesizes these multi-object observations into a coherent diagnosis: excessive grip force or misalignment is causing progressive product damage, supported by concurrent deformation flags on the gripper itself (frames 100--150). GPT-5 misses all of these because its caption-level analysis describes only the high-level motion (rotation and stop) without examining per-object physical integrity.

Both cases illustrate the same bottleneck identified in the main paper: baseline VLMs generate fluent and plausible reasoning but lack fine-grained object-level evidence to anchor their judgments. O-VAD's structured state trajectories provide the missing evidential layer, enabling the reasoning chain to detect subtle, localized anomalies and to correlate changes across multiple objects for causal diagnosis.

\begin{figure}[t]
    \centering
    \includegraphics[width=1\linewidth]{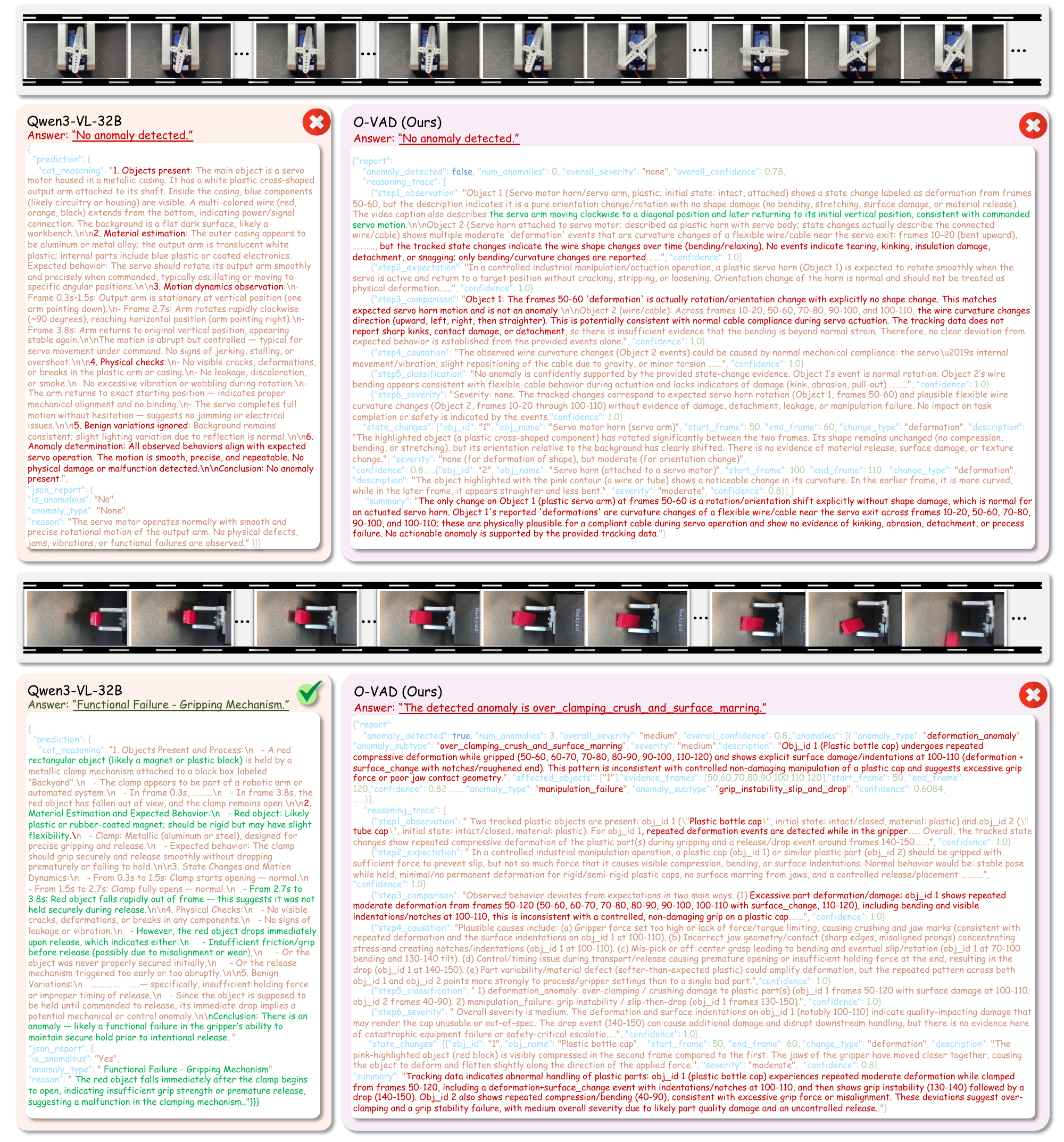}
    \caption{\textbf{Failure cases on servo and magnet categories.} \textit{Top:} A servo motor with a restricted rotation angle is misclassified as normal by both Qwen3-VL-32B and O-VAD. The observed small-angle motion is visually indistinguishable from a normal commanded rotation. \textit{Bottom:} A degaussed magnet falls off the board. Qwen3-VL-32B incorrectly attributes the anomaly to a gripping mechanism failure. O-VAD detects over-clamping deformation but fails to identify the root cause as loss of magnetism, since the magnet visually resembles an ordinary red plastic block.}
    \label{fig:servo-magnet}
\end{figure}

\begin{figure}[t]
    \centering
    \includegraphics[width=1\linewidth]{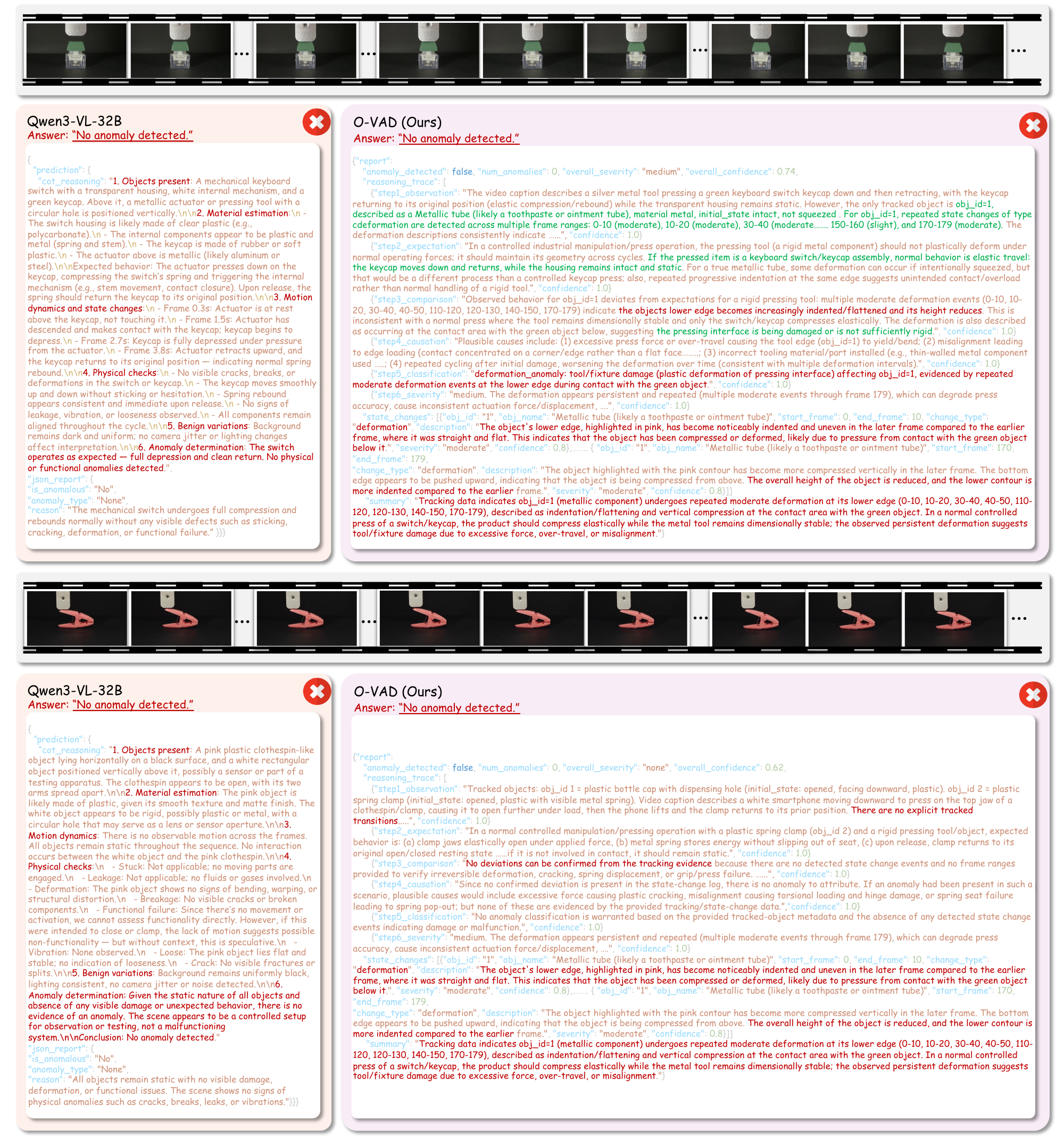}
    \caption{\textbf{Failure cases on button and clip categories.} \textit{Top:} A stuck button that fails to actuate is misclassified as normal by both methods. O-VAD detects repeated deformation of the pressing tool but attributes it to tool/fixture damage rather than recognizing the button's inability to depress. \textit{Bottom:} A clip that cannot be pressed shows no observable state changes, leading both methods to predict normal behavior despite the ground-truth anomaly.}
    \label{fig:buttion-clip}
\end{figure}

\subsection{Failure Cases Analysis}
\label{app:failcases}

We present four representative failure cases that reveal the fundamental limitations of perception-driven anomaly reasoning without predefined expert knowledge. These cases fall into two categories: \textit{specification-dependent anomalies}, where the ground truth is defined by quantitative thresholds invisible to visual inspection, and \textit{perception-ambiguous anomalies}, where the defect produces visually plausible behavior indistinguishable from normality.

\paragraph{Specification-dependent failures.}
In the servo angle-restricted case (Fig.~\ref{fig:servo-magnet}, top), the servo motor rotates to a small angle and returns smoothly. Both Qwen3-VL-32B and O-VAD classify this as normal, since the observed motion is visually consistent with a correctly functioning servo. The anomaly is a restricted rotation range, which is defined by a quantitative angular specification entirely absent from the visual evidence. Without knowing the \textit{expected} rotation angle, neither state tracking nor commonsense reasoning can flag this deviation.
Similarly, in the clip unable-to-press case (Fig.~\ref{fig:buttion-clip}, bottom), the pressing tool descends and retracts while the clip remains static. O-VAD detects no state change events for the clip and concludes no anomaly is present. The ground truth is that the clip \textit{should} have been pressed but was not. This requires knowledge of the expected actuation outcome, which is not visually inferrable.

\paragraph{Perception-ambiguous failures.}
In the button stuck case (Fig.~\ref{fig:buttion-clip}, top), O-VAD tracks repeated deformation of the metallic pressing tool across multiple frame intervals. It flags this as tool/fixture damage due to excessive force. However, the actual anomaly is the button's failure to actuate: the pressing tool deforms precisely \textit{because} the button is stuck. Both a stuck button and a normally stiff button produce similar visual signatures of tool compression. O-VAD captures the correct \textit{symptom} (tool deformation) but misattributes the \textit{cause}, illustrating how accurate state tracking can still lead to incorrect classification when the root cause requires understanding the intended function of the pressed object.

In the magnet degaussing case (Fig.~\ref{fig:servo-magnet}, bottom), a degaussed magnet detaches from the board during manipulation. Qwen3-VL-32B attributes this to a gripping mechanism failure. O-VAD detects over-clamping deformation and grip instability, identifying the anomaly as crushing and dropping rather than loss of magnetism. Since the magnet visually resembles an ordinary red plastic block, neither method can infer its magnetic properties. The detachment is caused by insufficient magnetic adhesion, not mechanical failure, but this distinction is invisible to visual perception.

These failure cases share a common pattern: the ground-truth anomaly is defined by an invisible physical property (magnetic strength, required actuation force, angular specification) or requires functional knowledge of what the object is supposed to do. Such information cannot be recovered from visual appearance alone. While O-VAD's object-centric state tracking correctly captures observable symptoms in most cases, it cannot bridge the gap between perceptual evidence and specification-level judgments. This motivates future work on integrating few-shot specification examples or lightweight expert priors into the reasoning pipeline.

\subsection{Human Evaluation}

\begin{table}[t]
\centering
\caption{\textbf{Human evaluation questionnaire.} All dimensions use a 1--5 Likert scale (1\,=\,strongly disagree, 5\,=\,strongly agree) except Q10 (preference ranking).}
\label{tab:questionnaire}
\setlength{\tabcolsep}{4pt}
\scriptsize
\begin{tabular}{cl p{7.8cm} c}
\toprule
\textbf{Part} & \textbf{ID} & \textbf{Dimension \& Description} & \textbf{Scale} \\
\midrule
\multirow{4}{*}{\rotatebox{90}{\scriptsize Det.}}
& Q1 & \textbf{Binary correctness.} Correctly identifies anomalous vs.\ normal? & 1--5 \\
& Q2 & \textbf{Type correctness.} Predicted type matches the ground-truth defect?\textsuperscript{$\dagger$} & 1--5 \\
& Q3 & \textbf{Object identification.} Affected object(s) correctly identified?\textsuperscript{$\ddagger$} & 1--5 \\
& Q4 & \textbf{Temporal localization.} Anomaly accurately localized in time?\textsuperscript{$\ddagger$} & 1--5 \\
\midrule
\multirow{3}{*}{\rotatebox{90}{\scriptsize Expl.}}
& Q5 & \textbf{Faithfulness.} Explanation reflects actual video content, free of hallucinations. & 1--5 \\
& Q6 & \textbf{Completeness.} All salient state changes covered without significant omissions. & 1--5 \\
& Q7 & \textbf{Causal coherence.} Causal reasoning is logically sound and evidence-consistent. & 1--5 \\
\midrule
{\scriptsize Act.}
& Q8 & \textbf{Diagnostic usefulness.} Operator could act on the report without re-watching the video. & 1--5 \\
\midrule
{\scriptsize Ovr.}
& Q9 & \textbf{Overall quality.} Overall quality of the anomaly report. & 1--5 \\
\midrule
{\scriptsize Pref.}
& Q10 & \textbf{Preference.} Rank all method reports for the same video for real deployment. & Rank 1--3 \\
\bottomrule
\multicolumn{4}{l}{\scriptsize \textsuperscript{$\dagger$}\,Scored only when both ground truth and prediction are anomalous. \textsuperscript{$\ddagger$}\,Scored only when provided.} \\
\end{tabular}
\end{table}

\paragraph{Human Evaluation Protocol.}
\label{app:human_eval}

To assess the practical utility of anomaly reports beyond automated metrics, we conduct a human evaluation study with $N{=}5$ domain experts (graduate researchers with industrial vision or robotics experience). Each evaluator independently reviews anomaly reports generated by three methods---Qwen3-VL-32B (direct prompting), URF-ZS-HVAA, and O-VAD---on 10 randomly sampled videos from the Phys-AD test set, spanning 10 object categories. Reports are anonymized and presented in randomized order to prevent method-identification bias. Each evaluator completes the questionnaire below for every report.

\paragraph{Instructions.} 
Each evaluator should independently review anomaly reports generated by three methods on 10 randomly sampled videos. 
For each video--report pair, first watch the video, then read reports, and rate each of the following on a 5-point Likert scale: \texttt{1}~(strongly disagree) to \texttt{5}~(strongly agree).

\paragraph{\textbf{Evaluation Questionnaire.}}
\label{app:questionnaire}

The questionnaire (Table~\ref{tab:questionnaire}) comprises 10 questions organized into five parts that progressively assess report quality from factual correctness to practical utility. Part~I (Q1--Q4) measures detection correctness at four granularities: binary anomaly/normal classification, semantic anomaly type matching, affected object identification, and temporal localization. Part~II (Q5--Q7) evaluates explanation quality along three complementary axes: faithfulness (no hallucinations), completeness (no omissions), and causal coherence (logically sound reasoning). Part~III (Q8) assesses actionability, i.e., whether the report alone suffices for an operator to diagnose and respond. Part~IV (Q9) captures a holistic quality judgment. Finally, Part~V (Q10) elicits a direct pairwise preference ranking across all methods for the same video, yielding a deployment-oriented win rate. All dimensions except Q10 use a 1--5 Likert scale; Q10 produces a rank ordering from best to worst.

\paragraph{\textbf{Human Evaluation Results}}
\label{sec:human_eval_results}

We report results from the human evaluation study described in Appendix~\ref{app:human_eval}. Five domain experts independently rated anomaly reports from three methods---URF-ZS-HVAA (Report~1), Qwen3-VL-32B (Report~2), and O-VAD (Report~3)---on 10 Phys-AD videos (7 anomalous, 3 normal) across 9 Likert-scale dimensions (Q1--Q9) and a pairwise preference ranking (Q10). Table~\ref{tab:quali_eval} summarizes the results.

\paragraph{Overall comparison.}
O-VAD achieves the highest scores on all nine dimensions and is ranked first by evaluators in 82\% of cases (41 out of 50 evaluator--video pairs). Its average detection score (4.42) exceeds Qwen3-VL-32B~\cite{bai2025qwen3} (2.89) by +1.53 points and URF-ZS-HVAA~\cite{lin2025unified} (1.14) by +3.28 points. The gap is similarly pronounced for explanation quality (+1.11 over Qwen3, +3.32 over URF-ZS) and actionability (+1.32 over Qwen3, +3.16 over URF-ZS). URF-ZS-HVAA consistently scores near the floor (1.0--1.2) across all dimensions, confirming that its chained reasoning framework, despite incorporating temporal detection and spatial localization modules, produces reports that evaluators judge as almost entirely uninformative for industrial anomaly diagnosis. This is consistent with its near-zero recall in the quantitative evaluation 
(Table 1 and Table 2).

\paragraph{Detection correctness.}
The most striking result is on binary detection (Q1), where O-VAD scores 4.98 out of 5---near-perfect agreement that it correctly identifies anomalous vs.\ normal videos. In contrast, URF-ZS-HVAA scores only 1.24, indicating that evaluators find its predictions almost always incorrect. Qwen3-VL-32B scores 2.70, performing better than URF-ZS but still far below O-VAD. O-VAD also leads on anomaly type identification (Q2: 4.26 vs.\ 2.60 vs.\ 1.08) and temporal localization (Q4: 3.96 vs.\ 2.32 vs.\ 1.08), though these finer-grained dimensions show more variance, reflecting the inherent difficulty of open-ended type classification and frame-level precision.

\paragraph{Explanation quality and actionability.}
Among the explanation dimensions, O-VAD's largest advantage is on completeness (Q6: 4.58 vs.\ 3.36 vs.\ 1.00), indicating that evaluators find its state-trajectory-grounded reports cover all relevant state changes without significant omissions. Causal coherence (Q7: 4.30 vs.\ 3.02 vs.\ 1.00) shows the second-largest gap, confirming that the structured chain-of-thought reasoning produces more logically sound root-cause analyses. On diagnostic usefulness (Q8), O-VAD scores 4.16, meaning evaluators believe an operator could largely act on the report without re-watching the video, while URF-ZS-HVAA's score of 1.00 indicates its reports provide no actionable information.


\paragraph{Anomalous vs.\ normal cases.}
On the 7 anomalous videos, O-VAD's detection average (4.40) substantially exceeds Qwen3-VL-32B (3.19) and URF-ZS-HVAA (1.00). On the 3 normal videos, O-VAD achieves a perfect binary correctness score of 5.00, whereas both baselines score below 1.80 on Q1, indicating severe false-positive tendencies. Notably, Qwen3-VL-32B's binary correctness drops to 1.13 on normal cases (compared to 3.37 on anomalous ones), suggesting it defaults to predicting anomalies. O-VAD's robustness on normal videos stems from its evidence-grounded reasoning: without tracked state changes that violate process expectations, the chain-of-thought correctly concludes normality.

\subsection{LLM-as-Judge}
\label{app:llm_judge}

\paragraph{LLM-as-Judge Evaluation Protocol.}

To enable scalable evaluation beyond the 10-video human study, we design an LLM-as-judge~\cite{zheng2023judging} protocol that mirrors the human questionnaire (\S\ref{app:questionnaire}) and can be applied to the full test set. We use GPT-4o as the judge model with structured prompts.

\begin{table}[t]
\centering
\caption{\textbf{Human and LLM-as-judge evaluation results} (mean scores, 1--5 scale). Human: $N{=}5$ evaluators $\times$ 10 videos; LLM: GPT-4o judge $\times$ 10 videos. URF-ZS: URF-ZS-HVAA~\cite{lin2025unified}; Qwen3: Qwen3-VL-32B~\cite{bai2025qwen3}. Q4 is excluded from LLM evaluation due to the absence of temporal ground truth. Best per row per evaluator is \textbf{bold}; ties share bold.}
\label{tab:quali_eval}
\setlength{\tabcolsep}{3.5pt}
\small
\resizebox{\textwidth}{!}{%
\begin{tabular}{cl ccc l ccc}
\toprule
& & \multicolumn{3}{c}{\textbf{Human Evaluation}} & & \multicolumn{3}{c}{\textbf{LLM-as-Judge}} \\
\cmidrule(lr){3-5}\cmidrule(lr){7-9}
\textbf{Part} & \textbf{Dimension} & \textbf{URF-ZS} & \textbf{Qwen3} & \textbf{O-VAD} & & \textbf{URF-ZS} & \textbf{Qwen3} & \textbf{O-VAD} \\
\midrule
\multirow{4}{*}{\rotatebox{90}{\scriptsize Det.}}
& Q1. Binary correctness      & 1.24 & 2.70 & \textbf{4.98} & & 1.40 & 2.60 & \textbf{4.60} \\
& Q2. Type correctness         & 1.08 & 2.60 & \textbf{4.26} & & 1.00 & \textbf{1.80} & \textbf{1.80} \\
& Q3. Object identification    & 1.14 & 3.94 & \textbf{4.46} & & 1.00 & 2.60 & \textbf{3.20} \\
& Q4. Temporal localization    & 1.08 & 2.32 & \textbf{3.96} & & \multicolumn{3}{c}{\textit{N/A}} \\
\cmidrule(lr){2-5}\cmidrule(lr){7-9}
& \textit{Detection avg.}      & 1.14 & 2.89 & \textbf{4.42} & & 1.13 & 2.33 & \textbf{3.20} \\
\midrule
\multirow{3}{*}{\rotatebox{90}{\scriptsize Expl.}}
& Q5. Faithfulness              & 1.04 & 3.28 & \textbf{4.10} & & 1.30 & \textbf{4.90} & 3.90 \\
& Q6. Completeness              & 1.00 & 3.36 & \textbf{4.58} & & 1.20 & \textbf{4.00} & 3.90 \\
& Q7. Causal coherence          & 1.00 & 3.02 & \textbf{4.30} & & 1.20 & \textbf{3.70} & 3.60 \\
\cmidrule(lr){2-5}\cmidrule(lr){7-9}
& \textit{Explanation avg.}     & 1.01 & 3.22 & \textbf{4.33} & & 1.23 & \textbf{4.20} & 3.80 \\
\midrule
{\scriptsize Act.}
& Q8. Diagnostic usefulness     & 1.00 & 2.84 & \textbf{4.16} & & 1.10 & 2.30 & \textbf{3.40} \\
\midrule
{\scriptsize Ovr.}
& Q9. Overall quality           & 1.04 & 3.00 & \textbf{4.28} & & 1.20 & 2.70 & \textbf{3.50} \\
\midrule
{\scriptsize Pref.}
& Q10. Win rate (\%)           & 0.0  & 18.0 & \textbf{82.0} & & 10.0 & 30.0 & \textbf{70.0} \\
\bottomrule
\end{tabular}
}
\end{table}

\paragraph{\textbf{LLM-as-Judge Results.}}
\label{sec:llm_judge_results}

To complement the human evaluation and scale assessment to a larger sample, we apply the LLM-as-judge protocol using GPT-4o as the judge model across the same 10 Phys-AD videos. Note that Q4 (temporal localization) is excluded from this evaluation due to the lack of frame-level ground-truth annotations in the judging set. Table~\ref{tab:quali_eval} reports the results.

\paragraph{Overall comparison.}
O-VAD is ranked first by the LLM judge in 70\% of cases, followed by Qwen3-VL-32B at 30\% and URF-ZS-HVAA at 10\%. O-VAD achieves the highest overall quality (Q9: 3.50 vs.\ 2.70 vs.\ 1.20) and diagnostic usefulness (Q8: 3.40 vs.\ 2.30 vs.\ 1.10), consistent with the human evaluation trends. URF-ZS-HVAA again scores near the floor across all dimensions (1.00--1.40), confirming its inability to produce meaningful anomaly reports regardless of the evaluation paradigm.
 
\paragraph{Detection vs.\ explanation: a revealing split.}
The most notable finding is the divergence between detection correctness and explanation quality. O-VAD leads convincingly on detection (avg.\ 3.20 vs.\ 2.33 vs.\ 1.13) and binary correctness in particular (Q1: 4.60 vs.\ 2.60 vs.\ 1.40). However, Qwen3-VL-32B achieves higher explanation scores than O-VAD on all three dimensions: faithfulness (4.90 vs.\ 3.90), completeness (4.00 vs.\ 3.90), and causal coherence (3.70 vs.\ 3.60). This reversal, absent in the human evaluation where O-VAD leads on all explanation metrics, reflects a known LLM-judge bias toward fluent, well-structured prose~\cite{zheng2023judging}. Qwen3-VL-32B generates polished natural-language narratives even when its detection is incorrect, which the LLM judge rewards on explanation dimensions evaluated independently of correctness. Human experts, by contrast, penalize explanations that describe the wrong conclusion, regardless of fluency. Despite this explanation advantage, Qwen3-VL-32B's poor detection correctness (Q1: 2.60) and low actionability (Q8: 2.30) result in a lower overall quality score and win rate, indicating that the LLM judge's holistic assessment still correctly prioritizes factual accuracy.
 
\paragraph{Anomaly type correctness remains challenging.}
Both O-VAD and Qwen3-VL-32B tie on type correctness (Q2: 1.80), the lowest-scoring dimension for both methods. This reflects the inherent difficulty of open-ended anomaly type classification: even when detection is correct, precisely matching the ground-truth anomaly taxonomy remains challenging for all methods, a gap that motivates future work on domain-adaptive type reasoning.
 
\paragraph{\textbf{Consistency with Human Evaluation}.}
The LLM-as-judge results broadly corroborate the human evaluation findings: O-VAD ranks first overall, URF-ZS-HVAA is consistently the weakest, and O-VAD's primary advantage lies in detection correctness and actionability. The main discrepancy is on explanation quality, where the LLM judge favors Qwen3-VL-32B's fluent but often factually incorrect narratives. This highlights a complementary strength of human evaluation: experts naturally integrate correctness into their explanation judgments, producing assessments more aligned with real deployment requirements.

\section{Algorithm of O-VAD}

Algorithm~\ref{alg:ovad} summarizes the complete O-VAD pipeline. Stage~1 discovers objects via VLM prompting and segments them with SAM, Stage~2 tracks each object through transformations and queries the VLM on sampled frame pairs to detect open-ended state changes, and Stage~3 performs a six-step chain-of-thought reasoning pass followed by confidence-tiered visual verification to produce the final anomaly report. All three stages are training-free and require no predefined taxonomy.

\begin{algorithm}[H]
\caption{O-VAD: Object-Centric Video Anomaly Detection}
\label{alg:ovad}
\scriptsize  
\begin{algorithmic}[1]

\REQUIRE Video $\mathcal{V}=\{I_t\}_{t=1}^T$, optional caption $\texttt{cap}$, optional task context $\texttt{ctx}$
\ENSURE Anomaly report $\mathcal{R}$

\STATE
\STATE \textcolor{gray}{\textit{// Stage~1: Automated Object Grounding }}
\STATE Sample candidate frames $\{I_{t_k}\}_{k=1}^K$ from $\mathcal{V}$
\FOR{each candidate frame $I_{t_k}$}
    \STATE $\mathcal{O}_{t_k} \leftarrow \phi_{\text{VLM}}(I_{t_k})$
        \hfill \textcolor{gray}{$\triangleright$ object inventory: names, descriptions, spatial cues}
\ENDFOR
\STATE $\mathcal{O} \leftarrow \bigcup_k \mathcal{O}_{t_k}$
    \hfill \textcolor{gray}{$\triangleright$ merge and deduplicate}
\FOR{each object $o_i \in \mathcal{O}$}
    \STATE $\mathcal{M}_1^i \leftarrow \text{SAM}(I_1, o_i)$
        \hfill \textcolor{gray}{$\triangleright$ grounded segmentation}
\ENDFOR

\STATE
\STATE \textcolor{gray}{\textit{// Stage~2: Object-Centric State Tracking }}
\STATE \textcolor{gray}{\textit{// \quad Spatiotemporal partitioning \& tubelet construction~\cite{sun2025tracking}}}
\STATE $\mathcal{P} \leftarrow \text{TubeletGraph}(\mathcal{V},\, \mathcal{M}_1)$
    \hfill \textcolor{gray}{$\triangleright$ partition video into tubelets}
\STATE \textcolor{gray}{\textit{// \quad State tracking: recover missing tracks}}
\FOR{each object $o_i$ with prompt track $P_i$}
    \STATE $\mathcal{T}_i \leftarrow P_i \cup \{C \!\in\! \mathcal{P} \mid S_{\text{prox}}(C,P_i)\!>\!\tau_{\text{prox}} \;\wedge\; S_{\text{sem}}(C,P_i)\!>\!\tau_{\text{sem}}\}$
\ENDFOR
\STATE \textcolor{gray}{\textit{// \quad Initial state identification}}
\FOR{each object $o_i$ with track $\mathcal{T}_i$}
    \STATE $\tilde{I}_1^i \leftarrow \text{Highlight}(I_1, \mathcal{M}^i_1)$
    \STATE $(\texttt{desc}_i,\, \texttt{state}_i,\, \texttt{mat}_i) \leftarrow \phi_{\text{VLM}}(\tilde{I}_1^i)$
        \hfill \textcolor{gray}{$\triangleright$ object name, state, material}
\ENDFOR
\STATE \textcolor{gray}{\textit{// \quad State change detection and understanding (ours)}}
\STATE $\mathcal{E} \leftarrow \emptyset$
\FOR{each object $o_i$ with track $\mathcal{T}_i$}
    \FOR{each sampled pair $(t_k,\, t_{k+1})$ from $\mathcal{T}_i$}
        \STATE $\tilde{I}_{t_k} \leftarrow \text{Highlight}(I_{t_k}, \mathcal{M}^i_{t_k})$;\quad $\tilde{I}_{t_{k+1}} \leftarrow \text{Highlight}(I_{t_{k+1}}, \mathcal{M}^i_{t_{k+1}})$
        \STATE $(d,\, \texttt{ty},\, \texttt{ca},\, \texttt{desc},\, \texttt{sev}) \leftarrow \phi_{\text{VLM}}(\tilde{I}_{t_k}, \tilde{I}_{t_{k+1}})$
            \hfill \textcolor{gray}{$\triangleright$ open-ended state change query}
        \IF{$d = \texttt{yes}$}
            \STATE $\mathcal{E} \leftarrow \mathcal{E} \cup \{(t_k,\, t_{k+1},\, \texttt{ty},\, \texttt{ca},\, \texttt{desc},\, \texttt{sev},\, i)\}$
        \ENDIF
    \ENDFOR
\ENDFOR

\STATE
\STATE \textcolor{gray}{\textit{// Stage~3: State-Aware Anomaly Reasoning }}
\STATE
\STATE \textcolor{gray}{\textit{// \quad Cognitive anomaly reasoning (single VLM call, Steps 1--6)}}
\STATE Sample key frames $\mathcal{F} = \{I_1,\, I_{\lfloor T/2 \rfloor},\, I_T\}$ from $\mathcal{V}$
\STATE $(\texttt{reasoning},\, \mathcal{A}_{\text{raw}}) \leftarrow \phi_{\text{VLM}}\!\big(\mathcal{O},\, \mathcal{E},\, \texttt{ctx},\, \texttt{cap},\, \mathcal{F}\big)$
    \hfill \textcolor{gray}{\parbox[t]{0.42\linewidth}{\raggedright
    $\triangleright$ Steps 1--6: observe $\to$ expect $\to$ compare $\to$ cause $\to$ classify $\to$ severity}}
\STATE
\STATE \textcolor{gray}{\textit{// \quad Post visual verification}}
\FOR{each anomaly $a \in \mathcal{A}_{\text{raw}}$}
    \IF{$c_{\text{orig}}(a) \geq \tau_{\text{hi}}$}
        \STATE Retain $a$
            \hfill \textcolor{gray}{$\triangleright$ high-confidence: skip verification}
    \ELSIF{$c_{\text{orig}}(a) \leq \tau_{\text{lo}}$}
        \STATE Discard $a$
            \hfill \textcolor{gray}{$\triangleright$ low-confidence: auto-discard}
    \ELSE
        \STATE $(\texttt{verified},\, c_{\text{ver}}) \leftarrow \phi_{\text{VLM}}(\mathcal{F},\, \texttt{cap},\, a)$
            \hfill \textcolor{gray}{$\triangleright$ check anomaly evidence in caption/frames}
        \IF{$\texttt{verified} = \texttt{true}$}
            \STATE $c_{\text{final}}(a) \leftarrow c_{\text{orig}}(a) \cdot c_{\text{ver}}$
        \ELSE
            \STATE $c_{\text{final}}(a) \leftarrow c_{\text{orig}}(a) \cdot (1 - c_{\text{ver}})$
        \ENDIF
        \IF{$c_{\text{final}}(a) < \tau_{\text{conf}}$}
            \STATE Discard $a$
                \hfill \textcolor{gray}{$\triangleright$ low post-verification confidence}
        \ENDIF
    \ENDIF
\ENDFOR
\STATE $\mathcal{A} \leftarrow$ retained anomalies
\STATE
\RETURN Report $\mathcal{R} = (\mathcal{A},\, \mathcal{E},\, \texttt{reasoning},\, \texttt{summary})$

\end{algorithmic}
\end{algorithm}

\end{document}